\documentclass[11pt]{article}
\usepackage{fullpage}

\usepackage[english]{babel}
\usepackage{xspace}
\usepackage{algorithm,algorithmicx}
\usepackage{tcolorbox}
\usepackage{tikz}

\usepackage[colorlinks=true,
linkcolor=blue,
urlcolor=blue,
citecolor=blue]{hyperref}

\usepackage[utf8]{inputenc} 
\usepackage[T1]{fontenc}    
\usepackage{hyperref}       
\usepackage{url}            
\usepackage{booktabs}       
\usepackage{amsfonts}  
\usepackage{amsmath}
\usepackage{nicefrac}       
\usepackage{microtype}      
\usepackage{xcolor, colortbl}        
\usepackage{wrapfig}
\usepackage{caption}
\usepackage{subcaption}
\usepackage{footnote}
\makesavenoteenv{tabular}
\makesavenoteenv{table}
\usepackage{multirow}
\usepackage{bbm}
\usepackage{amsthm}
\usepackage[makeroom]{cancel}
\usepackage{stmaryrd}
\usepackage{amsmath,bm}
\usepackage{amssymb}
\usepackage{mathtools, cuted}
\usepackage{kantlipsum,setspace}
\usepackage{enumitem}
\usepackage{algorithm}
\usepackage{algpseudocode}
 
\definecolor{Gray}{gray}{0.85}

\def\ee{\mathbb{E}}

\newtheorem{lemma}{Lemma}

\newtheorem{theorem}{Theorem}

\newtheorem{assumption}{Assumption}

\newtheorem{Corollary}{Corollary}

\title{Structural Estimation of Markov Decision Processes in High-Dimensional State Space with Finite-Time Guarantees}

\author{\large Siliang Zeng$^\dagger$, Mingyi Hong$^\dagger$, Alfredo Garcia$^{\ddagger}$\\[.5cm]
	\small $^{\dagger}$Department  of Electrical and Computer Engineering, \\
	\small University of Minnesota, MN, USA\\
	\small $^{\ddagger}$ Department of Industrial and Systems Engineering,\\
	\small Texas A\&M University, TX, USA\\
	\small Email: \texttt{\{zeng0176, mhong\}@umn.edu}, \texttt{alfredo.garcia@tamu.edu}} 

\begin{document}

\maketitle

\begin{abstract}
   We consider the task of estimating a structural model of dynamic decisions by a human
agent based upon the observable history of implemented actions and visited states. This problem has an inherent nested structure: in the inner problem, an optimal policy for a given reward function is identified while in the outer problem, a measure of fit is maximized.
Several approaches have been proposed to alleviate the computational burden of this nested-loop structure, but these methods still suffer from high complexity when the state space is either discrete with large cardinality or continuous in high dimensions. Other approaches in the inverse reinforcement learning (IRL) literature emphasize policy estimation at the expense of reduced reward estimation accuracy.
In this paper we propose a {\em single-loop} estimation algorithm with finite time guarantees that is equipped to deal with high-dimensional state spaces without compromising reward estimation accuracy. In the proposed
algorithm, each policy improvement step is followed by a stochastic gradient step for likelihood maximization. We show the proposed algorithm converges to a stationary solution with a finite-time guarantee. 
{Further, if the reward is parameterized linearly, the algorithm approximates the maximum likelihood estimator sublinearly.}
For robotics control problems
in MuJoCo and their transfer settings, the proposed algorithm achieves superior performance compared with other IRL and imitation learning benchmarks.\footnote{This paper is the journal version of \cite{zeng2022maximum}}
\end{abstract}

\section{Introduction}
\label{intro}

We consider the task of estimating a structural model of dynamic decisions by a single human agent based upon the observable history of implemented actions and visited states. This problem has been studied as the estimation of dynamic discrete choice (DDC) models in econometrics and inverse reinforcement learning (IRL) in artificial intelligence and machine learning research.

\cite{Rust} is a seminal piece in the literature on dynamic discrete choice estimation. In that paper, the estimation task is formulated as a bi-level optimization problem where the inner problem is a stochastic dynamic programming problem and the outer problem is the likelihood maximization of observed actions and states.
\cite{Rust} proposed an iterative nested fixed point algorithm
in which the inner dynamic programming problem is solved repeatedly followed by maximum likelihood updates of the structural parameters. Over the years, a significant literature on alternative estimation methods requiring less computational effort has been developed. For example, \cite{Hotz} and \cite{Hotz1994} proposed two-step algorithms which avoid the repeated solution of the inner stochastic dynamic programming problem. In the first step, a non-parametric estimator of the policy (also referred to as conditional choice probabilities) is obtained and the inverse of a map from differences in Bellman's value function for different states to randomized policies is computed. In the second step, a pseudo log-likelihood is maximized. Two-step estimators may suffer from substantial finite sample bias if the estimated policies in the first step are of poor quality. Sequential estimators which are recursively obtained by alternating between pseudo-likelihood maximization and improved policy estimation are considered in \cite{Aguirregabiria_2002}. 
In general, the computational burden for all these methods is significant when the state space is either discrete with large cardinality, or they are continuous in high dimensions. Discretization may be avoided by using forward Monte Carlo simulations \cite{Bajari_2007, Reich} but this also becomes computationally demanding in high dimensions. A constrained optimization approach for maximum likelihood estimation of dynamic discrete choice models is considered in \cite{Su_2012}. However, the number of constraints needed to represent Bellman's equation becomes a significant computational burden with discrete state space with large cardinality or continuous state space in high dimensions.

{ Recent work has addressed the computational challenges posed by high-dimensional state space. For example, in \cite{adusumilli2022temporaldifference} the authors extend the CCP estimator approach proposed in \cite{Hotz} by considering a functional approximation approach coupled with a temporal difference (TD) algorithm to maximize pseudo-likelihood. In \cite{Chernozhukov} the authors consider an approach to adjust the CCP estimator to account for finite simple bias in high-dimensional settings. }

The literature in IRL features the seminal work \cite{ziebart2008maximum} in which a model for the expert's behavior is formulated as the policy that maximizes entropy subject to a constraint requiring that the expected features under such policy match the empirical averages in the expert's observation dataset.\footnote{ In section \ref{Sec:Duality}, we show that if the reward is linearly parametrized, the maximum entropy formulation in \cite{ziebart2008maximum} is the dual of the maximum likelihood formulation of the estimation problem.} 
The algorithms developed for maximum entropy estimation \cite{ziebart2008maximum,ziebart2010modeling,wulfmeier2015maximum} have a nested loop structure, alternating between an outer loop with a reward update step, and an inner loop that calculates the explicit policy estimates.
The computational burden of this nested structure is manageable in tabular environments, but it becomes significant in high dimensional settings requiring value function approximation.

Recent works have developed algorithms to alleviate the computational burden of nested-loop estimation methods. For example, in \cite{garg2021iq}, the authors propose to transform the standard formulation of IRL into a single-level problem by estimating the Q-function rather than estimating the reward function and associated optimal policy separately. However, the implicit reward function in the Q-function identified is a poor estimate since it is not guaranteed to satisfy Bellman's equation. Finally, \cite{ni2020f} considers an approach called $f$-IRL for estimating rewards based on the minimization of several measures of divergence with respect to the expert's state visitation measure. The approach is limited to estimating rewards that only depend on state. While the results reported are based upon a single-loop implementation, the paper does not provide a convergence guarantee to support performance. 

{ In contrast to the lines of works surveyed above, we focus our efforts in developing estimation algorithms with finite-time guarantees for computing high-quality estimators}. { Towards addressing this challenge}, in this paper, we propose a class of new algorithms which only require a finite number of computational steps for a class of (non-linearly parameterized) structural estimation problems { assuming the environment dynamics are known (or samples from the environment dynamics are available ``on-line")}. Specifically,  the proposed algorithm has a single-loop structure wherein a single-step update of the estimated policy is followed by an update of the reward parameter. 
We show that the algorithm has strong theoretical guarantees: to achieve certain $\epsilon$-approximate stationary solution for a non-linearly parameterized problem, it requires  $\mathcal{O}(\epsilon^{-2})$ steps of policy and reward updates each. 
To our knowledge, it is the first algorithm which has finite-time guarantee for the structural estimation of an MDP under nonlinear parameterization of the reward function. We conduct extensive experiments to demonstrate that the proposed algorithm outperforms many state-of-the-art IRL algorithms in {\it both} policy estimation and reward recovery.  
In particular, when transferring to a new environment, the performance of state-of-the-art reinforcement learning (RL) algorithms, using estimated rewards outperform those that use rewards recovered from existing IRL and imitation learning benchmarks.

{Finally, we consider the extension to the ``off-line" case in which the estimation task also includes the environment dynamics. Referring to our recent work (see \cite{offline-irl23}), we consider a two-stage estimation approach. First, a maximum likelihood model of dynamics is identified.
However, this first stage estimator of the environment dynamics may be inaccurate due to limited data coverage.
Thus, in the second stage, a ``conservative" reward estimator is obtained by introducing a penalty for model uncertainty.
} 

The structure of this paper is as follows. In Sec. \ref{Sec:Background}, we introduce the basic setting for structural estimation of MDPs. In Sec. \ref{Sec:MLE}, we introduce the problem formulation of the maximum likelihood  IRL. In Sec. \ref{Sec:Algorithm}, we introduce a single-loop algorithm for estimation and formalize a finite-time performance guarantee for high dimensional states. In Sec. \ref{Sec:Analysis}, we present the convergence results of our proposed single-loop algorithm. In Sec. \ref{Sec:Duality}, we consider the case with linearly parameterized rewards to show the proposed algorithm converges sublinearly to the maximum likelihood estimator. This results is proven by establishing a duality relationship between maximum entropy IRL and maximum likelihood IRL. In Sec. \ref{Sec: State-only}, we consider the case in which the agent's preferences can be represented by a reward that is only a function of the state. {In Sec. \ref{sec:offline} we outline the extension of the proposed algorithms to the offline case.} Finally, in Sec. \ref{Sec:Testbed}, we present the numerical results. 

\section{Background}
\label{Sec:Background}

\subsection{Dynamic Discrete Choice Model}
We now review the basic setting for dynamic discrete choice model as given for example in \cite{Rust1994}.
At time $t \geq 0$, the agent implements an action $a_{t}$ from a finite (discrete) action space $\mathcal{A}$ and receives a reward $r(s_{t}, a_{t};\theta) + \epsilon_t(a_t)$, where $s_t\in \mathcal{S}$ is the state at time $t$, $r(s_{t}, a_{t};\theta)$ is the reward associated to the state-action pair $(s_{t}, a_{t})$ with $\theta \in \mathbb{R}^p$ a parameter and $\epsilon_t(a_t):\mathbb{R}^{|\mathcal{A}|}\to\mathbb{R}$ is a random perturbation which is observable by the agent (decision maker) but not by the modeler.

Upon implementing the action $a_{t} \in \mathcal{A}$, the state evolves according to a Markov
process with kernel $P(s_{t+1}|s_t, a_t)$. Moreover, let $\mu(\epsilon_t|s_t)$ denote the probability distribution for the random perturbation, where the probability distribution is a function of the state.

Let $\pi(\cdot\mid s_t,\epsilon_t) $ denote a randomized policy, i.e. $\pi(a_t|s_t,\epsilon_t)$ is the probability that action $a_t$ is implemented when the state is $s_t$ and the observed reward perturbation vector is $\epsilon_t$.

The agent's optimal policy is characterized by the value function:
\begin{align*}
V_{\theta}(s_0,\epsilon_0)= \max_{\pi} ~ \mathbb{E}_{s_0 \sim \rho, \tau \sim \pi} \bigg[\sum_{t = 0}^{\infty} \gamma^t \big(r(s_t,a_t; \theta)+\epsilon_t(a_t)\big)\bigg|s_0,\epsilon_0\bigg]
\end{align*}
where the expectation is taken with respect to $a_t \sim \pi(\cdot|s_t,\epsilon_t), s_{t+1} \sim P(\cdot|s_t,a_t), \epsilon_{t+1} \sim \mu(\cdot|s_{t+1})$ and $\gamma \in [0,1)$ is the discount factor. The Bellman equation is:
\begin{align*}
    V_{\theta}(s_t, \epsilon_t) &= \max_{a_t \in \mathcal{A}}\bigg[ r(s_t, a_t;\theta) + \epsilon_t(a_t) + \gamma \mathbb{E}_{s_{t+1}\sim P(\cdot|s_t, a_t), \epsilon_{t+1} \sim \mu(\cdot|s_{t+1})} \big[ V_{\theta}(s_{t+1}, \epsilon_{t+1}) \big] \bigg], \notag \\
    &=\max_{a_t \in \mathcal{A}}\big[ Q_{\theta}(s_t, a_t) + \epsilon_t(a_t) \big]
\end{align*}
where
$Q_{\theta}:\mathcal{S} \times \mathcal{A} \mapsto \mathbb{R}$ is the fixed point of the {\em soft}-Bellman operator:
\begin{align}
\Lambda_{\theta}\big(Q(s_t, a_t)\big) = r(s_t, a_t;\theta)+ \gamma \mathbb{E}_{s_{t+1}\sim P(\cdot|s_t, a_t), \epsilon_{t+1} \sim \mu(\cdot|s_{t+1})} \bigg[ \max_{a \in \mathcal{A}}\bigg( Q(s_{t+1}, a) + \epsilon_{t+1}(a) \bigg) \bigg].
\label{def:soft_Bellman_operator}
\end{align}

As the realization of the reward perturbations is not observable by the modeler, a parametrized model of the agent's behavior is a map $\pi_{\theta}(\cdot|s_t)$ which satisfies Bellman's optimality as follows:
\begin{equation}
\pi_{\theta}(a_t|s_t)
= P\Big(a_t \in \arg \max_{a \in \mathcal{A}}\big[Q_{\theta}(s_t, a) + \epsilon_t(a)\big]\Big). 
\label{DDC:optimal_policy}
\end{equation}

{Assume observations are in the form of expert state-action trajectories $\tau^{\rm E}=\{(s_t,a_t)\}_{t\geq0}$ drawn from a ground-truth (or ``expert") policy $\pi^{\rm E}$, i.e. $a_t \sim \pi^{\rm E}(\cdot|s_t)$, $s_{t+1} \sim P(\cdot|s_t,a_t)$ and $s_0 \sim \rho(\cdot)$, where $\rho(\cdot)$ denotes the initial distribution of the first state $s_0$}. The expected discounted log-likelihood of observing such trajectory under model $\pi_{\theta}$ can be written as:{\small
\begin{align*}
{\mathbb{E}_{\tau^{\rm E} \sim \pi^{\rm E}}} \bigg[  \sum_{t = 0}^{\infty} \gamma^t \log \big( P(s_{t+1}|s_t,a_t)\pi_{\theta}(a_t | s_t) \big)  \bigg] & = {\mathbb{E}_{\tau^{\rm E} \sim \pi^{\rm E}}}\bigg[  \sum_{t = 0}^{\infty} \gamma^t \log \pi_{\theta}(a_t | s_t)  \bigg] +  {\mathbb{E}_{\tau^{\rm E} \sim \pi^{\rm E}}} \bigg[\sum_{t = 0}^{\infty} \gamma^t \log P(s_{t+1}|s_t,a_t) \bigg].
\end{align*}}

Given that the term $ {\mathbb{E}_{\tau^{\rm E} \sim \pi^{\rm E}}} \bigg[\sum_{t = 0}^{\infty} \gamma^t \log P(s_{t+1}|s_t,a_t) \bigg] $ is independent of the reward parameter $\theta$, the maximum likelihood estimation problem can be formulated as follows:
\begin{subequations}\label{DDC}
    \begin{align}
	\max_{\theta} 	&~~~~{{L}(\theta):=\mathbb{E}_{\tau^{\rm E} \sim \pi^{\rm E}}} \Big[  \sum_{t = 0}^{\infty} \gamma^t \log \pi_{\theta}(a_t | s_t)  \Big]\label{DDC:objective}\\
	{\rm s.t} &~~~~ 
\pi_{\theta}(a_t|s_t)
= P\Big(a_t \in \arg \max_{a \in \mathcal{A}}\big[Q_{\theta}(s_t, a) + \epsilon_t(a)\big]\Big), \label{DDC:constraint}
\end{align}
\end{subequations}
where $Q_{\theta}$ is the fixed point of the {\em soft}-Bellman operator in \eqref{def:soft_Bellman_operator}.

In the next subsection we review the literature on the entropy-regularized RL model and then highlight the formal equivalence of entropy-regularized  IRL with the dynamic discrete choice model just introduced in \eqref{DDC}.

\subsection{Maximum Likelihood Inverse Reinforcement Learning (ML-IRL)}
\label{Sec:MLE}

A recent literature has considered MDP models with information processing costs \cite{Tishby2011,Ortega_2013,Matejka_2015,Hansen_2018}. In these papers, optimal behavior is modeled as the solution to the following problem:
\begin{equation*}    
\max_{\pi \in \Pi} ~ J_{\theta}(\pi; \rho)\triangleq \mathbb{E}_{s_0 \sim \rho, \tau^{\rm A} \sim \pi} \bigg[ \sum_{t = 0}^{\infty} \gamma^{t} \Big(r(s_t, a_t;\theta)-c(\pi(\cdot|s_t)\Big)\bigg],
\end{equation*} 
where $\rho(\cdot)$ denotes the initial distribution of the first state $s_0$, $\tau^A$ is a trajectory generate from the agent policy $\pi$ and $c(\cdot)$ is a function representing the information processing cost. A common specification is $c(\pi(\cdot|s_t))=\alpha D_{\rm KL}(\pi(\cdot|s_t) ||\pi_{0}(\cdot|s_t))$, where  $D_{\rm KL}(\pi(\cdot|s_t) ||\pi_{0}(\cdot|s_t))=\sum\limits_{a\in \mathcal{A}}\pi (a|s_t)\log \frac{\pi (a|s_t)}{\pi
_{0}(a|s_t)} $ is Kullback-Leibler divergence between $\pi(\cdot|s_t)$ and
a reference (or default) policy $\pi_0(\cdot|s_t)$ and $\alpha \geq 0$ is a scale parameter. As the objective function above can be re-scaled by $\frac{1}{\alpha}$, we can set $\alpha=1$. 
To model no prior knowledge the reference policy $\pi_0$ is the uniformly random policy, i.e. $\pi _{0}(a)=\frac{1}{|\mathcal{A}|}$ for any $a \in \mathcal{A}$. In this case,
we can further rewrite the problem
as: 
\begin{equation*}    
\max_{\pi \in \Pi} ~ J_{\theta}(\pi; \rho) \triangleq \mathbb{E}_{s_0 \sim \rho, \tau^{\rm A} \sim \pi}\bigg[ \sum_{t = 0}^{\infty} \gamma^{t} \Big(r(s_t, a_t;\theta) +
\mathcal{H}(\pi(\cdot|s_t ))
\Big)\bigg] + \frac{\log|\mathcal{A}|}{1-\gamma},
\end{equation*} 
where $
\mathcal{H}(\pi(\cdot|s_t))=-\sum\limits_{a\in A}\pi (a|s_t)\log \pi (a|s_t)$ is the entropy of $\pi(\cdot|s_t)$. This model has also been recently used in the RL literature
\cite{haarnoja2017reinforcement, haarnoja2018soft,cayci2021linear,cen2021fast} where it is commonly referred to as an {\em entropy regularized} MDP. 

Denoting the expert policy as $\pi^{\rm E}$ and assuming an entropy-regularized MDP model for behavior. and the model for dynamic behavior described, the inverse reinforcement learning (IRL) problem can be formulated as follows:
\begin{subequations}\label{eq:ML:problem}
\begin{align}
	\max_{\theta} 	&~~~~{{L}(\theta)} := \mathbb{E}_{\tau^{\rm E} \sim \pi^{\rm E}} \bigg[ \sum_{t = 0}^{\infty} \gamma^t \log \pi_{\theta}(a_t | s_t)  \bigg] \label{eq:ML} \\
	s.t. &~~~~ 
\pi_{\theta} := \arg \max_{\pi} ~ \mathbb{E}_{s_0 \sim \rho, \tau^{\rm A} \sim \pi} \bigg[ \sum_{t = 0}^{\infty} \gamma^t \bigg( r(s_t, a_t; \theta) + \mathcal{H}(\pi(\cdot | s_t)) \bigg) \bigg]. \label{def:inner_problem}
\end{align}
\end{subequations}
When the reward perturbations $\epsilon_t(a)$ follow i.i.d. Gumbel distribution with zero mean and variance $\frac{\pi^2}{6}$ for $a \in A$, the
models of behavior \eqref{DDC:constraint} and \eqref{def:inner_problem} are equivalent (see Proposition 1 in \cite{Mai_2020}). 
{Specifically, the fixed point $Q_{\theta}(s,a)$ of the Bellman operator $\Lambda_{\theta}$ in \eqref{def:soft_Bellman_operator} and the optimal policy \eqref{DDC:optimal_policy} are of the form:
\begin{subequations}
    \begin{align}
    Q_{\theta}(s,a) &:= r(s,a; \theta) + \gamma \mathbb{E}_{s^\prime \sim P(\cdot| s,a)} \big[ V_{\theta}(s^\prime) \big]. \label{optimal_soft_Q} \\
V_{\theta}(s) &=  \log \bigg( \sum_{\tilde{a} \sim \mathcal{A}} \exp Q_{\theta}(s,\tilde{a}) \bigg) , \label{def:soft_V_closed_form} \\
\pi_{\theta}(a|s) & = \frac{\exp \big( Q_{\theta}(s,a) \big)}{\sum_{\tilde{a} \in \mathcal{A}} \exp \big( Q_{\theta}(s,\tilde{a}) \big)}.\label{def:policy_closed_form}
\end{align}
\end{subequations}
As has been shown in \cite{haarnoja2018soft,cen2021fast}, the policy described in \eqref{def:policy_closed_form} corresponds to the optimal policy in \eqref{def:inner_problem} so that
\begin{subequations}
    \begin{align}
    V_{\theta}(s) &= \max_{\pi} \mathbb{E}_{s_0\sim \rho,\tau^{\rm A} \sim \pi} \bigg[ \sum_{t = 0}^{\infty} \gamma^t \bigg( r(s_t, a_t;\theta) + \mathcal{H}(\pi(\cdot | s_t)) \bigg) \bigg | s_0 = s \bigg].\label{def:optimal_soft_V_func} 
\end{align}
\end{subequations}

}


{
\subsection{Computational Effort and Estimation Quality of Existing Algorithms}

The existing solution and approximation methodologies for solving 
\eqref{DDC} (or equivalently, \eqref{eq:ML:problem}) 
are ill-equipped for dealing with the high dimensional state space. 
For example, the algorithms considered in (e.g. \cite{Rust1994,ziebart2010modeling,wulfmeier2015maximum}) rely on a nested-loop structure which requires the solution of a fixed-point problem in the inner-loop before making any updates to the parameter estimates for the outer-loop. Evidently, in high dimensional environments. the inner-loop solution renders the nested-loop structure computationally intractable. 

Similarly, with high dimensional continuous state, a discretization approach (e.g. \cite{Su_2012}) to solving the inner problem \eqref{DDC:constraint} (or equivalently, \eqref{def:inner_problem}) is computationally intractable. Forward Monte Carlo simulations \cite{Bajari_2007, Reich} is an alternative to discretization but this is also computationally demanding in high dimensions.
 
Approximation algorithms (e.g. \cite{Hotz,Hotz1994, garg2021iq, ni2020f}) reduce the computational burden of the nested-loop structure. However, the resulting estimates may be of poor quality. For example, the CCP estimator from \cite{Hotz,Hotz1994}
may suffer from finite sample bias because in high dimensional state space, initial policy estimates (i.e. conditional choice probabilities) based upon empirical frequencies are likely of poor quality. Sequential estimators \cite{Aguirregabiria_2002} reduce bias at the expense of significant computational burden. The reward estimates in \cite{garg2021iq} do not approximate a solution to the inner problem and are thus likely to be of poor quality. Recently, \cite{adusumilli2022temporaldifference} and  \cite{Chernozhukov} have proposed approaches to account for finite-sample bias in CCP estimators in high dimensional environments.

In the present paper we introduce a new class of single-loop algorithms which exhibit finite-time guarantees of performance for solving  \eqref{DDC}) (or more precisely, its approximated version to be introduced in the next section).

As many papers in the 
dynamic discrete choice (DDC) estimation literature (e.g. \cite{Rust1994} rely on a two-stage approach to estimating dynamics (first stage) and rewards (second stage), the results obtained in this paper address the computational complexity of the {\em second} stage estimation task. This issue was ignored in \cite{Rust} due the scale of the problem. However, computational complexity is an important concern in high-dimensional environments. 

In Section \ref{sec:offline}, we also discuss how to extend the proposed method to the two-stage problem / offline setting where estimating dynamics should be considered.
}

\section{Problem Approximation in High Dimensional State Space}
{
In practice, the IRL problem \eqref{eq:ML:problem} (and its equivalent \eqref{DDC}) can only be approximated with a {\em finite} set of observed trajectories as the ground-truth behavior model (or ``expert" policy) $\pi^{\rm E}$ is not known. Let $\mathcal{D} := \{ \tau^{\rm E} \}$ denote a finite dataset of state-action trajectories independently drawn from the expert policy and the environment dynamics.  
Let $\tau^{\rm E}\sim \mathcal{D}$ denote a uniformly sampled trajectory from $\mathcal{D}$. Using a finite dataset, a natural choice for an {\em empirical} approximation to the estimation problem is the following:
\begin{subequations}\label{ML:app}
    \begin{align}
	\max_{\theta} 	&~~~~ \widetilde{L}(\theta; \mathcal{D}):=
 \mathbb{E}_{\tau^{\rm E} \sim \mathcal{D}} \bigg[ \sum_{t = 0}^{\infty} \gamma^t \log \pi_{\theta}(a_t | s_t)  \bigg] 
 \label{ML:app:objective}\\
	{\rm s.t} &~~~~ 
\pi_{\theta}(a_t|s_t)
:= \arg \max_{\pi} ~ \mathbb{E}_{s_0 \sim \rho, \tau^{\rm A} \sim \pi} \bigg[ \sum_{t = 0}^{\infty} \gamma^t \bigg( r(s_t, a_t; \theta) + \mathcal{H}(\pi(\cdot | s_t)) \bigg) \bigg].  \label{ML:app:constraint}
\end{align}
\end{subequations}
However, with high-dimensional state space, the above approximation $\widetilde{L}(\theta,\mathcal{D})$ is likely to incur significant error because the observed transitions in the data may not adequately describe the ground-truth transition kernel. In what follows, we introduce a different {\em surrogate} empirical objective $\widehat{L}(\theta,\mathcal{D})$ which provides a better approximation to the original likelihood function $L(\theta)$ given in \eqref{eq:ML:problem} with high-dimensional state.

To motivate the definition of $\widehat{L}(\theta,\mathcal{D})$, let us start by expressing the likelihood function ${L}(\theta) := \mathbb{E}_{\tau^{\rm E} \sim \pi^{\rm E}} \bigg[ \sum_{t = 0}^{\infty} \gamma^t \log \pi_{\theta}(a_t | s_t)  \bigg]$ in terms of the difference in expected value:
\begin{align}
    L(\theta) &= \mathbb{E}_{\tau^{\rm E} \sim \pi^{\rm E}} \bigg[ \sum_{t = 0}^{\infty} \gamma^t \log \pi_{\theta}(a_t | s_t)  \bigg] \nonumber \\
    &\overset{(i)}{=} \mathbb{E}_{\tau^{\rm E} \sim \pi^{\rm E}} \bigg[ \sum_{t = 0}^{\infty} \gamma^t \log \Big( \frac{\exp Q_{\theta}(s_t, a_t)}{\sum_{a \in \mathcal{A}} \exp Q_{\theta}(s_t, a)} \Big) \bigg] \nonumber \\
    &\overset{(ii)}{=} \mathbb{E}_{\tau^{\rm E} \sim \pi^{\rm E}} \bigg[ \sum_{t = 0}^{\infty} \gamma^t \Big( Q_{\theta}(s_t, a_t) - V_{\theta}(s_t) \Big) \bigg] \nonumber \\
    &\overset{(iii)}{=} \mathbb{E}_{\tau^{\rm E} \sim \pi^{\rm E}} \bigg[ \sum_{t = 0}^{\infty} \gamma^t \bigg( r(s_t, a_t; \theta) + { \gamma} \mathbb{E}_{s_{t+1} \sim P(\cdot| s_t, a_t)} [ V_{\theta}(s_{t+1}) ] - V_{\theta}(s_t) \bigg) \bigg] \nonumber \\
    &= \mathbb{E}_{\tau^{\rm E} \sim \pi^{\rm E}} \bigg[ \sum_{t = 0}^{\infty} \gamma^t r(s_t, a_t; \theta) \bigg] + \sum_{t = 0}^{\infty} \gamma^{t+1} \mathbb{E}_{(s_t, a_t) \sim \tau^{\rm E}} \bigg[ \mathbb{E}_{s_{t+1} \sim P(\cdot| s_t, a_t)} [ V_{\theta}(s_{t+1}) ] \bigg] - \mathbb{E}_{\tau^{\rm E} \sim \pi^{\rm E}} \bigg[ \sum_{t = 0}^{\infty} \gamma^t V_{\theta}(s_t)  \bigg] \nonumber \\
    & =\mathbb{E}_{\tau^{\rm E} \sim \pi^{\rm E}} \bigg[ \sum_{t = 0}^{\infty} \gamma^t r(s_t, a_t; \theta) \bigg] - \mathbb{E}_{s_0 \sim \rho(\cdot)} \Big[ V_{\theta}(s_0) \Big]  \label{rewrite:maximum_likelihood}
\end{align}
where $(i)$ follows the closed-form expression of the optimal policy $\pi_{\theta}$ in \eqref{def:policy_closed_form}, $(ii)$ follows the expression of the soft value function $V_{\theta}$ in \eqref{def:soft_V_closed_form} and $(iii)$ follows from the fixed point definition in \eqref{optimal_soft_Q}. 

Observe that in the above decomposition, the first term is related to the expert policy $\pi^{\rm E}$, while the second term is related to the initial distribution $\rho$ and the transition kernel $P$. Note that in practice we only have limited observations of expert trajectories from a fixed dataset $\mathcal{D}$, but cannot directly sample the trajectory from the expert policy $\pi^{\rm E}$ in an online manner. Hence, we need to construct an estimation problem which utilizes limited observations of expert trajectories to approximate the original maximum likelihood objective in \eqref{rewrite:maximum_likelihood}. Since we assume the {\it online} setting in which the transition kernel $P$ and the initial distribution $\rho$ are either available for access or known, we can construct a {\it surrogate} approximation to the likelihood as follows:
\begin{align}
\widehat{L}(\theta; \mathcal{D})&:=
 \mathbb{E}_{\tau \sim \mathcal{D}} \bigg[ \sum_{t = 0}^{\infty} \gamma^t r(s_t, a_t; \theta) \bigg] - \mathbb{E}_{s_0 \sim \rho} \Big[ V_{\theta}(s_0) \Big]. 
 \label{ML:estimation:objective}
 \end{align}



In contrast, if we conduct the same analysis on the {\it empirical} approximation $\widetilde{L}(\theta;\mathcal{D})$ presented in \eqref{ML:app}, we obtain:
\begin{align}
    \widetilde{L}(\theta; \mathcal{D}) &= \mathbb{E}_{\tau^{\rm E} \sim \mathcal{D}} \bigg[ \sum_{t = 0}^{\infty} \gamma^t \log \pi_{\theta}(a_t | s_t)  \bigg] \nonumber \\
    &\overset{(i)}{=} \mathbb{E}_{\tau^{\rm E} \sim \mathcal{D}} \bigg[ \sum_{t = 0}^{\infty} \gamma^t \log \Big( \frac{\exp Q_{\theta}(s_t, a_t)}{\sum_{a \in \mathcal{A}} \exp Q_{\theta}(s_t, a)} \Big) \bigg] \nonumber \\
    &\overset{(ii)}{=} \mathbb{E}_{\tau^{\rm E} \sim \mathcal{D}} \bigg[ \sum_{t = 0}^{\infty} \gamma^t \Big( Q_{\theta}(s_t, a_t) - V_{\theta}(s_t) \Big) \bigg] \nonumber \\
    &\overset{(iii)}{=} \mathbb{E}_{\tau \sim \mathcal{D}} \bigg[ \sum_{t = 0}^{\infty} \gamma^t \bigg( r(s_t, a_t; \theta) + { \gamma} \mathbb{E}_{s_{t+1} \sim P(\cdot| s_t, a_t)} [ V_{\theta}(s_{t+1}) ] - V_{\theta}(s_t) \bigg) \bigg] \nonumber \\
    &= \mathbb{E}_{\tau \sim \mathcal{D}} \bigg[ \sum_{t = 0}^{\infty} \gamma^t r(s_t, a_t; \theta) \bigg] + \sum_{t = 0}^{\infty} \gamma^{t+1} \mathbb{E}_{(s_t, a_t) \sim \mathcal{D}} \bigg[ \mathbb{E}_{s_{t+1} \sim P(\cdot| s_t, a_t)} [ V_{\theta}(s_{t+1}) ] \bigg] - \mathbb{E}_{\tau \sim \mathcal{D}} \bigg[ \sum_{t = 0}^{\infty} \gamma^t V_{\theta}(s_t)  \bigg] \nonumber \\
    &= \underbrace{\bigg( \mathbb{E}_{\tau \sim \mathcal{D}} \bigg[ \sum_{t = 0}^{\infty} \gamma^t r(s_t, a_t; \theta) \bigg] - \mathbb{E}_{s_0 \sim \mathcal{D}} \bigg[ V_{\theta}(s_0) \bigg] \bigg)}_{\rm T1: ~\text{\footnotesize surrogate likelihood}} \nonumber \\
    & \quad + \underbrace{\bigg( \sum_{t = 0}^{\infty} \gamma^{t+1} \mathbb{E}_{(s_t, a_t) \sim \mathcal{D}, s_{t+1} \sim P(\cdot| s_t, a_t)} \big[ V_{\theta}(s_{t+1}) \big] - \sum_{t = 0}^{\infty} \gamma^{t+1} \mathbb{E}_{(s_t, a_t, s_{t+1}) \sim \mathcal{D}} \big[ V_{\theta}(s_{t+1}) \big] \bigg)}_{\rm T2:~\text{\footnotesize error term due to transition probability mismatch}} \label{eq:empirical_likelihood_expression}
\end{align}
where the second term is the error introduced by approximating the transition using  finite data. 

From the above analysis, we argue that the {\it surrogate} approximation $\widehat{L}(\theta;\mathcal{D})$ in \eqref{ML:estimation:objective} is a more accurate objective function as compared to the {\it empirical} likelihood $\widetilde{L}(\theta;\mathcal{D})$ in  \eqref{ML:app:objective}.
Below we show under a mild assumption, $\widehat{L}(\theta;\mathcal{D})$ can well-approximate $L(\theta)$ when data is large enough.
\begin{assumption} \label{assumption:bound_reward}
    For any reward parameter $\theta$, the following condition holds:
        \begin{align}
          0 \leq r(s,a;\theta) \leq C_r, \quad \forall s \in \mathcal{S}, a \in \mathcal{A}\label{ineq:bound_reward} 
        \end{align}
        where $C_r>0$ is a fixed constant.
\end{assumption}


\begin{lemma} \label{lemma:objective_concentration}
    Suppose Assumption \ref{assumption:bound_reward} hold. Consider the likelihood function $L(\theta)$ in \eqref{eq:ML} and its surrogate empirical version $\widehat{L}(\theta; \mathcal{D})$ defined in  \eqref{ML:estimation:objective}. Then, with probability greater than $ 1 - \delta$, we have:
    \begin{align}
        |L(\theta) - \widehat{L}(\theta; \mathcal{D})| \leq \frac{C_r}{1 - \gamma} \sqrt{\frac{\ln(2 / \delta)}{2|\mathcal{D}|}}. \label{eq:approximation_error:concentration}
    \end{align}
\end{lemma}
The proof of Lemma \ref{lemma:objective_concentration} can be found in Section \ref{proof:surrogate_objective_approximation} in the electronic companion. 

In the rest of this work, we will consider the following surrogate estimation problem: 
\begin{subequations}\label{ML:estimation}
    \begin{align}
	\max_{\theta} 	&~~~~ \widehat{L}(\theta; \mathcal{D}) :=
 \mathbb{E}_{\tau \sim \mathcal{D}} \bigg[ \sum_{t = 0}^{\infty} \gamma^t r(s_t, a_t; \theta) \bigg] - \mathbb{E}_{s_0 \sim \rho} \Big[ V_{\theta}(s_0) \Big]\label{ML:estimation:surrogate} \\
	{\rm s.t} &~~~~ 
\pi_{\theta}(a_t|s_t)
:= \arg \max_{\pi} ~ \mathbb{E}_{s_0 \sim \rho, \tau^{\rm A} \sim \pi} \bigg[ \sum_{t = 0}^{\infty} \gamma^t \bigg( r(s_t, a_t; \theta) + \mathcal{H}(\pi(\cdot | s_t)) \bigg) \bigg].  \label{ML:estimation:constraint}
\end{align}
\end{subequations}
}

\section{The Proposed Algorithm} 
\label{Sec:Algorithm}

The main idea in the proposed algorithm is to alternate between one step of policy update to improve the solution of the lower-level problem, and one step of the parameter update which improves the upper-level likelihood objective. At each iteration $k$, given the current policy $\pi_k$ and the reward parameter $\theta_k$, a new policy $\pi_{k+1}$ is generated from the policy improvement step, and $\theta_{k+1}$ is generated by the reward optimization step. 

{In Sec. \ref{Sec:Algorithm} -  Sec. \ref{Sec:Analysis}, we will design an algorithm to solve the approximated maximum likelihood problem \eqref{ML:estimation}. We emphasize that in Sec. \ref{Sec:Algorithm} - Sec. \ref{Sec: State-only}, we assume an {\it online} setting where the learner knows the  transition kernel $P(s_{t+1}|s_t,a_t)$ or can sample from it. The motivation is that understanding how to develop efficient algorithms for the {\it online} setting is the basis for addressing the more challenging {\it offline} setting. In Sec. \ref{sec:offline} we will briefly outline how to extend this work to the offline setting.}
Below we present the details of our algorithm at a given iteration $k$.




{\bf Policy Improvement Step.} Let us consider optimizing the lower-level problem \eqref{def:inner_problem}, when the reward parameter $\theta_k$ is held fixed. 
Towards this end, we define the so-called soft Q-function and soft value functions under a given policy-reward pair $(\pi_k,\theta_k)$:
\begin{subequations}
    \begin{align}
      V_{k}(s) &= \mathbb{E}_{\tau^{\rm A} \sim \pi_k} \bigg[ \sum_{t = 0}^{\infty} \gamma^t \bigg( r(s_t, a_t; \theta_k) + \mathcal{H}(\pi_k(\cdot | s_t)) \bigg) \bigg| s_0 = s \bigg], \label{def:soft_Value_function} \\ 
      Q_{k}(s,a) &= r(s,a; \theta_k) + \gamma \ee_{s^\prime \sim P(\cdot | s,a)} \big[ V_{k}(s^\prime) \big]. \label{def:soft_Q_function}
    \end{align}
\end{subequations}
Similarly, if the policy is {\it optimal} for a given parameter $\theta$ (as defined in \eqref{def:inner_problem}), then we will denote the associated soft Q-function and soft value function as $Q_{\theta}$ and $V_{\theta}$.

To obtain an estimate of the policy at iteration $k$, let us suppose that we have access to an estimate of the soft Q-function, denoted as $\widehat{Q}_k(s,a)$, which satisfies $\| \widehat{Q}_k - Q_k \|_{\infty} \leq \epsilon_{\rm app}$, with $\epsilon_{\rm app}>0$
being the approximation error. 
Then the estimated policy will be generated according to 
\begin{align}
    \pi_{k+1}(a|s) \propto \exp\big( \widehat{Q}_k(s,a)\big), \quad \forall s \in \mathcal{S}, a \in \mathcal{A}. \label{def:approximated_SPI}
\end{align}
When $\epsilon_{\rm app}=0$, or equivalently when $\widehat{Q}_k(s,a)={Q}_k(s,a),\;\forall~s \in \mathcal{S}, a \in \mathcal{A}$, and when $r(\cdot, \cdot;\theta_k)$ is fixed, the above update is referred to as the {\it soft policy iteration};
{it is known that the policy will be monotonically improved by soft policy iteration  
and will converge linearly to the optimal policy; see \cite[Theorem 1]{cen2021fast}.}
{In practice, when we do not have direct access to the exact soft Q-function $Q_k$, one could use an {\it estimated} soft Q-function $\widehat{Q}_k$ to perform the approximated soft policy iteration in \eqref{def:approximated_SPI}, which can be obtained by following the update schemes in soft Q-learning \cite{haarnoja2017reinforcement} or soft Actor-Critic (SAC) \cite{haarnoja2018soft}.} 

{\bf Reward Optimization Step.}
We propose to use a stochastic gradient-type algorithm to optimize the reward parameter $\theta$. Towards this end, let us first derive the exact gradient $\nabla_{\theta} L(\theta)$. See the supplementary material for detailed proof. 
\begin{lemma} \label{lemma:outer_gradient}
	The gradient of the { $L(\theta)$ and $ \widehat{L}(\theta; \mathcal{D})$}, as defined in \eqref{eq:ML} and \eqref{ML:estimation:objective} respectively, can be expressed as:
 \begin{subequations}
     \begin{align}
 \nabla_{\theta}  {{L}}
  (\theta) & = 
  {\mathbb{E}_{\tau^{\rm E} \sim \pi^{\rm E}}}\bigg[ \sum_{t = 0}^{\infty} \gamma^t 
		\nabla_{\theta} r(s_t, a_t;\theta)
		\bigg] - \mathbb{E}_{\tau^{\rm A} \sim \pi_{\theta}}\bigg[ \sum_{t = 0}^{\infty} \gamma^t 
		\nabla_{\theta}r(s_t, a_t;\theta)
		\bigg], \label{eq:likelihood_grad}\\
		{\nabla_{\theta}  \widehat{L}
  (\theta; \mathcal{D})} & = 
  {\mathbb{E}_{\tau^{\rm E} \sim \mathcal{D}}}\bigg[ \sum_{t = 0}^{\infty} \gamma^t 
		\nabla_{\theta} r(s_t, a_t;\theta)
		\bigg] - \mathbb{E}_{\tau^{\rm A} \sim \pi_{\theta}}\bigg[ \sum_{t = 0}^{\infty} \gamma^t 
		\nabla_{\theta}r(s_t, a_t;\theta)
		\bigg]. \label{eq:likelihood_grad_app}
	\end{align}
 \end{subequations}
\end{lemma}

{We note that the gradient expression \eqref{eq:likelihood_grad}  takes the same form as the one given in a recent work \cite[Eq. (1)]{sanghvi2021inverse}. However, our proof which focuses on the {\em infinite} horizon case is different. Moreover, we further derive the gradient expression of the sample-based estimation problem $\widehat{L}(\theta; \mathcal{D})$ which has not been considered in \cite{sanghvi2021inverse}. }

In order to obtain stochastic estimators of the empirical gradient $\nabla_{\theta} \widehat{L}(\theta_k; \mathcal{D})$, we take { two} approximation steps: 1) approximate the optimal policy $\pi_{\theta_k}$ by $\pi_{k+1}$ in \eqref{def:approximated_SPI}, since the optimal policy $\pi_{\theta_k}$ is not available throughout the algorithm; 
{ 2) sample the trajectory $\tau^{\rm A}$ from the current policy $\pi_{k+1}$.}

Following the approximation steps mentioned above, we construct a stochastic estimator $g_k$ to approximate the {empirical gradient $\nabla_{\theta} \widehat{L}(\theta_k; \mathcal{D})$ in \eqref{eq:likelihood_grad_app}} as follows:
\begin{align}
    g_k := h(\theta_{k}; \tau_k^{\rm E}) - h(\theta_{k}; \tau_k^{\rm A}), \;\; \mbox{\rm where}\;\;   h(\theta; \tau) := \sum_{t = 0}^{\infty} \gamma^t \nabla_{\theta} r(s_t, a_t; \theta)  \label{eq:stochastic_grad_estimator}.
\end{align}
With the stochastic gradient estimator $g_k$, the reward parameter $\theta_k$ is updated as:
\begin{align}
    \theta_{k+1} = \theta_k + \alpha g_k \label{eq:reward_parameter_update}
\end{align}
where $\alpha$ is the stepsize in updating the reward parameter. 

Alg. \ref{alg:IRL} summarizes the proposed two-step approach for solving the IRL problem \ref{eq:ML:problem}. {It is worth mentioning that the proposed algorithm can also be used to solve the DDC problem \eqref{DDC} due to the equivalence between \eqref{DDC} and \eqref{eq:ML:problem}.}
\begin{algorithm}[tbh]
	\caption{Maximum Likelihood Inverse Reinforcement Learning (ML-IRL)
}%
	\begin{algorithmic}	
		\State {\bfseries Input:} Initialize reward parameter $ \theta_0 $ and policy $ \pi_0 $. Set the reward parameter's stepsize as $\alpha$.
		\For{$k=0,1,\ldots, K-1$}
		\State \textbf{{Policy Evaluation:}} Approximate the soft Q-function 
		{$Q_{k}(\cdot,\cdot)$}
		by $ \widehat{Q}_{k}(\cdot, \cdot)$.
		\State \textbf{Policy Improvement:} $ \pi_{k+1}( a | s) \propto \exp \big( \widehat{Q}_{k}(s, \cdot) \big), ~ \forall s \in \mathcal{S}, a \in \mathcal{A} $. {  (Lower-Level Update)} 
		\State \textbf{Data Sampling I:} Sample a trajectory $\tau_k^{\rm E}$
  {from the dataset $\mathcal{D}$}.
		\State \textbf{Data Sample II:} Sample a trajectory $\tau_k^{\rm A} := \{ s_t, a_t \}_{t \geq 0}$ from the current policy $\pi_{k+1}$
		\State \textbf{Estimating Gradient:} $ g_k := h(\theta_{k}, \tau_k^{\rm E}) - h(\theta_{k}, \tau_k^{\rm A}) $ where $h(\theta, \tau) := \sum_{t = 0}^{\infty} \gamma^t { \nabla_{\theta} r(s_t, a_t; \theta)}$
		\State \textbf{Reward Parameter Update:} $ \theta_{k+1} := \theta_{k} + \alpha g_k $  { (Upper-Level Update)}
		\EndFor
	\end{algorithmic}
	\label{alg:IRL}
\end{algorithm}

Before closing this section, let us note that the generic alternating update strategy adopted by our algorithm is efficient, since completely solving the policy optimization subproblem all the time could be redundant, and could induce heavy computation burden. 
Such a kind of strategy has been used in many other RL related settings as well. For example, the well-known actor-critic (AC) algorithm for policy optimization \cite{konda1999actor,wu2020finite,hong2020two} alternates between one step of policy  update, and one step of critic parameter update. However, these types of algorithm are known to be challenging to analyze, partly because when the inner problem (e.g., the policy optimization problem \eqref{def:inner_problem}) is not solved exactly, the update direction for the main parameter (e.g., $\theta$ in \eqref{eq:ML:problem}) can be very far from the desired descent directions. That is, $g_k$ in \eqref{eq:stochastic_grad_estimator} can be a very coarse approximation of the exact gradient {$\nabla_{\theta} \widehat{L}(\theta_k; \mathcal{D})$ as expressed in \eqref{eq:likelihood_grad_app}}. 
In the subsequent sections, we develop techniques to address the above mentioned changes.

\section{Theoretical Analysis}
\label{Sec:Analysis}


Our analysis is based upon the so-called {\it two-timescale} stochastic approximation (TTSA) approach \cite{borkar1997stochastic,hong2020two}, where the lower-level problem updates in a faster time-scale (i.e., converges faster) compared with its upper-level counterpart.  
Intuitively, the TTSA enables the $\pi_{k+1}$ to track the optimal $\pi_{\theta_k}$, so that the gradient estimate $g_k$ will stay close to the gradient {$\nabla_{\theta} \widehat{L}(\theta_k)$}. 
Indeed, Alg. \ref{alg:IRL} has the desired two-timescale phenomenon because the policy update \eqref{def:approximated_SPI} converges linearly to the optimal policy under a fixed reward function  \cite[Theorem 2]{cen2021fast} (hence it is `fast'); while the reward parameter update \eqref{eq:reward_parameter_update} does not have such linear convergence property (hence it is  `slow').
To begin our analysis, let us first present a few technical assumptions. 

\begin{assumption}[Ergodic Dynamics]
	For any policy $\pi$, assume the Markov chain with transition kernel $\mathcal{P}$ is irreducible and aperiodic under policy $\pi$. Then there exist constants $ \kappa > 0 $ and $ \rho \in (0, 1) $ such that 
	\begin{align}
		\sup_{s \in \mathcal{S}} ~ \| P(s_t \in \cdot| s_0 = s, \pi) -  \mu_{\pi}(\cdot) \|_{TV} \leq \kappa \rho^t, \quad \forall ~ t \geq 0  \nonumber
	\end{align}
	where $ \| \cdot \|_{TV} $ is the total variation (TV) norm; $ \mu_{\pi} $ is the stationary state distribution under $ \pi $.
\label{Assumption:Ergodicity_Markov_chain}
\end{assumption}

Assumption \ref{Assumption:Ergodicity_Markov_chain} assumes the Markov chain mixes at a geometric rate. It is a common assumption in the iterature of RL \cite{bhandari2018finite,zou2019finite,wu2020finite}, which
holds for any time-homogeneous Markov chain with finite-state space or any uniformly ergodic Markov chain
with general state space.

\begin{assumption}[Lipschitz Reward]
\label{Assumption:reward_grad_bound}
	For any $s \in \mathcal{S}$, $a \in \mathcal{A}$ and any reward parameter $\theta$, the following holds: 
		\begin{align}
		\big \| \nabla_{\theta} r(s, a; \theta) \big  \|  \leq L_r, \quad
		\big \|	\nabla_{\theta} r(s, a; \theta_1)  - \nabla_{\theta} r(s, a; \theta_2)	\big \|	  \leq  L_g \|  \theta_{1} - \theta_{2}	\|  \label{ineq:Lipschitz_smooth_reward}
		\end{align}
	where $L_r$ and $L_g$ are positive constants.
\end{assumption}

Assumption \ref{Assumption:reward_grad_bound} assumes that the parameterized reward function has bounded gradient and is Lipschitz smooth. Such assumption in Lipschitz property are common in the literature of min-max / bi-level optimization \cite{jin2020local,hong2020two,guan2021will,khanduri2021near,chen2021closing}. Based on Assumptions \ref{assumption:bound_reward} - \ref{Assumption:reward_grad_bound}, we next provide the following Lipschitz properties:
\begin{lemma} \label{lemma:Lipschitz_properties}
	Suppose Assumptions \ref{assumption:bound_reward} - \ref{Assumption:reward_grad_bound} hold. There are positive constant $L_q$ and $L_c$ such that the following results hold for any reward parameter $\theta_1$ and $\theta_2$:
	\begin{subequations}
	    	\begin{align}
	    | Q_{\theta_1}(s,a)  - Q_{\theta_2}(s,a) |& \leq L_q \| \theta_{1} - \theta_2 \|, \quad \forall s \in \mathcal{S}, a \in \mathcal{A}  \label{ineq:soft_Q_Lipschitz} \\
		\| \nabla_{\theta} { \widehat{L}}(\theta_{1}; \mathcal{D}) - \nabla_{\theta} { \widehat{L}}(\theta_{2}; \mathcal{D}) \|& \leq L_c \| \theta_1 - \theta_2 \|  \label{ineq:objective_lipschitz_smooth}
	\end{align}
	\end{subequations}
	where $Q_{\theta}(\cdot, \cdot)$ denotes the soft Q-function under the reward parameter $\theta$ and the optimal policy $\pi_{\theta}$. 
\end{lemma}

{The full proof of the result is delegated to  Sec. \ref{sub:lip} in the companion of the paper.}

Next we present the main results, which show the convergence speed of the policy $\{ \pi_k \}_{k \geq 0}$ and the reward parameter $\{ \theta_k \}_{k \geq 0}$ in the Alg. \ref{alg:IRL}. Please see Appendix \ref{proof:main_convergence_theorem} for the detailed proof.
\begin{theorem}
\label{theorem:main_convergence_results}
	Suppose Assumptions 1 - 2 hold. Let  $K$ denote the total number of iterations to be run by the algorithm. Let us select $\alpha := \frac{\alpha_0}{K^\sigma} $ for the reward update step  \eqref{eq:reward_parameter_update}, where $\alpha_0>0$ and $ \sigma\in(0,1) $ are some fixed constants. Then the follow holds:
	\begin{subequations}
	    \begin{align}
		&\frac{1}{K}\sum_{k = 0}^{K-1} \mathbb{E} \left[ \big \|  \log\pi_{k+1} - \log\pi_{\theta_k} \big \|_{\infty} \right] = \mathcal{O}(K^{-1}) + \mathcal{O}(K^{-\sigma}) + \mathcal{O}(\epsilon_{\rm app}) \label{rate:lower_error} \\
		&\frac{1}{K} \sum_{k = 0}^{K - 1} \mathbb{E} \left[ \|  \nabla_{\theta} { \widehat{L}}(\theta_{k}; \mathcal{D})  \|^2 \right] = \mathcal{O}(K^{-\sigma}) + \mathcal{O}(K^{-1 + \sigma}) + \mathcal{O}(K^{-1}) + \mathcal{O}(\epsilon_{\rm app}) \label{rate:upper_grad_norm}
	\end{align}
	\end{subequations}
	where $\|  \log\pi_{k+1} - \log\pi_{\theta_k} \|_{\infty} := \max_{s \in \mathcal{S}, a \in \mathcal{A}} \big| \log\pi_{k+1}(a|s) - \log\pi_{\theta_k}(a|s) \big|$. In particular, setting $\sigma=1/2$, then both quantities in \eqref{rate:lower_error} and \eqref{rate:upper_grad_norm} converge with the rate $\mathcal{O}(K^{-1/2}) + \mathcal{O}(\epsilon_{\rm app}).$
\end{theorem}

In Theorem \ref{theorem:main_convergence_results}, we present the finite-time guarantee for the convergence of the Alg. \ref{alg:IRL}. 
We note that our theoretical guarantee is different from the existing works, such as \cite{cen2021fast} which showed the convergence rate of soft policy iteration under a {\it fixed} reward function. Theorem \ref{theorem:main_convergence_results} analyzes a more challenging setting where {\it both} the policy and reward parameter are kept changing. To our knowledge, this is the first result that characterizes the finite-time convergence for an algorithm developed for either the structural estimation problem \eqref{DDC} or the maximum likelihood IRL problem \eqref{eq:ML:problem}.
{In the following result, we characterize the dimension dependence of the performance of the policy estimated with Algorithm 1.}

{
{\bf Remark 3}:
It is worth mentioning here that the Lipschitz constant $L_c$ in \eqref{ineq:objective_lipschitz_smooth} is given by: $$L_c = \frac{2L_q L_r C_d \sqrt{ | \mathcal{S}| \cdot |\mathcal{A}| } }{1 - \gamma}  + \frac{2 L_g}{1 - \gamma},$$ 
where $C_d$ is a constant given in \eqref{ineq:lipschitz_measure}.
Hence, $L_c$, as well as the subsequent convergence rate of the algorithm in Theorem 1, are dependent on the dimension of the problem (i.e., the size of the state and action space). However, the empirical evidence (to be presented in Sec. \ref{Sec:Testbed}) strongly indicates that the proposed algorithm performs well with high-dimensional neural network representations. This is mainly because our formulation allows us to directly take (approximate) gradient steps on updating $\theta_k$, and that for fixed reward parameterization $\theta_k$, the lower-level policy optimization problem we are interested in has a closed-form solution (as a function of the corresponding $Q_k$).
We believe that the extension of the analysis for our algorithm with function approximations (for the parameterized Q-function and the policy) will result in bounds that have less dependence upon the dimension of the basis at the expense of additional approximation error term. The extension of our convergence analysis with function approximations is left for future research.
}

\section{The Linearly Parameterized Reward Function Case}
\label{Sec:Duality}

The result in Theorem 1 can be further strengthened when rewards are a linear function of (possibly non-linear) features, i.e. $ r(s,a;\theta) = \phi(s,a)^{\top} {\theta}$ with $\phi:\mathbb{R}^{|S|\times |A|} \rightarrow \mathbb{R}^p$ and the distribution of observations is consistent with optimal behavior for a ground truth parameter $\theta^*$, $\pi^E=\pi_{\theta^*}$.


In this setting, the result in Theorem 1 can be strengthened to finite-time convergence to the optimal solution. To show this result we first establish a duality relationship between the estimation problem in \eqref{ML:estimation},
and the maximum entropy estimator \cite{Ziebart_2013} which is the solution to the following problem:
\begin{subequations}\label{eq:max_ent}
	\begin{align} 
		\max_{\pi} &\quad -\mathbb{E}_{\tau^{\rm A} \sim \pi} \bigg[ \sum_{t = 0}^{\infty} \gamma^t \log \pi(a_t | s_t) \bigg]
		 \label{obj:max_ent} \\
		{\rm s.t.} &\quad \mathbb{E}_{\tau^{\rm A} \sim \pi} \bigg[ \sum_{t = 0}^{\infty} \gamma^t \phi(s_t, a_t) \bigg]	= 	{ \mathbb{E}_{\tau^{\rm E} \sim \mathcal{D}}} \bigg[ \sum_{t = 0}^{\infty} \gamma^t \phi(s_t, a_t) \bigg] \label{constraint:feature} \\
		&~~~~~~~~\sum_{a_t \in \mathcal{A}} \pi(a_t | s_t) = 1, \quad \forall ~ s_t \in \mathcal{S}, t \geq 0 \label{constraint:prob_sum} \\
		&~~~~~~~~{\pi(a_t | s_t) \geq 0}, \quad \forall ~ s_t \in \mathcal{S}, a_t \in \mathcal{A}, t \geq 0 \label{constraint:prob_positive}
	\end{align}
\end{subequations}
where \eqref{constraint:feature} requires that the expected discounted feature value under the model matches the { expected} discounted feature under the finite dataset $\mathcal{D}$ of collected expert trajectories. When the expert policy is known or available for access, the maximum entropy estimation problem is defined as in \eqref{eq:max_ent} by replacing \eqref{constraint:feature} with
\begin{align}
    \mathbb{E}_{\tau^{\rm A} \sim \pi} \bigg[ \sum_{t = 0}^{\infty} \gamma^t \phi(s_t, a_t) \bigg]	&= 	{ \mathbb{E}_{\tau^{\rm E} \sim \pi^{\rm E}}} \bigg[ \sum_{t = 0}^{\infty} \gamma^t \phi(s_t, a_t) \bigg]. \label{max_ent_infinite}
\end{align}
The following result formalizes the relationship between the maximum entropy estimation problem \eqref{eq:max_ent} and the estimation problem \eqref{ML:estimation}. 


Please see the detailed proof in Appendix \ref{proof:theorem:duality}.
 
 \begin{theorem} \label{theorem:duality}
	Under linear parameterization for reward function $r(s, a; \theta) = \phi(s, a)^{\top} \theta$, the estimation problem defined in \eqref{ML:estimation} ({ resp. the maximum likelihood IRL problem \eqref{eq:ML:problem}}) is the Lagrangian dual of the maximum entropy estimation problem \eqref{eq:max_ent} ({ resp. the problem defined by \eqref{obj:max_ent},\eqref{max_ent_infinite},\eqref{constraint:prob_sum},\eqref{constraint:prob_positive}}). Moreover, strong duality holds between the two problems.
\end{theorem}



\noindent
{
\begin{Corollary} \label{corollary:duality} (i) The surrogate objective defined in \eqref{ML:estimation:surrogate} (dual objective) is a {\it concave} function of $\theta$;  (ii) If the ground-truth reward values $r(s,\tilde{a};\theta^*)$ for a reference action $\tilde{a} \in A$ and $s \in S$ are known, the optimal solution to \eqref{ML:estimation} is unique.
\end{Corollary}

{\bf Proof:} The first result is a direct consequence of Theorem \ref{theorem:duality} since the estimation problem \eqref{ML:estimation} is a dual problem. Then we prove {\em (ii)} by contradiction.
Let $\widehat{\theta}_1, \widehat{\theta}_2$ denote two distinct solutions of the estimation problem \eqref{ML:estimation}, which is the dual problem w.r.t. the maximum entropy IRL problem \eqref{eq:max_ent}. 
From \eqref{eq:likelihood_grad_app} it follows that
\begin{align}
\nabla_{\theta}  \widehat{L}
  (\widehat{\theta}_i; \mathcal{D})=
{\mathbb{E}_{\tau^{\rm E} \sim \mathcal{D}}}\bigg[ \sum_{t = 0}^{\infty} \gamma^t 
		 \phi (s_t, a_t;\theta)
		\bigg] - \mathbb{E}_{\tau^{\rm A} \sim \pi_{\widehat{\theta}_i}}\bigg[ \sum_{t = 0}^{\infty} \gamma^t 
		\phi(s_t, a_t;\theta)
		\bigg]=0, \quad i=1,2.
  \label{feasible}
  \end{align}
Let $Q_{\widehat{\theta}_i}$ denote the unique fixed point of the soft-Bellman operator and $\tilde{Q}_{\widehat{\theta}_i}(s,a):=Q_{\widehat{\theta}_i}(s,a)-Q_{\widehat{\theta}_i}(s,\tilde{a})$ for all $a\in A$.
 The following mapping   (from the parameter space to the policy space)
$$\pi_{\widehat{\theta}_i}(a|s):=\frac{\exp{\tilde{Q}_{\widehat{\theta}_i}(s,a)}}{\sum_{a'\in A}\exp{\tilde{Q}_{\widehat{\theta}_i}(s,a')}}$$ is one-to-one (see Proposition 1 in \cite{Hotz}) and $\pi_{\widehat{\theta}_1}\neq \pi_{\widehat{\theta}_2}$.
By Theorem 2 (strong duality), it holds that
$$
\mathbb{E}_{\tau^{\rm A} \sim \pi_{\widehat{\theta}_i}} \big[ \sum_{t = 0}^{\infty} \gamma^t \log \pi_{\widehat{\theta}_i}(a_t | s_t) \big] 
=
\mathbb{E}_{\tau^{\rm A} \sim \widehat{\pi}} \big[ \sum_{t = 0}^{\infty} \gamma^t \log \widehat{\pi} (a_t | s_t) \big]
$$
where $\widehat{\pi}$ is an optimal solution to primal problem \eqref{eq:max_ent}. 
This is a contradiction to the uniqueness of the optimal solution $\widehat{\pi}$ since the maximum entropy objective \eqref{obj:max_ent} is {strictly concave}. Hence, we can show that the optimal solution to \eqref{ML:estimation} is unique.
\hfill $\blacksquare$


Note that the concavity property does not hold for the estimation objective in \cite{Rust1994}. 
For example, the undiscounted empirical likelihood for Group 2 data in \cite{Rust} can be shown to be non-concave.


Moreover, we note that the bi-level formulations \eqref{eq:ML:problem} and \eqref{ML:estimation} are quite involved, and it is difficult to directly show the concavity of the problems \eqref{eq:ML:problem} and \eqref{ML:estimation} with nonlinear reward parameterization. Based upon our observations under linear reward parameterization, as well as the finite sample guarantee given in Lemma \ref{lemma:objective_concentration}, we have the following corollary: 

\begin{Corollary} \label{corollary:optimality_guarantee}
	Assume that the reward is linearly parameterized, i.e. $r(s, a; \theta) = \phi(s, a)^{\top} \theta$ with $\theta \in  \Theta \subset \mathbb{R}^p$ where $\Theta$ is a  compact set. Assume the ground-truth reward value $r(s,\tilde{a};\theta^*)$ for a reference action $\tilde{a} \in A$ and $s \in S$ are known. 
 Let $\widehat{\theta}$ denote the optimal solution to \eqref{ML:estimation}. From Algorithm 1's output, define $\widehat{\theta}_K:=\theta_{k^*(K)}$ where
$$
k^*(K) := \arg\min_{k\in \{0,K\}} \{\|\nabla \widehat{L}(\theta_k,\mathcal{D})\|^2\}
$$
then $\widehat{\theta}_K \rightarrow \widehat{\theta}$ in probability with finite-time guarantee $\mathbb{E}\big[\|\nabla \widehat{L}(\widehat{\theta}_K,\mathcal{D})\|^2\big]
\leq
\mathcal{O}(K^{-1/2})$.
Furthermore, if
    $|\mathcal{D}|\geq \frac{2C_r^2}{\epsilon^2 (1 - \gamma)^2} \ln\big( \frac{2}{\delta} \big) $
then with probability greater than $1 - \delta$:
    \begin{align}
        L(\theta^*) - L(\widehat{\theta}) \leq \epsilon \label{eq:epsilon_optimal_estimator}
    \end{align}
    where $\theta^*$ is the ground truth parameter. 
\end{Corollary}

{\bf Proof:} {The finite-time guarantee
$\mathbb{E}\big[\|\nabla \widehat{L}(\widehat{\theta}_K,\mathcal{D})\|^2\big]
\leq
\mathcal{O}(K^{-1/2})$ implies $\|\nabla \widehat{L}(\widehat{\theta}_K,\mathcal{D})\|^2 \rightarrow 0$ in probability. By compactness, the set of accumulation points of the sequence $\{\widehat{\theta}_K:K\in \mathbb{N}^+\}$ is non-empty. By Corollary 1{\em (ii)}, the set of limit points is a singleton, hence $\widehat{\theta}_K \rightarrow \widehat{\theta}$ in probability. } 
To prove the performance guarantee in \eqref{eq:epsilon_optimal_estimator}, we can show the following decomposition of the error between the log likelihood objective evaluated at $\theta^*$ and $\widehat{\theta}$, respectively. With probability greater than $1 - \delta$, the following result holds:
\begin{align}
    L(\theta^*) - L(\widehat{\theta}) &= \big( L(\theta^*) - \widehat{L}(\theta^*; \mathcal{D}) \big) + \big( \widehat{L}(\theta^*;\mathcal{D}) - \widehat{L}(\widehat{\theta};\mathcal{D}) \big) + \big( \widehat{L}(\widehat{\theta};\mathcal{D}) - L(\widehat{\theta}) \big) \nonumber \\
    &\overset{(i)}{\leq} \frac{C_r}{1 - \gamma} \sqrt{\frac{\ln(2 / \delta)}{2|\mathcal{D}|}}  + \big( \widehat{L}(\theta^*) - \widehat{L}(\widehat{\theta}) \big) + \frac{C_r}{1 - \gamma} \sqrt{\frac{\ln(2 / \delta)}{2|\mathcal{D}|}}     \nonumber \\
    &= \frac{2C_r}{1 - \gamma} \sqrt{\frac{\ln(2 / \delta)}{2|\mathcal{D}|}} + \big( \widehat{L}(\theta^*;\mathcal{D}) - \widehat{L}(\widehat{\theta};\mathcal{D}) \big) \label{ineq:likelihood_difference_decomposition}
\end{align}
where $(i)$ follows \eqref{eq:approximation_error:concentration} in Lemma \ref{lemma:objective_concentration}. Since we have defined $\widehat{\theta}$ as the optimal solution to $\widehat{L}(\cdot;\mathcal{D})$, we know that $\widehat{L}(\theta;\mathcal{D}) - \widehat{L}(\widehat{\theta};\mathcal{D}) \leq 0$ for any $\theta$. Plugging this result into \eqref{ineq:likelihood_difference_decomposition}, the following result holds with probability greater than $1 - \delta$:
\begin{align}
    L(\theta^*) - L(\widehat{\theta}) &\leq \frac{2C_r}{1 - \gamma} \sqrt{\frac{\ln(2 / \delta)}{2|\mathcal{D}|}} + \big( \widehat{L}(\theta^*;\mathcal{D}) - \widehat{L}(\widehat{\theta};\mathcal{D}) \big)  \nonumber \\
    &\leq \frac{2C_r}{1 - \gamma} \sqrt{\frac{\ln(2 / \delta)}{2|\mathcal{D}|}}. \label{optimality_gap:concentration_bound}
\end{align}
Hence, when the number of expert trajectories in the demonstration dataset satisfies $|\mathcal{D}|\geq \frac{2C_r^2}{\epsilon^2 (1 - \gamma)^2} \ln\big( \frac{2}{\delta} \big) $,
then with probability greater than $1 - \delta$, we obtain \begin{align}
        L(\theta^*) - L(\widehat{\theta}) \leq \epsilon \nonumber
    \end{align}
    where $\theta^*$ is the ground truth parameter, which is optimal w.r.t. the log-likelihood objective $L(\cdot)$ defined in \eqref{eq:ML}. The corollary is proved.
\hfill $\blacksquare$

It is worth mentioning that when relaxing the assumption that the ground-truth reward value $r(s,\tilde{a};\theta^*)$ for a reference action $\tilde{a} \in A$ and $s \in S$ is known, we will no longer have a guarantee on parameter convergence. However, as shown below, the policy obtained by Algorithm 1 still converges to the expert policy. 

{By defining the state-action visitation measure $d^{\rm E}(s,a) := (1-\gamma)\pi^{\rm E}(a|s)\sum_{t = 0}^{\infty} \gamma^t P^{\pi^{\rm E}}(s_t = s| s_0 \sim \rho)$ under the expert policy $\pi^{\rm E}$, we can rewrite the expression of the log-likelihood objective $L(\cdot)$ in \eqref{eq:ML} for any reward parameter $\theta$ as below:
\begin{align}
    L(\theta) := \mathbb{E}_{\tau^{\rm E} \sim \pi^{\rm E}} \bigg[ \sum_{t = 0}^{\infty} \gamma^t \log \pi_{\theta}(a_t | s_t)  \bigg] = \frac{1}{1 - \gamma} \mathbb{E}_{s \sim d^{\rm E}(\cdot), a \sim \pi^{\rm E}(\cdot | s)} \Big[ \log \pi_{\theta}(a|s) \Big]. \nonumber
\end{align}
}
Then the $\epsilon$-optimal solution on the maximum likelihood IRL problem \eqref{eq:ML:problem} implies:
\begin{align}
    L(\theta^*) - L(\widehat{\theta}) = \frac{1}{1 - \gamma} \mathbb{E}_{s \sim d^{\rm E}(\cdot), a \sim \pi^{\rm E}(\cdot | s)} \big[ \log \big( \frac{\pi_{\theta^*}(a|s)}{\pi_{\widehat{\theta}}(a|s)} \big) \big] \leq \varepsilon   \nonumber
\end{align}
where $d^{\rm E}(s,a) := (1-\gamma)\pi^{\rm E}(a|s)\sum_{t = 0}^{\infty} \gamma^t P^{\pi^{\rm E}}(s_t = s| s_0 \sim \rho)$ denotes the state-action visitation measure under the expert policy $\pi^{\rm E}$. Assume the expert behaviors are consistent with optimal behavior for a ground truth reward parameter $\theta^*$, then it follows
 $\pi^{\rm E} = \pi_{\theta^*}$. Due to this property, we can obtain the following result: \begin{align}
    L(\theta^*) - L(\widehat{\theta}) = \frac{1}{1 - \gamma} \mathbb{E}_{s \sim d^{\rm E}(\cdot), a \sim \pi^{\rm E}(\cdot | s)} \big[ \log \big( \frac{\pi^{\rm E}(a|s)}{\pi_{\widehat{\theta}}(a|s)} \big) \big] = \frac{1}{1 - \gamma} \mathbb{E}_{s \sim d^{\rm E}(\cdot)} \big[ D_{KL}\big( \pi^{\rm E}(\cdot | s) || \pi_{\widehat{\theta}}(\cdot | s) \big) \big] \leq \varepsilon. \nonumber
\end{align}
Hence, Corollary \ref{corollary:optimality_guarantee} provides a formal guarantee that the recovered policy $\pi_{\widehat{\theta}}$ solved from the empirical estimation problem \eqref{ML:estimation} is $\epsilon$-close to the expert policy $\pi^{\rm E}$ measured by the KL divergence.




{\bf Remark 4:} We also believe the results for the linear reward parametrization case can be generalized to certain nonlinear parametric rewards representations. Such is the case, for example, of  {\it overparameterized} neural networks. In this setting, under certain structural assumptions such as Neural Tangent Kernel and Local Linearity (see \cite{du2018gradient} and \cite{Jacot_2018}) we expect that the resulting reward representation is approximately linear in the  parameters. Hence, it would be possible to identify the global optimal reward estimator. These directions are left for future research. %
}



\section{The Case with State-only Dependent Rewards}
\label{Sec: State-only}

{In this section we consider the IRL problems when the reward is only a function of the state. A lower dimensional representation of the agent's preferences (i.e. in terms only of states as opposed to states {\em and} actions) is more likely to facilitate counterfactual analysis such as predicting
the optimal policy under different environment dynamics and/or learning new
tasks. This is because the estimation of preferences which are only defined in terms of states is less sensitive to the specific environment dynamics in the expert's demonstration dataset. } {Moreover, in application such as healthcare \cite{yu2021reinforcement} and autonomous driving \cite{kiran2021deep}, where simply imitating the expert policy can potentially result in poor performance, since the learner and the expert may have different transition dynamics. Similar points have also been argued in recent works \cite{gangwani2020state,ni2020f,viano2021robust}.



Next, let us briefly discuss how we can understand \eqref{eq:ML:problem} and Alg. \ref{alg:IRL}, when the reward is parameterized as a state-only function.} First, it turns out that there is an equivalent formulation of \eqref{eq:ML}, when the expert trajectories only contain the visited states. 


\begin{lemma} \label{lemma:formulation_equivalence}
	Suppose the reward is parameterized as a state-only function $r(s;\theta)$. Then \eqref{eq:ML:problem} is equivalent to the following:
	\begin{subequations}
	    \label{formulation:min_soft_value_gap}
	    \begin{align}
	    &\min_{\theta} ~~ \ee_{s_0 \sim \rho(\cdot)}\big[ V_{\theta}(s_0) \big] - \ee_{s_0 \sim \rho(\cdot)}\big[ V_{\theta}^{\rm E}(s_0) \big] \label{obj:gap_soft_value_func} \\
	    & s.t. ~~ \pi_{\theta} := \arg \max_{\pi} ~ \mathbb{E}_{\tau^{\rm A} \sim \pi} \bigg[ \sum_{t = 0}^{\infty} \gamma^t \bigg( r(s_t; \theta) + \mathcal{H}(\pi(\cdot | s_t)) \bigg) \bigg] \label{def:state_only_inner_problem}
	    \end{align}
	\end{subequations}

where $V_{\theta}^{\rm E}(\cdot)$ denotes the soft value function under reward parameter $\theta$ and the expert policy $\pi^{\rm E}$. 
\end{lemma}
Please see the Section \ref{proof:formulation_equivalence_lemma} in the supplementary material for detailed derivation. 
Intuitively, the above lemma says that, when dealing with the state-only IRL, \eqref{obj:gap_soft_value_func} minimizes the gap between the soft value functions of the optimal policy $\pi_{\theta}$ and the { expert policy $\pi^{\rm E}$}.
{Moreover, Alg. \ref{alg:IRL} can also be easily implemented with the state-only reward. In fact, the entire algorithm essentially stays the same, and the only change is that $r(s,a;\theta)$ will be replaced by $r(s;\theta)$. In this way, by only using the visited states in the trajectories, one can still  compute the stochastic gradient estimator in \eqref{eq:stochastic_grad_estimator}. 
Therefore, even under the state-only IRL setting where the expert dataset only contains visited states, our formulation and the proposed algorithm still work if we parameterize the reward as a state-only function.}  Moreover, it is straightforward to show that the convergence results in Theorem \ref{theorem:main_convergence_results} also hold under the state-only IRL setting.

\section{Extension to the Offline Setting}\label{sec:offline}

{ Throughout this paper, we have focused on the {\it online} setting where the transition kernel $P(s_{t+1}|s_t,a_t)$ is known or alternatively, samples from such kernel are available to the learner in an online fashion. However, in many applications this assumption does not hold and the available data is fixed. In such an {\em offline} setting, one strategy to deal with the problem is to estimate {\it both} the transition kernel and the reward function based on the finite dataset of state-action sequences.  
In our follow-up work \cite{offline-irl23} to the present paper, we have extended Algorithm 1 to the offline setting. In particular, a two-stage estimation procedure has been proposed, where in the first stage a maximum likelihood estimate of the transition kernel is obtained from transition triples $(s,a,s')$ in a transition dataset denoted as $\mathcal{D}^T$, i.e. $
      \hat{P} := \arg \max_{\tilde{P}} \mathbb{E}_{(s,a,s^\prime) 
\sim \mathcal{D}^T}\bigg[ \log \tilde{P}(s^\prime|s,a)\bigg].$ 
Given that finite-data estimation of high-dimensional environment dynamics likely leads to an inaccurate model, in the second stage, a ``conservative" reward estimator is obtained using $\hat{P}$ by introducing a regularization term $U(s,a)$ to account model uncertainty:
 \begin{subequations} \label{formulation:offline_irl}
     \begin{align} 
	\max_{\theta} 	&~~~~ \widehat{L}(\theta):=\mathbb{E}_{\tau^{\rm E} \sim \mathcal{D}} \big[ \sum_{t\geq 0} \gamma^t \log \pi_{\theta}(a_t | s_t)  \big] \label{upper:likelihood} \\
	s.t. &~~~~ 
\pi_{\theta} := \arg \max_{\pi} ~ \mathbb{E}_{\tau^{\rm A} \sim (\rho, \pi, \widehat{P})} \Big[ \sum_{t\geq0} \gamma^t \Big( r(s_t, a_t;\theta) +\mathcal{H}(\pi(\cdot|s_t)) -U(s_t, a_t)  \Big) \Big] \label{lower:offline_policy_optimization}
\end{align}
 \end{subequations}
 
The regularization term in the lower level problem \eqref{lower:offline_policy_optimization} induces {\em conservative} policies which assign low probability to state-action pairs in which $\widehat{P}$ cannot provide an accurate prediction on the dynamics. Clearly, the second stage is closely related to the online setting discussed in this work. Therefore algorithms and intuitions developed in the present work for the online setting is crucial for the offline setting as well. 

There are many other outstanding issues to be resolved for the offline setting. For example, how well the estimated transition  function  can be recovered,  how the error will propagate to the error of the reward estimation, and how to compute (stochastic) gradient for the new formulation
\eqref{upper:likelihood} and
\eqref{lower:offline_policy_optimization}. Since these investigations are out of the scope of this paper, we refer the readers to in \cite{offline-irl23} for more details.}

\section{Testbed}
\label{Sec:Testbed}

In this section, we test the performance of our algorithm { with limited expert trajectories} on a diverse collection of RL tasks and environments. In each experiment set, we train algorithms until convergence and average the scores of the trajectories over multiple random seeds. 



{\bf Mujoco Tasks For Inverse Reinforcement Learning.} In this experiment set, we test the performance of our algorithm on imitating the expert behavior. We consider several high-dimensional robotics control tasks in Mujoco \cite{todorov2012mujoco}. 
{Two classes of existing algorithms are considered as the comparison baselines: 1) imitation learning algorithms that only learn the policy to imitate the expert, including Behavior Cloning (BC) \cite{pomerleau1988alvinn} and Generative Adversarial Imitation Learning (GAIL) \cite{ho2016generative}; 2) IRL algorithms which learn a reward function {\it and} a policy simultaneously, including Adversarial Inverse Reinforcement Learning (AIRL) \cite{fu2017learning}, $f$-IRL \cite{ni2020f} and IQ-Learn \cite{garg2021iq}.}
To ensure fair comparison, all imitation learning / IRL algorithms use soft Actor-Critic \cite{haarnoja2018soft} as the base RL algorithm. 
For the expert dataset, we use the data provided in the official implementation\footnote{\url{https://github.com/twni2016/f-IRL}} of $f$-IRL.

\begin{table}[t]
\begin{tabular}{cccccccc}
\hline
Task&BC&GAIL&IQ-Learn&$f$-IRL&ML-IRL&ML-IRL&Expert\\
&&&&&(State-Only)&(State-Action)&\\
\hline
Hopper&102.74&2762.77&3039.21&3116.02&3131.45&{\bf3290.02}&3530.63 \\
Half-Cheetah&155.64&3085.18&4562.51&4751.63&4661.04&{\bf 4846.43}&5072.53\\
Walker&283.43&3610.49&4361.27&4562.48&4367.81&{\bf4703.35}&5471.58\\
Ant&961.58&2971.57&$4362.90^*$&5124.13&4832.38&{\bf5157.03}&5856.84\\
Humanoid&547.62&3174.66&$5227.10^*$&{\bf 5399.67}&5149.39&5281.93&5339.12\\
\hline
\end{tabular}
\caption{\footnotesize{\bf Mujoco Results.} The performance of benchmark algorithms under {five expert trajectories}.}
\label{table:mujoco_benchmarks}
\end{table}

In this experiment, we implement two versions of our proposed algorithm: ML-IRL(State-Action) where the reward is parameterized as a function of state and action; ML-IRL(State-Only) which utilizes the state-only reward function. In Table \ref{table:mujoco_benchmarks}, we present the simulation results under a limited data regime
where only five expert trajectories are collected. 
{The scores (cumulative rewards) reported in the table is averaged over $5$ random seeds. In each random seed, we train algorithm from initialization and collect $20$ trajectories to average their cumulative rewards after the algorithms converge.
}
According to the results reported in Table \ref{table:mujoco_benchmarks} { where we run the experiments with only five expert trajectories in the demonstration dataset $\mathcal{D}$}, it shows that our proposed algorithms outperform the baselines on most tasks. 

{We observe that BC fails to imitate the expert's behavior. It is likely due to the fact that BC is based on supervised learning and thus could not learn a good policy under such a limited data regime.} Moreover, we notice the training of IQ-Learn is unstable, likely due to its inaccurate approximation to the soft Q-function. 
Therefore, in the Mujoco tasks where IQ-Learn does not perform well, we cannot match the results presented in the original paper \cite{garg2021iq}. For those cases, we directly report results from the original paper (and mark them by $*$ in Table \ref{table:mujoco_benchmarks}). The results of AIRL are not presented in Table \ref{table:mujoco_benchmarks} since it performs poorly even
after spending significant efforts in parameter tuning; note that similar observations have been made in \cite{liu2019state,ni2020f}.

\begin{table}[t]
\centering
\begin{tabular}{ccccccc}
\hline
Setting&IQ-Learn&AIRL&$f$-IRL&ML-IRL (State-Only)& Ground-Truth\\
\hline
Data Transfer&-11.78&-5.39& 188.85 &{\bf 221.51}&320.15\\
Reward Transfer&-1.04&130.3& 156.45 &{\bf 187.69}&320.15\\
\hline
\end{tabular}
\caption{\footnotesize{\bf Transfer Learning.} The performance of benchmark algorithms under a single expert trajectory.
The scores in the table are obtained similarly as in Table \ref{table:mujoco_benchmarks}.}
\label{experiment:transfer_learning}
\end{table}

{\bf Transfer Learning Across
Changing Dynamics.} We further evaluate  IRL algorithms on the transfer learning setting. We follow the environment setup in \cite{fu2017learning}, where two environments with different dynamics are considered: {\tt{Custom-Ant}} vs {\tt{Disabled-Ant}}. We compare ML-IRL(State-Only) with several existing IRL methods: 1) AIRL \cite{fu2017learning}, 2) $f$-IRL
 \cite{ni2020f}; 3) IQ-Learn \cite{garg2021iq}.
 
We consider two transfer learning settings: 1) data transfer; 2) reward transfer. For both settings, the expert dataset / trajectories are generated in {\tt{Custom-Ant}}. In the data transfer setting, we train IRL agents in {\tt{Disabled-Ant}} by using the expert trajectories, which are generated in {\tt{Custom-Ant}}. In the reward transfer setting, we first use IRL algorithms to infer the reward functions in {\tt{Custom-Ant}}, and then transfer these recovered reward functions to {\tt{Disabled-Ant}} for further evaluation. {In both settings, we also train SAC with the ground-truth reward in {\tt{Disabled-Ant}} and report the scores.}

The numerical results are reoprted in Table \ref{experiment:transfer_learning}. the proposed  ML-IRL(State-Only) achieves superior performance compared with the existing IRL benchmarks in both settings. We notice that IQ-Learn fails in both settings since it indirectly recovers the reward function from a soft Q-function approximator, which could be inaccurate and is highly dependent upon the environment dynamics. Therefore, the reward function recovered by IQ-Learn cannot be disentangled from the expert actions and environment dynamics, which leads to its failures in the transfer learning tasks.

\section{Conclusions}
\label{Sec:Conclusions}
The nested structure of the structural estimation of MDPs entails a significant computational burden in environments with a high-dimensional continuous state or discrete state with large cardinality. To alleviate such burden several approaches have been proposed in both the econometrics (dynamic discrete choice estimation) and artificial intelligence (inverse reinforcement learning) literature.
For example, the approximation algorithms in \cite{Hotz,Hotz1994} reduce the computational burden but the resulting estimates suffer from finite sample bias because in high dimensional state space, initial policy estimates are likely of poor quality. Recent approaches in inverse reinforcement learning that lessen the computational burden \cite{garg2021iq, ni2020f} do so either at the expense of reward estimation accuracy or lack theoretical guarantees.

In this paper we introduce a class of single-loop algorithms for the structural estimation of MDPs with non-linear parametrization. In each iteration a policy improvement step is followed by a stochastic gradient step for likelihood maximization. We show that the proposed algorithm converges to a stationary solution with a finite-time guarantee. Further, if the reward is parameterized linearly, we show that the algorithm approximates the maximum likelihood estimator in sub-linear time. Extensive experimentation in standard testbeds for robotics control problems
 show that the proposed algorithm achieves superior performance compared with other IRL and imitation learning approaches.
 In future work we will consider extensions of the proposed algorithm when a model of the state dynamics is not available and thus must also be estimated.

\section*{Appendix}


\section{Auxiliary Lemmas}
Before starting the proof of the main theorems in this paper, we first introduce several supporting lemmas in this section.
Throughout this section, we assume Assumptions \ref{Assumption:Ergodicity_Markov_chain} - \ref{Assumption:reward_grad_bound} hold true.
\begin{lemma} \label{lemma:Lipschitz_visitation_measure}
	\cite[Lemma 3]{xu2020improving} Consider the initialization distribution $\rho(\cdot)$ and transition kernel $P( \cdot | s,a)$. Under $\rho(\cdot)$ and $P( \cdot | s,a)$, denote $d_w(\cdot, \cdot)$ as the state-action visitation distribution of MDP with the softmax policy parameterized by parameter $w$. Suppose Assumption \ref{Assumption:Ergodicity_Markov_chain} holds, for all policy parameter $w$ and $w^\prime$, we have 
	\begin{align}
		\| d_{w}(\cdot, \cdot) - d_{w^\prime}(\cdot, \cdot)  \|_{TV} \leq C_d \| w - w^\prime \| 	\label{ineq:lipschitz_measure}
	\end{align}
	where $C_d$ is a positive constant.
\end{lemma}

\begin{lemma}\label{lemma:Lipschitz_Q_different_r}
	Suppose Assumption \ref{Assumption:reward_grad_bound} holds. Under the approximated soft policy iteration in \eqref{def:approximated_SPI}, denote the soft Q-function under reward parameter $\theta_k$ and policy $\pi_{k+1}$ as $Q_{k + \frac{1}{2}}$; further note that $Q_{k+1}$ has been defined as the soft Q-function under the reward parameter $\theta_{k+1}$ and policy $\pi_{k+1}$. Then for any $s \in \mathcal{S}$, $a \in \mathcal{A}$ and $k \geq 0$, the following inequality holds: 
	\begin{align}
	     | Q_{k+	\frac{1}{2}} (s,a) -  Q_{k+1} (s,a) | \leq L_q \| \theta_{k} - \theta_{k+1} \|, 
	\end{align}
	where $L_q := \frac{L_r}{1 - \gamma}$ and $L_r$ is the positive constant defined in Assumption \ref{Assumption:reward_grad_bound}.
\end{lemma}

\begin{lemma}  
\label{lemma:approx_policy_improvement}
Using approximated soft policy iteration \eqref{def:approximated_SPI}, the following holds for any iteration $k\geq0$:
\begin{align}
     Q_{k}(s,a)  &\leq Q_{k+\frac{1}{2}}(s,a) + \frac{2\gamma \epsilon_{\rm app}}{1 - \gamma}, \quad \forall s \in \mathcal{S}, a \in \mathcal{A} \label{ineq:approx_soft_policy_improvement}, \\
     \| Q_{\theta_k} - Q_{k+\frac{1}{2}} \|_{\infty} &\leq \gamma \| Q_{\theta_k} - Q_{k} \|_{\infty} + \frac{2\gamma \epsilon_{\rm app}}{1 - \gamma} \label{ineq:soft_Q_contraction}
\end{align}
where $Q_{k+\frac{1}{2}}(\cdot, \cdot)$ denotes the soft Q-function under reward parameter $\theta_k$ and updated policy $\pi_{k+1}$, and $\| Q_{\theta_k} - Q_{k+\frac{1}{2}} \|_{\infty} = \max_{s \in 
\mathcal{S}} \max_{a \in \mathcal{A}} | Q_{\theta_k}(s,a) - Q_{k+\frac{1}{2}}(s,a) |$.
\end{lemma}

\section{Proof of Theorem \ref{theorem:main_convergence_results}}
\label{proof:main_convergence_theorem}

In this section, we prove \eqref{rate:lower_error} and \eqref{rate:upper_grad_norm} respectively, to show the convergence of the lower-level problem and the upper-level problem.

\subsection{Proof of Relation \eqref{rate:lower_error}}
	In this proof, we first show the convergence of the lower-level variable $\{\pi_k\}_{k\geq0}$. Recall that we approximate the optimal policy $\pi_{\theta_k}$ by $\pi_{k+1}$ at each iteration $k$. Moreover, the policy $\pi_{k+1}$ is generated as below:
	\begin{align}
	    \pi_{k+1}(a|s) \propto \exp\big( \widehat{Q}_{k}(s,a) \big), \text{ where } \| \widehat{Q}_{k} - Q_k \|_{\infty} \leq \epsilon_{\rm app}. \label{update:policy}
	\end{align}
	We first analyze the approximation error between $\pi_{\theta_k}$ and $\pi_{k+1}$. Recall that both policies $\pi_{k+1}$ and $\pi_{\theta_k}$ are in the softmax form parameterized by $\widehat{Q}_{k}$ and $Q_{\theta_k}$, then it holds
	\begin{align}
	   \| \log \pi_{k+1} - \log \pi_{\theta_k} \|_{\infty} &\overset{(i)}{\leq} 2 \| \widehat{Q}_{k} - Q_{\theta_k} \|_{\infty} = 2 \| \widehat{Q}_{k} - Q_{k} + Q_{k} - Q_{\theta_k} \|_{\infty} \leq 2\epsilon_{\rm app} + 2\| Q_{k} - Q_{\theta_k} \|_{\infty}
	   \label{ineq:infty_policy_gap}
	\end{align}
	where (i) follows the Lipschitz property of softmax policy, which is shown in proof of Lemma \ref{lemma:approx_policy_improvement}.
	
	Based on the inequality \eqref{ineq:infty_policy_gap}, we further analyze $\| Q_{k} - Q_{\theta_k} \|_{\infty}$ to show the convergence of the policy estimates. Here, we use an auxiliary sequence $\{ Q_{k + \frac{1}{2}} \}_{k \geq 0}$, where $Q_{k + \frac{1}{2}}$ is defined as the soft Q-function under reward parameter $\theta_k$ and the policy $\pi_{k+1}$, its expression follows its\begin{align}
	    Q_{k+\frac{1}{2}}(s,a) := r(s,a;\theta_k) +  \mathbb{E}_{\tau^{\rm A} \sim \pi_{k+1}} \bigg[ \sum_{t = 1}^{\infty} \gamma^t \bigg( r(s_t, a_t;\theta_k) + \mathcal{H}(\pi_{k+1}(\cdot | s_t)) \bigg) \bigg | (s_0, a_0) = (s,a) \bigg]. \label{def:soft_Q_semi_expression}
	\end{align} Then, the following relations hold:
	\begin{align}
		 \|  Q_{k} - Q_{\theta_k} \|_{\infty}
		&=  \|  Q_{k} -  Q_{\theta_k} +  Q_{\theta_{k-1}} -  Q_{\theta_{k-1}}  +  Q_{k- \frac{1}{2}}  -  Q_{k-\frac{1}{2}} \|_{\infty}  \nonumber  \\
		&\leq  \| Q_{\theta_k} - Q_{\theta_{k-1}} \|_{\infty}  + \| Q_{k-\frac{1}{2}} - Q_{\theta_{k-1}} \|_{\infty} + \| Q_k -  Q_{k-\frac{1}{2}} \|_{\infty}		\nonumber \\
		&\overset{(i)}{\leq}  L_q \| \theta_{k} - \theta_{k-1} \|  +  \| Q_{k-\frac{1}{2}} - Q_{\theta_{k-1}} \|_{\infty}  + \| Q_k -  Q_{k-\frac{1}{2}} \|_{\infty} \nonumber \\
		&\overset{(ii)}{\leq}  \| Q_{k-\frac{1}{2}} - Q_{\theta_{k-1}} \|_{\infty} + 2 L_q \| \theta_k - \theta_{k-1} \|  \label{bound:soft_Q_difference}
	\end{align}
	where (i) is from \eqref{ineq:soft_Q_Lipschitz} in Lemma \ref{lemma:Lipschitz_properties}; (ii) follows Lemma \ref{lemma:Lipschitz_Q_different_r}. Based on \eqref{bound:soft_Q_difference}, we further analyze the two terms in \eqref{bound:soft_Q_difference} as below.
	
	Recall that we have already shown the following relation in \eqref{ineq:soft_Q_contraction}:
	\begin{align}
	    \| Q_{\theta_k} - Q_{k+\frac{1}{2}} \|_{\infty} \leq \gamma \| Q_{\theta_k} - Q_{k} \|_{\infty} + \frac{2\gamma\epsilon_{\rm app}}{1 - \gamma}.  \label{ineq:approximate_policy_improvement}
	\end{align}
	
	Through plugging \eqref{ineq:approximate_policy_improvement} into \eqref{bound:soft_Q_difference}, we have the following result:
	\begin{align}
	     \|  Q_k - Q_{\theta_k} \|_{\infty} 
		&\leq  \| Q_{k-\frac{1}{2}} - Q_{\theta_{k-1}} \|_{\infty} + 2 L_q \| \theta_k - \theta_{k-1} \| 
		\nonumber \\
		&\leq \gamma \| Q_{\theta_{k-1}} - Q_{k-1} \|_{\infty} + \frac{2\gamma\epsilon_{\rm app}}{1 - \gamma}  + 2 L_q \| \theta_k - \theta_{k-1} \|. \label{constraction:approx_soft_q_update}
	\end{align}
	To show the convergence of the soft Q-function based on \eqref{constraction:approx_soft_q_update}, we further analyze the error between the reward parameters $\theta_k$ and $\theta_{k-1}$. Recall that in Alg. \ref{alg:IRL}, the reward parameter is updated as:
	\begin{align}
		& \theta_k = \theta_{k-1} + \alpha g_{k-1} = \theta_{k-1} + \alpha \big(h(\theta_{k - 1}, \tau^{\rm E}_{k-1}) -  h(\theta_{k - 1}, \tau^{\rm A}_{k-1}) \big) \nonumber 
	\end{align}
	where we denote $\tau := \{ (s_t, a_t) \}_{t = 0}^{\infty}$, $h(\theta, \tau) := \sum_{t = 0}^{\infty} \gamma^t \nabla_{\theta} r(s_t,a_t; \theta) $ and $g_{k-1}$ is the stochastic gradient estimator at iteration $k-1$. Here, $\tau^{\rm E}_{k-1}$ denotes the trajectory sampled from the expert's dataset $D$ at iteration $k-1$, and $\tau^{\rm A}_{k-1}$ denotes the trajectory sampled from the agent's policy $\pi_k$ at time $k-1$. Then according to the inequality \eqref{ineq:Lipschitz_smooth_reward} in Assumption \ref{Assumption:reward_grad_bound}, we could show that \begin{align}
		\|  g_{k-1}  \| \leq \| h(\theta_{k - 1}, \tau^{\rm E}_{k-1}) \| + \| h(\theta_{k - 1}, \tau^{\rm A}_{k-1}) \| \leq  2 L_r \sum_{t = 0}^{\infty} \gamma^t  = \frac{2 L_r}{1 - \gamma} = 2L_q \label{bound:upper_grad}
	\end{align}
	where the last equality follows the fact that we have defined the constant $L_q := \frac{L_r}{1 - \gamma}$. Then we could further show that
	\begin{align}
	     \|  Q_k - Q_{\theta_k} \|_{\infty} 
	&\overset{(i)}{\leq} \gamma \| Q_{\theta_{k-1}} - Q_{k-1} \|_{\infty} + \frac{2\gamma\epsilon_{\rm app}}{1 - \gamma}  + 2 L_q \| \theta_k - \theta_{k-1} \|
		\nonumber \\
		&\overset{(ii)}{=} \gamma \| Q_{\theta_{k-1}} - Q_{k-1} \|_{\infty} + \frac{2\gamma\epsilon_{\rm app}}{1 - \gamma} + 2 \alpha L_q \| g_{k-1} \| \nonumber \\
		&\overset{(iii)}{\leq} \gamma \| Q_{\theta_{k-1}} - Q_{k-1} \|_{\infty} + \frac{2\gamma\epsilon_{\rm app}}{1 - \gamma} + 4\alpha L_q^2 \label{contraction_bound:optimal_gap_soft_Q}
	\end{align}
	where (i) is from \eqref{constraction:approx_soft_q_update}; (ii) follows the reward update scheme in \eqref{eq:reward_parameter_update}; (iii) is from \eqref{bound:upper_grad}. 
	
	Summing the inequality \eqref{contraction_bound:optimal_gap_soft_Q} from $k = 1$ to $k = K$, it holds that \begin{align}
		\sum_{k = 1}^{K} \|  Q_k - Q_{\theta_k} \|_{\infty} \leq \gamma \sum_{k = 0}^{K - 1} \| Q_{k} - Q_{\theta_{k-1}} \|_{\infty} + K \frac{2\gamma\epsilon_{\rm app}}{1 - \gamma} +  4\alpha K L_q^2. \label{bound:sum_lower_error}
	\end{align}
	Rearranging the inequality \eqref{bound:sum_lower_error} and divided \eqref{bound:sum_lower_error} by $K$ on both sides, it holds that
	\begin{align}
		& \frac{1 - \gamma}{K} \sum_{k = 1}^{K} \| Q_k - Q_{\theta_k} \|_{\infty} \leq \frac{\gamma}{K} \bigg( \| Q_{0} - Q_{\theta_0} \|_{\infty} -  \| Q_{K} - Q_{\theta_K} \|_{\infty} \bigg) + \frac{2\gamma\epsilon_{\rm app}}{1 - \gamma} +  4\alpha L_q^2.  \label{bound:lower_rate}
	\end{align}
	Dividing the constant $ 1 - \gamma $ on both sides of \eqref{bound:lower_rate}, it holds that 
	\begin{align}
		\frac{1}{K} \sum_{k = 1}^{K} \| Q_k - Q_{\theta_k}  \|_{\infty} \leq \frac{\gamma C_0}{K(1 - \gamma)} +  \frac{2\gamma\epsilon_{\rm app}}{(1-\gamma)^2} + \frac{4 L_q^2}{1 - \gamma} \alpha \label{gap:soft_Q_difference}
	\end{align}
	where we denote $C_0 := \| Q_{0} - Q_{\theta_0} \|_{\infty} $. Add $\|Q_0 - Q_{\theta_0} \|_{\infty}$ and subtract $\|Q_K - Q_{\theta_K} \|_{\infty}$ on both sides of \eqref{gap:soft_Q_difference}, it follows that 
	\begin{align}
		 \frac{1}{K} \sum_{k = 0}^{K-1} \| Q_{k} - Q_{\theta_k} \|_{\infty}
		& \leq \frac{\gamma C_0}{K(1 - \gamma)} + \frac{C_0}{K} - \frac{\| Q_{K} - Q_{\theta_K} \|_{\infty}}{K} + \frac{2\gamma\epsilon_{\rm app}}{(1 - \gamma)^2} + \frac{4 L_q^2}{1 - \gamma} \alpha \nonumber \\
		& \leq \frac{C_0}{K(1 - \gamma)} + \frac{2\gamma\epsilon_{app}}{(1 - \gamma)^2}  + \frac{4 L_q^2}{1 - \gamma} \alpha. \nonumber
	\end{align}
	
	Recall the stepsize is defined as $\alpha = \frac{\alpha_0}{K^\sigma}$ where $\sigma>0$. Then we have:
	\begin{align}
		\frac{1}{K} \sum_{k = 0}^{K-1} \| Q_{k} - Q_{\theta_k} \|_{\infty}   = \mathcal{O}(K^{-1}) + \mathcal{O}(K^{-\sigma}) + \mathcal{O}(\epsilon_{\rm app}). \label{convergence_rate:soft_Q_gap}
	\end{align}
	Summing the inequality \eqref{ineq:infty_policy_gap} from $k = 0$ to $K-1$, it holds that
	\begin{align}
	   \frac{1}{K} \sum_{k = 0}^{K-1} \| \log \pi_{k+1} - \log \pi_{\theta_k} \|_{\infty}  \leq \frac{2}{K} \sum_{k = 0}^{K-1} \big( \epsilon_{\rm app} + \| Q_{k} - Q_{\theta_k} \|_{\infty}\big)
	   = \mathcal{O}(K^{-1}) + \mathcal{O}(K^{-\sigma}) + \mathcal{O}(\epsilon_{\rm app}). \nonumber
	\end{align}
	Therefore, we complete the proof of \eqref{rate:lower_error} in Theorem \ref{theorem:main_convergence_results}. \hfill $\blacksquare$

\subsection{Proof of relation \eqref{rate:upper_grad_norm}}

    In this part, we prove the convergence of reward parameters $\{\theta_k\}_{k \geq 0}$.
    
	We have the following result of the empirical estimation objective $\widehat{L}(\theta;\mathcal{D})$:
	\begin{align}
		\widehat{L}(\theta_{k+1};\mathcal{D}) &\overset{(i)}{ \geq } \widehat{L}(\theta_{k};\mathcal{D}) + \langle \nabla_{\theta} \widehat{L}(\theta_{k};\mathcal{D}), \theta_{k+1} - \theta_{k}  \rangle - \frac{L_c}{2} \| \theta_{k+1}  - \theta_{k} \|^2 \nonumber \\
		&\overset{(ii)}{=} \widehat{L}(\theta_{k};\mathcal{D}) + \alpha \langle \nabla_{\theta} \widehat{L}(\theta_{k};\mathcal{D}), g_k \rangle - \frac{L_c \alpha^2}{2} \| g_k \|^2 \nonumber \\
		&=  \widehat{L}(\theta_k;\mathcal{D}) + \alpha \langle \nabla_{\theta} \widehat{L}(\theta_{k};\mathcal{D}), g_k - \nabla_{\theta} \widehat{L}(\theta_{k};\mathcal{D}) \rangle + \alpha \|  \nabla_{\theta} \widehat{L}(\theta_{k};\mathcal{D}) \|^2  - \frac{L_c \alpha^2}{2} \| g_k \|^2 \nonumber \\
		&\overset{(iii)}{ \geq } \widehat{L}(\theta_{k};\mathcal{D}) + \alpha \langle \nabla_{\theta} \widehat{L}(\theta_{k};\mathcal{D}), g_k - \nabla_{\theta} \widehat{L}(\theta_{k};\mathcal{D}) \rangle + \alpha \|  \nabla_{\theta} \widehat{L}(\theta_{k};\mathcal{D}) \|^2  - 2L_c L_q^2 \alpha^2 \label{ineq:upper_grad_ascent}
	\end{align}
	where (i) is from the Lipschitz smooth property in \eqref{ineq:objective_lipschitz_smooth} of Lemma \ref{lemma:Lipschitz_properties}; (ii) follows the reward update scheme \eqref{eq:reward_parameter_update}; (iii) is from constant bound of the gradient estimator $g_k$ in \eqref{bound:upper_grad}. 
	
	Taking an expectation over the both sides of \eqref{ineq:upper_grad_ascent}, it holds that
	\begin{align}
		& \mathbb{E} \left[ \widehat{L}(\theta_{k+1}; \mathcal{D}) \right] \nonumber \\
		&\geq \mathbb{E} \left[ \widehat{L}(\theta_{k};\mathcal{D}) \right] + \alpha \mathbb{E} \bigg[ \langle \nabla_{\theta} \widehat{L}(\theta_{k};\mathcal{D}) , g_k - \nabla_{\theta} \widehat{L}(\theta_{k};\mathcal{D}) \rangle \bigg] + \alpha \mathbb{E} \bigg[ \|  \nabla_{\theta} \widehat{L}(\theta_{k};\mathcal{D}) \|^2 \bigg] - 2L_c L_q^2 \alpha^2  \nonumber \\ 
		&= \mathbb{E} \left[ \widehat{L}(\theta_{k};\mathcal{D}) \right] + \alpha \mathbb{E} \bigg[ \langle \nabla_{\theta} \widehat{L}(\theta_{k};\mathcal{D}), \mathbb{E} \big[g_k - \nabla_{\theta} \widehat{L}(\theta_{k};\mathcal{D}) \big | \theta_k] \rangle \bigg] + \alpha \mathbb{E} \bigg[ \|  \nabla_{\theta} \widehat{L}(\theta_{k};\mathcal{D}) \|^2 \bigg] - 2L_c L_q^2 \alpha^2  \nonumber \\
		&\overset{(i)}{=}  \mathbb{E} \left[ \widehat{L}(\theta_{k};\mathcal{D}) \right] + \alpha \ee \bigg[ \bigg \langle  \nabla_{\theta} \widehat{L}(\theta_{k};\mathcal{D}), \ee_{ \tau^{\rm A} \sim \pi_{\theta_{k}}}\bigg[ \sum_{t \geq 0} \gamma^t \nabla_{\theta} r(s_t, a_t; \theta_{k}) \bigg]  -  \ee_{ \tau^{\rm A} \sim \pi_{k+1}}\bigg[ \sum_{t \geq 0} \gamma^t \nabla_{\theta} r(s_t, a_t; \theta_{k}) \bigg] \bigg \rangle \bigg] \nonumber \\
		& \quad + \alpha \mathbb{E} \bigg[ \|  \nabla_{\theta} \widehat{L}(\theta_{k};\mathcal{D}) \|^2 \bigg] - 2L_c L_q^2 \alpha^2 \nonumber  \\
		&\overset{(ii)}{ \geq }  \mathbb{E} \left[ \widehat{L}(\theta_{k};\mathcal{D}) \right]  - 2\alpha L_q \underbrace{ \ee \left[ \bigg \| \ee_{ \tau^{\rm A} \sim \pi_{\theta_{k}}}\bigg[ \sum_{t = 0}^{\infty} \gamma^t \nabla_{\theta} r(s_t, a_t; \theta_{k}) \bigg]  -  \ee_{ \tau^{\rm A} \sim \pi_{k+1}}\bigg[ \sum_{t = 0}^{\infty} \gamma^t \nabla_{\theta} r(s_t, a_t; \theta_{k}) \bigg] \bigg \| \right] }_{\rm term~A} \nonumber \\ 
		& \quad + \alpha \mathbb{E} \bigg[ \|  \nabla_{\theta} \widehat{L}(\theta_{k};\mathcal{D}) \|^2 \bigg] - 2L_c L_q^2 \alpha^2 \label{bound:objective_gradient_ascent}
	\end{align}
	where (i) follows the expressions of $\nabla_{\theta} \widehat{L}(\theta;\mathcal{D})$ in \eqref{eq:likelihood_grad_app} and the gradient estimator $g_k$ in \eqref{eq:stochastic_grad_estimator}; (ii) is due to the fact $\| \nabla_{\theta} \widehat{L}(\theta;\mathcal{D}) \| \leq 2L_q$ according to \eqref{bound:upper_grad}.
	
	Then we further analyze the term A as below:
	\begin{align}
		& \ee \left[ \bigg \| \ee_{ \tau^{\rm A} \sim \pi_{\theta_{k}}}\bigg[ \sum_{t = 0}^{\infty} \gamma^t \nabla_{\theta} r(s_t, a_t; \theta_{k}) \bigg]  -  \ee_{ \tau^{\rm A} \sim \pi_{k+1}}\bigg[ \sum_{t = 0}^{\infty} \gamma^t \nabla_{\theta} r(s_t, a_t; \theta_{k}) \bigg] \bigg \| \right]  \nonumber \\
		&\overset{(i)}{=}  \ee \left[ \bigg \|  \frac{1}{1 - \gamma} \ee_{ (s,a) \sim d(\cdot, \cdot; \pi_{\theta_k}) }\big[ \nabla_{\theta} r(s, a; \theta_{k}) \big]  -  \frac{1}{1 - \gamma} \ee_{(s,a) \sim d(\cdot, \cdot; \pi_{k+1})}  \big[ \nabla_{\theta} r(s, a; \theta_{k}) \big] \bigg \| \right]  \nonumber \\
		&= \frac{1}{1 - \gamma} \mathbb{E} \bigg[ \bigg  \| \sum_{s \in \mathcal{S}, a \in \mathcal{A}}  \nabla_{\theta} r(s_t, a_t; \theta_k) \bigg(d(s,a; \pi_{\theta_k}) - d(s,a; \pi_{k+1})\bigg) \bigg \| \bigg] \nonumber \\ 
		&\leq  \frac{1}{1 - \gamma} \mathbb{E} \bigg[ \sum_{s \in \mathcal{S}, a \in \mathcal{A}} \big \|  \nabla_{\theta} r(s_t, a_t; \theta_k) \big \| \cdot \big | d(s,a; \pi_{\theta_k}) - d(s,a; \pi_{k+1}) \big | \bigg]  \nonumber \\
		&\overset{(ii)}{\leq} \frac{2L_r}{1 - \gamma}   \ee \big[ \| d(\cdot, \cdot;\pi_{\theta_k}) - d(\cdot, \cdot;\pi_{k+1}) \|_{TV} \big] \nonumber \\
		&= 2L_q \ee \big[ \| d(\cdot, \cdot;\pi_{\theta_k}) - d(\cdot, \cdot;\pi_{k+1}) \|_{TV} \big] \nonumber \\
		&\overset{(iii)}{\leq}  2L_q C_d  \ee \left[  \|  Q_{\theta_k} - \widehat{Q}_{k}  \| \right]  \nonumber \\
		&\overset{(iv)}{\leq} 2L_q C_d \sqrt{|\mathcal{S} | \cdot | \mathcal{A} |}  \ee \left[  \|  Q_{\theta_k} - \widehat{Q}_{k}  \|_{\infty} \right] \nonumber \\
		&\leq  2L_q C_d \sqrt{|\mathcal{S} | \cdot | \mathcal{A} |}  \ee \left[ \epsilon_{\rm app} + \|  Q_{\theta_k} - Q_{k}  \|_{\infty} \right]
		\label{bound:trajectory_mismatch}
	\end{align}
	where (i) follows the definition of the state-action visitation measure $d(s,a;\pi) = (1-\gamma)\pi(a|s)\sum_{t = 0}^{\infty} \gamma^t P^{\pi}(s_t = s| s_0 \sim \rho)$; (ii) follows the inequality \eqref{ineq:Lipschitz_smooth_reward} in Assumption \ref{Assumption:reward_grad_bound} and the definition of the total variation norm $\| \cdot \|_{TV}$; (iii) follows the definition of the constant $L_q := \frac{L_r}{1 - \gamma}$; (iii) follows Lemma \ref{lemma:Lipschitz_visitation_measure} and the fact that $\pi_{\theta_k}(\cdot | s) \propto \exp \big(Q_{\theta_k}(s, \cdot) \big)$, $\pi_{k + 1}(\cdot | s) \propto \exp \big(\widehat{Q}_{k}(s, \cdot) \big)$; follows the conversion between Frobenius norm and infinity norm.
	
	Through plugging the inequality \eqref{bound:trajectory_mismatch} into \eqref{bound:objective_gradient_ascent}, it leads to 
	\begin{align}
		& \mathbb{E} \left[ \widehat{L}(\theta_{k+1};\mathcal{D}) \right] \nonumber \\
		&\geq  \mathbb{E} \left[ \widehat{L}(\theta_{k};\mathcal{D}) \right]  - 2\alpha L_q  \ee \left[ \bigg \| \ee_{ \tau^{\rm A} \sim \pi_{\theta_{k}}}\bigg[ \sum_{t = 0}^{\infty} \gamma^t \nabla_{\theta} r(s_t, a_t; \theta_{k}) \bigg]  -  \ee_{ \tau^{\rm A} \sim \pi_{k+1}}\bigg[ \sum_{t = 0}^{\infty} \gamma^t \nabla_{\theta} r(s_t, a_t; \theta_{k}) \bigg] \bigg \| \right] \nonumber \\ 
		& \quad + \alpha \mathbb{E} \bigg[ \|  \nabla_{\theta} \widehat{L}(\theta_{k};\mathcal{D}) \|^2 \bigg] - 2L_c L_q^2 \alpha^2 \nonumber \\
		&\overset{(i)}{\geq} \mathbb{E} \left[ \widehat{L}(\theta_{k};\mathcal{D}) \right] - 4\alpha C_d L_q^2 \sqrt{|\mathcal{S}| \cdot | \mathcal{A} |}  \ee \left[  \|  Q_{\theta_k} - Q_{k} \|_{\infty} + \epsilon_{\rm app} \right] + \alpha \mathbb{E} \bigg[ \|  \nabla_{\theta} \widehat{L}(\theta_{k};\mathcal{D}) \|^2 \bigg] - 2L_c L_q^2 \alpha^2 \nonumber
	\end{align}
	where (i) follows the inequality \eqref{bound:trajectory_mismatch}. 
	
	Denoting $C_1 := 4C_d L_q^2 \sqrt{|\mathcal{S}| \cdot |\mathcal{A}|}$ and rearranging the inequality above, it holds that 
	\begin{align}
		\alpha \mathbb{E} \big[ \|  \nabla_{\theta} \widehat{L}(\theta_{k};\mathcal{D}) \|^2 \big] \leq 2L_c L_q^2 \alpha^2 +  \alpha C_1 \ee \left[ \|  Q_{\theta_{k}} -  Q_{k}  \|_{\infty} + \epsilon_{app} \right] + \ee \big[ \widehat{L}(\theta_{k+1};\mathcal{D}) - \widehat{L}(\theta_{k};\mathcal{D}) \big] \nonumber \nonumber
	\end{align} 
	Summing the inequality above from $k = 0$ to $K-1$ and dividing both sides by $\alpha K$, it holds that 
	\begin{align}
		\frac{1}{K}\sum_{k = 0}^{K-1} \ee \left[ \|  \nabla_{\theta} \widehat{L}(\theta_{k};\mathcal{D}) \|^2 \right] \leq 2L_c L_q^2 \alpha + \frac{C_1}{K} \sum_{k = 0}^{K-1} \ee \left[ \|  Q_{\theta_{k}} -  Q_{k} \|_{\infty} + \epsilon_{\rm app} \right] + \ee \left[ \frac{\widehat{L}(\theta_{K};\mathcal{D}) - \widehat{L}(\theta_{0};\mathcal{D})}{K \alpha} \right] \nonumber
	\end{align}
	According to Assumption \ref{assumption:bound_reward}, we assume that the reward function is bounded. Based on this assumption, we know that the empirical estimation objective $\widehat{L}(\cdot; \mathcal{D})$ is bounded. Then we could plug \eqref{convergence_rate:soft_Q_gap} into the inequality above, we obtain \begin{align}
		\frac{1}{K}\sum_{K = 0}^{K-1} \ee \left[ \|  \nabla_{\theta} \widehat{L}(\theta_{K};\mathcal{D}) \|^2 \right]  = \mathcal{O}(K^{-\sigma}) + \mathcal{O}(K^{-1}) + \mathcal{O}(K^{-1 + \sigma}) + \mathcal{O}(\epsilon_{\rm app}) \label{rate_analysis:lower_problem}.
	\end{align}
	This completes the proof of this result.
\hfill $\blacksquare$

\section{Proof of Theorem \ref{theorem:duality}}
\label{proof:theorem:duality}
{In this section, we prove the duality between the estimation problem \eqref{ML:estimation} and the maximum entropy IRL problem \eqref{eq:max_ent}. To state the proof, we first write down the {\it partial} Lagrangian function, when only dualizing the constraint \eqref{constraint:feature} - \eqref{constraint:prob_sum}. After we derive the dual form for the problem with constraint \eqref{constraint:feature} - \eqref{constraint:prob_sum}, we will make sure that the constraint \eqref{constraint:prob_positive} is satisfied.}

Let $\theta$ and $C_{s_t}$  denote the dual variables of the constraints \eqref{constraint:feature}, and \eqref{constraint:prob_sum}, respectively; define $\phi(\pi^{\rm E}; \mathcal{D}) := \mathbb{E}_{\tau^{\rm E} \sim \mathcal{D}} \big[ \sum_{t = 0}^{\infty} \gamma^t \phi(s_t, a_t) \big]$. Then the partial Lagrangian can be expressed as:
\begin{align}
		\mathcal{L}(\pi,\theta) &:= - \mathbb{E}_{ \tau^{\rm A} \sim \pi} \bigg[  \sum_{t = 0}^{\infty} \gamma^t  \log \pi(a_t | s_t) \bigg]  + \theta^{\top} \bigg( \mathbb{E}_{ \tau^{\rm A} \sim \pi} \bigg[ \sum_{t = 0}^{\infty} \gamma^t \phi(s_t, a_t) \bigg] - \phi(\pi^{\rm E}; \mathcal{D}) \bigg) \nonumber \\
		& \quad \quad + \sum_{t\geq0, s_t \in \mathcal{S}} C_{s_t} \bigg(\sum_{a \in \mathcal{A}} \pi(a| s_t) - 1\bigg)	\label{eq:Lagrangian_maxent}.
	\end{align} 
	Our plan is to show that the dual function, as defined by $\bar{\mathcal{L}}(\theta):= \max_{\pi} ~ 	\mathcal{L}(\pi,\theta)$, has the following expression: 
	\begin{align}
	 \bar{\mathcal{L}}(\theta) =  \mathbb{E}_{s_0 \sim \rho} \Big[ V_{\theta}(s_0) \Big] - \mathbb{E}_{\tau^{\rm E} \sim \mathcal{D}} \bigg[ \sum_{t = 0}^{\infty} \gamma^t r(s_t, a_t; \theta) \bigg],
	\end{align}
	so that the dual problem can be shown to be equivalent to problem \eqref{eq:max_ent}, as follows:
		\begin{align}
		\min_{\theta} ~ \bar{\mathcal{L}}(\theta)  =  \min_{\theta} ~ \mathbb{E}_{s_0 \sim \rho} \Big[ V_{\theta}(s_0) \Big] - \mathbb{E}_{\tau^{\rm E} \sim \mathcal{D}} \bigg[ \sum_{t = 0}^{\infty} \gamma^t r(s_t, a_t; \theta) \bigg] = \max_{\theta} ~ \mathbb{E}_{\tau^{\rm E} \sim \mathcal{D}} \bigg[ \sum_{t = 0}^{\infty} \gamma^t r(s_t, a_t; \theta) \bigg] - \mathbb{E}_{s_0 \sim \rho} \Big[ V_{\theta}(s_0) \Big]. \nonumber
	\end{align}
	
	Towards this end, let us compute the gradient of $\mathcal{L}(\pi,\theta)$ with respect to the policy $\pi(a|s_t = s)$:
	\begin{align}
		&\quad \nabla_{\pi(a|s_t = s)} ~ \mathcal{L}(\pi,\theta) \nonumber \\
		&= \nabla_{\pi(a|s_t = s)} \bigg( - \mathbb{E}_{ \tau^{\rm A} \sim \pi} \bigg[  \sum_{\kappa = 0}^{\infty} \gamma^{\kappa}  \log \pi(a_{\kappa} | s_{\kappa}) \bigg]  + \theta^{\top} \mathbb{E}_{ \tau^{\rm A} \sim \pi} \bigg[ \sum_{\kappa = 0}^{\infty} \gamma^{\kappa} \phi(s_{\kappa}, a_{\kappa}) \bigg] \bigg) \nonumber \\
		& \quad \quad + \nabla_{\pi(a|s_t = s)} \bigg( -\theta^{\top} \phi(\pi^{\rm E}; \mathcal{D}) + \sum_{\kappa\geq0, s \in \mathcal{S}} C_{s_{\kappa} = s} \Big(\sum_{a \in \mathcal{A}} \pi(a| s_{\kappa}) - 1\Big) \bigg) \nonumber \\
		&\overset{(i)}{=} \nabla_{\pi(a|s_t = s)} \bigg( - \mathbb{E}_{ \tau^{\rm A} \sim \pi} \bigg[ \sum_{\kappa = t}^{\infty} \gamma^{\kappa}  \log \pi(a_{\kappa} | s_{\kappa}) \bigg] + \theta^{\top} \mathbb{E}_{ \tau^{\rm A} \sim \pi} \bigg[ \sum_{\kappa = t}^{\infty} \gamma^t \phi(s_t, a_t) \bigg] \bigg) + C_{s_t = s} \nonumber \\
		&=  \nabla_{\pi(a|s_t = s)} \bigg( -\sum_{s \in \mathcal{S}, a \in \mathcal{A}} P^{\pi}(s_t = s| s_0 \sim \rho)\pi(a| s_t = s) \mathbb{E}_{\tau^{\rm A} \sim \pi} \Big[  \sum_{\kappa = t}^{\infty} \gamma^{\kappa  }\log \pi(a_{\kappa}|s_{\kappa}) \Big| (s_t, a_t) = (s, a) \Big] \bigg)    \nonumber \\
		&\quad \quad + \nabla_{\pi(a|s_t = s)} \bigg( \sum_{s \in \mathcal{S}, a \in \mathcal{A}} P^{\pi}(s_t = s| s_0 \sim \rho)\pi(a| s_t = s) \mathbb{E}_{\tau^{\rm A} \sim \pi} \Big[  \sum_{\kappa = t}^{\infty} \gamma^{\kappa  } \theta^{\top} \phi(s_{\kappa}, a_{\kappa}) \Big| (s_t, a_t) = (s, a) \Big] \bigg) + C_{s_t = s} \nonumber \\
		&= P^{\pi}(s_t = s | s_0 \sim \rho)\bigg( -\gamma^t \big(\log \pi(a|s_t = s) + 1 \big) + \mathbb{E}_{\tau^{\rm A} \sim \pi} \Big[ \sum_{\kappa = t}^{\infty} - \gamma^{\kappa+1} \log \pi(a_{\kappa+1} | s_{\kappa+1}) \Big| (s_t, a_t) = (s, a) \Big] \nonumber \\
		& \quad \quad + \mathbb{E}_{\tau^{\rm A} \sim \pi} \Big[ \sum_{\kappa = t}^{\infty} \gamma^{\kappa} \theta^{\top} \phi(s_\kappa, a_\kappa) \Big| (s_t, a_t) = (s, a) \Big] \bigg) + C_{s_t = s} \label{grad:Lagrangian}
	\end{align}
	where (i) follows the fact that the probability $\pi(a|s_t = s)$ has no effect on the trajectory generated before time $t$. Setting $\nabla_{\pi(a|s_t = s)} ~ \mathcal{L}(\pi,\theta) = 0$, we obtain the following first-order condition:
	\begin{align*}
		\quad \log \pi(a|s_t = s) & =
		 \bigg( \frac{C_{s_t = s}}{\gamma^t \cdot P^{\pi}(s_t = s| s_0 \sim \rho)} - 1 \bigg) - \mathbb{E}_{\tau^{\rm A} \sim \pi} \bigg[ \sum_{\kappa = t}^{\infty} \gamma^{\kappa + 1 - t} \log \pi(a_{\kappa + 1} | s_{\kappa + 1}) \Big| (s_t, a_t) = (s, a) \bigg]   \\
		 &~~~~ + \mathbb{E}_{\tau^{\rm A} \sim \pi} \bigg[ \sum_{\kappa = t}^{\infty} \gamma^{\kappa - t} \theta^{\top} \phi(s_{\kappa}, a_{\kappa}) \Big| (s_t, a_t) = (s, a) \bigg].
	\end{align*}
	 Then we could express $\pi(a|s_t = s)$ as below:
	  \begin{align}
		\pi(a|s_t = s) &= \exp\bigg(  -\mathbb{E}_{\tau^{\rm A} \sim \pi} \bigg[ \sum_{\kappa = t}^{\infty} \gamma^{\kappa + 1 - t} \log \pi(a_{\kappa+1} | s_{\kappa+1}) \Big| (s_t, a_t) = (s, a) \bigg] \nonumber \\
		& \quad + \mathbb{E}_{\tau^{\rm A} \sim \pi} \bigg[ \sum_{\kappa = t}^{\infty} \gamma^{\kappa - t} \theta^{\top}  \phi(s_{\kappa}, a_{\kappa}) \Big| (s_t, a_t) = (s, a) \bigg] + \frac{C_{s_t = s}}{\gamma^t \cdot P^{\pi}(s_t = s)} - 1\bigg). \label{stationarity}
	\end{align}
	Note that $\big( \frac{C_{s_t = s}}{\gamma^t \cdot P^{\pi}(s_t = s| s_0 \sim \rho)} - 1 \big)$ is independent of the action $a_t$. Hence, the following result holds:
	\begin{align}
	    \pi(a|s_t = s) \propto \exp\bigg(  \mathbb{E}_{\tau^{\rm A} \sim \pi} \bigg[ \sum_{\kappa = t}^{\infty} \gamma^{\kappa - t} \big( \theta^T \phi(s_{\kappa}, a_{\kappa}) - \gamma \log \pi(a_{\kappa+1} | s_{\kappa+1}) \big) \Big| (s_t, a_t) = (s, a) \bigg] \bigg) \nonumber \\
	    = \exp\bigg(  \mathbb{E}_{\tau^{\rm A} \sim \pi} \bigg[ \sum_{\kappa = 0}^{\infty} \gamma^{\kappa} \big( \theta^{\top} \phi(s_{\kappa}, a_{\kappa}) - \gamma \log \pi(a_{\kappa+1} | s_{\kappa+1}) \big) \Big| (s_0, a_0) = (s, a) \bigg] \bigg) \label{derive:policy_expression}
	\end{align}
	According to \eqref{derive:policy_expression}, we could conclude that $ \pi(a|s_t = s) $ only depends on the state-action pair $(s,a)$ and is independent of the time index $t\geq 0$. Hence, we have shown that the policy $\pi$ is a stationary policy and $\pi(a|s_t = s) = \pi(a|s)$ for any $t \geq 0$. 
	
	Therefore, we can rewrite \eqref{derive:policy_expression} with $t=0$ as follows:
	\begin{align}
		\pi(a|s) &\propto \exp\bigg(  \mathbb{E}_{\tau^{\rm A} \sim \pi} \bigg[ \sum_{\kappa = 0}^{\infty} \gamma^{\kappa} \big( \theta^T \phi(s_{\kappa}, a_{\kappa}) - \gamma \log \pi(a_{\kappa+1} | s_{\kappa+1}) \big) \Big| (s_0, a_0) = (s, a) \bigg] \bigg) \nonumber \\
		&\overset{(i)}{=} \exp\bigg( r(s_0, a_0; \theta)
		+ \mathbb{E}_{\tau^{\rm A} \sim \pi} \bigg[ \sum_{\kappa = 0}^{\infty} \gamma^{\kappa+1} \big(r(s_{\kappa+1}, a_{\kappa+1}; \theta)- \log \pi(a_{\kappa+1} | s_{\kappa+1}) \big)  \Big| (s_0, a_0) = (s, a) \bigg] \bigg) \nonumber \\
		&\overset{(ii)}{=} \exp \big( Q^{\pi}(s,a) \big) \label{duality:policy_expression}
	\end{align}
	where (i) follows the linear approximation of the reward function that $r(s,a;\theta) := \theta^T \phi(s,a)$. Clearly, the right hand side of (i) is the soft Q-function under reward parameter $\theta$ and the stationary policy $\pi$, therefore in (ii) we use $Q^{\pi}(s,a)$ to denote such a soft Q-function.
	
	Recall that we have defined $V_{\theta}$, $Q_{\theta}$ as the soft value function, soft Q-function under reward parameter $\theta$ and the optimal policy $\pi_{\theta}$. For any $s \in \mathcal{S}$ and $a \in \mathcal{A}$, it follows that
\begin{subequations}
    \begin{align}
    &V_{\theta}(s) := \mathbb{E}_{\tau^{\rm A} \sim \pi_{\theta}} \bigg[ \sum_{t = 0}^{\infty} \gamma^t \bigg( r(s_t, a_t;\theta) + \mathcal{H}(\pi_{\theta}(\cdot | s_t)) \bigg) \bigg | s_0 = s \bigg], \label{def:optimal_soft_V_closed_form} \\
    &Q_{\theta}(s,a) := r(s,a; \theta) + \gamma \mathbb{E}_{s^\prime \sim P(\cdot| s,a)} \big[ V_{\theta}(s^\prime) \big]. \label{def:optimal_soft_Q_expression}
\end{align}
\end{subequations}
	According to \cite{haarnoja2017reinforcement} and \cite{cen2021fast}, the the optimal policy $\pi_{\theta}$ in the entropy-regularized MDP satisfy the following expression for any $s \in \mathcal{S}$ and $a \in \mathcal{A}$: 
\begin{subequations}
    \begin{align}
    \pi_{\theta}(a|s) = \frac{\exp \big( Q_{\theta}(s,a) \big)}{\sum_{\tilde{a} \in \mathcal{A}} \exp \big( Q_{\theta}(s,\tilde{a}) \big)} \label{eq:policy_closed_form_expression}
\end{align}
\end{subequations}
Therefore, we know the policy in \eqref{duality:policy_expression} is the optimal policy $\pi_{\theta}$. Using $\pi_{\theta}$ to replace the policy $\pi$ in the Lagrangian function $L(\pi, \theta)$ as given by \eqref{eq:Lagrangian_maxent}, we can express the dual function as {\small
	\begin{align}
		& \quad \bar{\mathcal{L}}(\theta)  \nonumber \\
		&= - \mathbb{E}_{ \tau^{\rm A} \sim \pi_{\theta}} \bigg[  \sum_{t = 0}^{\infty} \gamma^t  \log \pi_{\theta}(a_t | s_t) \bigg]  + \theta^{\top} \bigg( \mathbb{E}_{ \tau^{\rm A} \sim \pi_{\theta}} \bigg[ \sum_{t = 0}^{\infty} \gamma^t \phi(s_t, a_t) \bigg] - \mathbb{E}_{\tau^{\rm E} \sim \mathcal{D}} \bigg[ \sum_{t = 0}^{\infty} \gamma^t \phi(s_t, a_t) \bigg] \bigg) \nonumber \\
		&\quad \quad + \sum_{t\geq0, s_t \in \mathcal{S}} C_{s_t} \bigg(\sum_{a \in \mathcal{A}} \pi_{\theta}(a| s_t) - 1\bigg) \nonumber \\
		&\overset{(i)}{=}  -  \mathbb{E}_{ \tau^{\rm A} \sim \pi_{\theta}} \bigg[ \sum_{t = 0}^{\infty} \gamma^t \log \bigg( \frac{\exp Q_{\theta}(s_t, a_t)}{\sum_{a \in \mathcal{A}} \exp Q_{\theta}(s_t, a)} \bigg) \bigg] +  \mathbb{E}_{ \tau^{\rm A} \sim \pi_{\theta}} \bigg[ \sum_{t = 0}^{\infty} \gamma^t r(s_t, a_t; \theta)  \bigg] - \mathbb{E}_{\tau^{\rm E} \sim \mathcal{D}} \bigg[ \sum_{t = 0}^{\infty} \gamma^t r(s_t, a_t; \theta) \bigg]  \nonumber \\
		&= -  \mathbb{E}_{ \tau^{\rm A} \sim \pi_{\theta}} \bigg[ \sum_{t = 0}^{\infty} \gamma^t \bigg( Q_{\theta}(s_t, a_t) - \log \big( \sum_{a \in \mathcal{A}} \exp Q_{\theta}(s_t, a) \big) \bigg) \bigg]  +  \mathbb{E}_{ \tau^{\rm A} \sim \pi_{\theta}} \bigg[ \sum_{t = 0}^{\infty} \gamma^t r(s_t, a_t; \theta)  \bigg] \nonumber \\
		& \quad \quad - \mathbb{E}_{\tau^{\rm E} \sim \mathcal{D}} \bigg[ \sum_{t = 0}^{\infty} \gamma^t r(s_t, a_t; \theta) \bigg] \nonumber \\
		&\overset{(ii)}{=} -  \mathbb{E}_{ \tau^{\rm A} \sim \pi_{\theta}} \bigg[ \sum_{t = 0}^{\infty} \gamma^t \bigg( r(s_t, a_t; \theta) + \gamma V_{\theta}(s_{t+1}) - V_{\theta}(s_t) \bigg) \bigg] +  \mathbb{E}_{ \tau^{\rm A} \sim \pi_{\theta}} \bigg[ \sum_{t = 0}^{\infty} \gamma^t r(s_t, a_t; \theta)  \bigg] - \mathbb{E}_{\tau^{\rm E} \sim \mathcal{D}} \bigg[ \sum_{t = 0}^{\infty} \gamma^t r(s_t, a_t; \theta) \bigg] \nonumber \\
		&=  -  \mathbb{E}_{ \tau^{\rm A} \sim \pi_{\theta}} \bigg[ \sum_{t = 0}^{\infty} \gamma^t \bigg( \gamma V_{\theta}(s_{t+1}) - V_{\theta}(s_t) \bigg) \bigg] - \mathbb{E}_{\tau^{\rm E} \sim \mathcal{D}} \bigg[ \sum_{t = 0}^{\infty} \gamma^t r(s_t, a_t; \theta) \bigg] \nonumber \\
		&= \mathbb{E}_{s_0 \sim \rho} \big[ V_{\theta}(s_0) \big] -  \mathbb{E}_{\tau^{\rm E} \sim \mathcal{D}} \bigg[ \sum_{t = 0}^{\infty} \gamma^t r(s_t, a_t; \theta) \bigg]
		\label{dual:theta}
	\end{align}}
	where (i) follows the fact that $\pi_{\theta}(a_t | s_t) = \frac{\exp Q_{\theta}(s_t, a_t)}{\sum_{a \in \mathcal{A}}\exp Q_{\theta}(s_t, a)} $ (see \eqref{eq:policy_closed_form_expression}) and $r(s,a;\theta) := \theta^T \phi(s,a)$ ; (ii) follows \eqref{def:optimal_soft_Q_expression} and \eqref{def:optimal_soft_V_closed_form}. 
 Then we can show the equivalence between \eqref{dual:theta} and \eqref{ML:estimation:surrogate}:
 {\small
	\begin{align}
		\min_{\theta} ~ \bar{\mathcal{L}}(\theta)  =  \min_{\theta} ~ \mathbb{E}_{s_0 \sim \rho} \big[ V_{\theta}(s_0) \big] -  \mathbb{E}_{\tau^{\rm E} \sim \mathcal{D}} \bigg[ \sum_{t = 0}^{\infty} \gamma^t r(s_t, a_t; \theta) \bigg] = \max_{\theta} ~ \mathbb{E}_{\tau^{\rm E} \sim \mathcal{D}} \bigg[ \sum_{t = 0}^{\infty} \gamma^t r(s_t, a_t; \theta) \bigg] - \mathbb{E}_{s_0 \sim \rho} \big[ V_{\theta}(s_0) \big].\nonumber
	\end{align}}
	 {Hence, we have proved that  \eqref{ML:estimation:surrogate} - \eqref{ML:estimation:constraint} is the dual form of \eqref{obj:max_ent} - \eqref{constraint:prob_sum} and the constraint \eqref{constraint:prob_positive} is satisfied due to the closed form of the optimal policy $\pi_{\theta}$ in \eqref{eq:policy_closed_form_expression}.
	
	Note the objective \eqref{obj:max_ent} is concave and \eqref{constraint:feature}, \eqref{constraint:prob_sum}  are affine. In addition, the interior of the feasible region is not empty (i.e Slater's condition). Hence, under linear parameterization of the reward function, there is strong duality (no gap) between the solutions of \eqref{ML:estimation} and \eqref{eq:max_ent}.

 When the expert policy is known or available for access, following the derivations in \eqref{dual:theta}, we show the dual problem of the maximum entropy estimation problem (\eqref{obj:max_ent},\eqref{max_ent_infinite},\eqref{constraint:prob_sum},\eqref{constraint:prob_positive}) follows:
 \begin{subequations}
    \begin{align}
	\max_{\theta} 	&~~~~ \mathbb{E}_{\tau^{\rm E} \sim \pi^{\rm E}} \bigg[ \sum_{t = 0}^{\infty} \gamma^t r(s_t, a_t; \theta) \bigg] - \mathbb{E}_{s_0 \sim \rho} \Big[ V_{\theta}(s_0) \Big] \label{obj:value_gap:expectation} \\
	{\rm s.t} &~~~~ 
\pi_{\theta}(a_t|s_t)
:= \arg \max_{\pi} ~ \mathbb{E}_{s_0 \sim \rho, \tau^{\rm A} \sim \pi} \bigg[ \sum_{t = 0}^{\infty} \gamma^t \bigg( r(s_t, a_t; \theta) + \mathcal{H}(\pi(\cdot | s_t)) \bigg) \bigg].  \nonumber
\end{align}
\end{subequations}
Then based on our derivations in \eqref{rewrite:maximum_likelihood}, we obtain the equivalence between \eqref{obj:value_gap:expectation} and \eqref{eq:ML}:
\begin{align}
    \mathbb{E}_{\tau^{\rm E} \sim \pi^{\rm E}} \bigg[ \sum_{t = 0}^{\infty} \gamma^t r(s_t, a_t; \theta) \bigg] - \mathbb{E}_{s_0 \sim \rho} \Big[ V_{\theta}(s_0) \Big] = \mathbb{E}_{\tau^{\rm E} \sim \pi^{\rm E}} \bigg[ \sum_{t = 0}^{\infty} \gamma^t \ln \pi_{\theta}(a_t | s_t)  \bigg]. \nonumber
\end{align}
Therefore, we obtain the duality between the maximum likelihood estimation problem \eqref{eq:ML:problem} and the maximum entropy estimation problem (\eqref{obj:max_ent},\eqref{max_ent_infinite},\eqref{constraint:prob_sum},\eqref{constraint:prob_positive}).}
	\hfill $\blacksquare$




\bibliographystyle{IEEEtran} 
\bibliography{PaperBIB,ref_inverse} 


	
\clearpage

\section*{Electronic Companions}

\section{Supplementary Experiment}

\begin{figure}[H]
    \centering
    \includegraphics[width=0.4\textwidth]{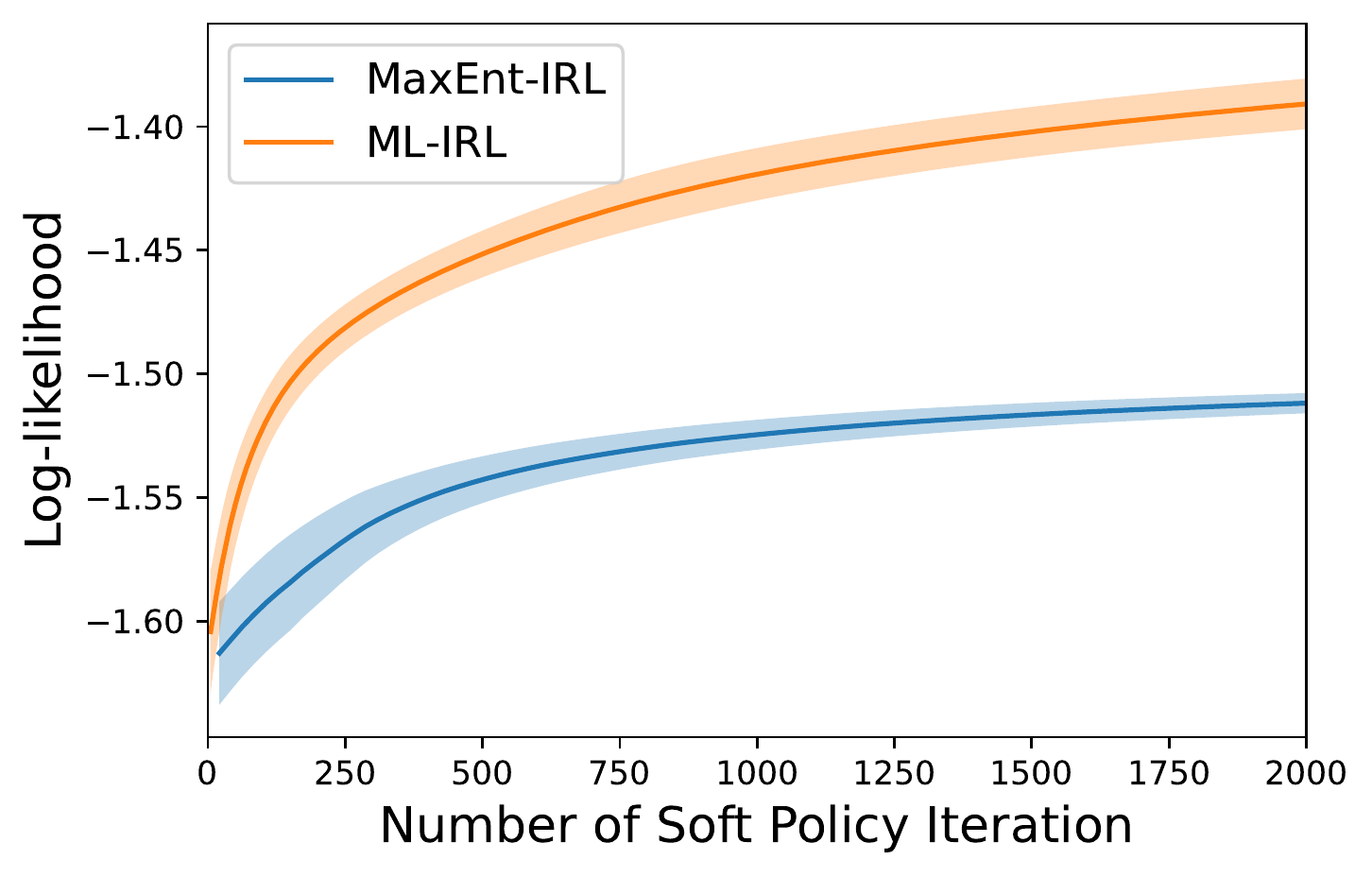}
    \includegraphics[width=0.4\textwidth]{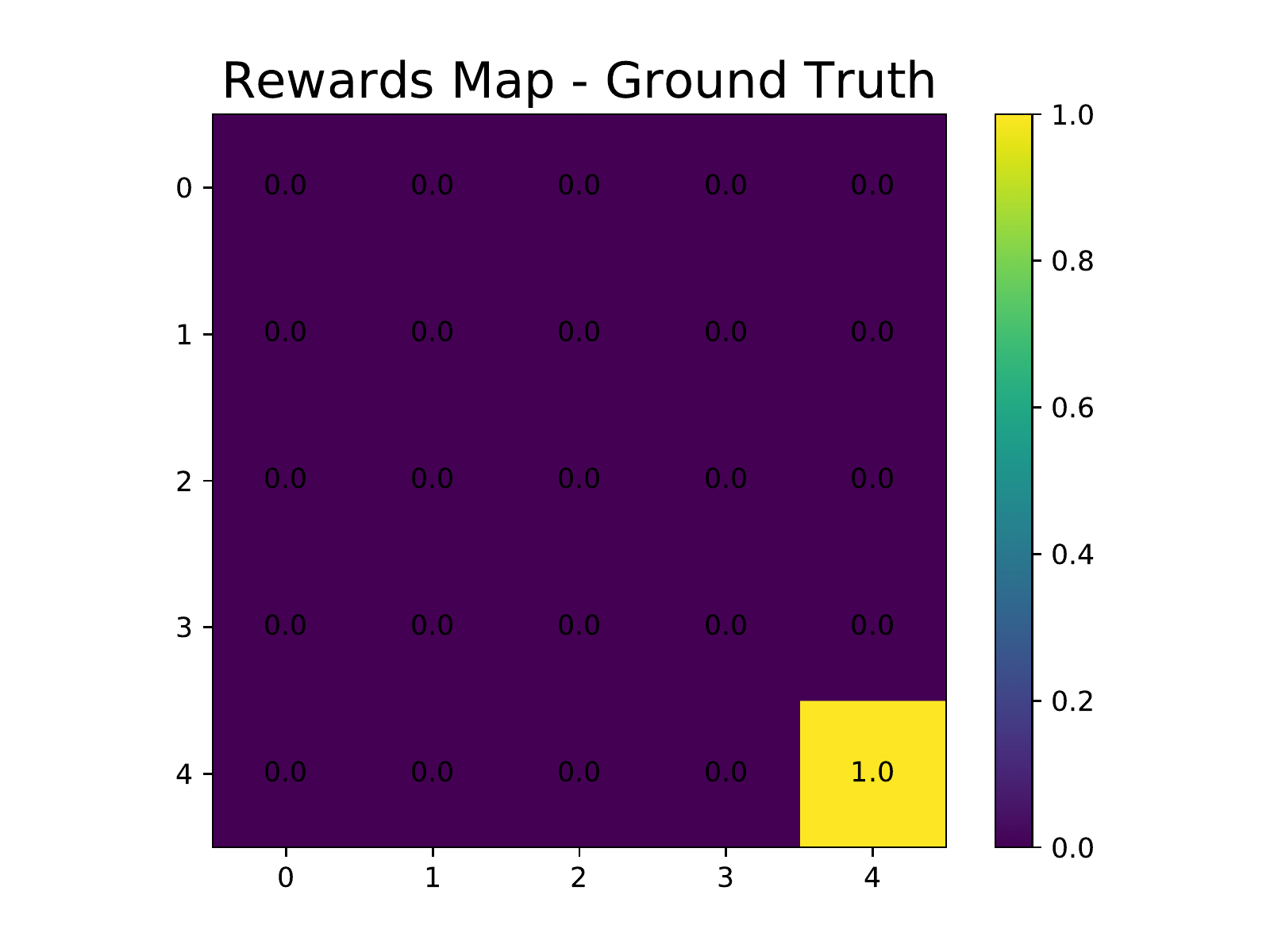}
    \includegraphics[width=0.4\textwidth]{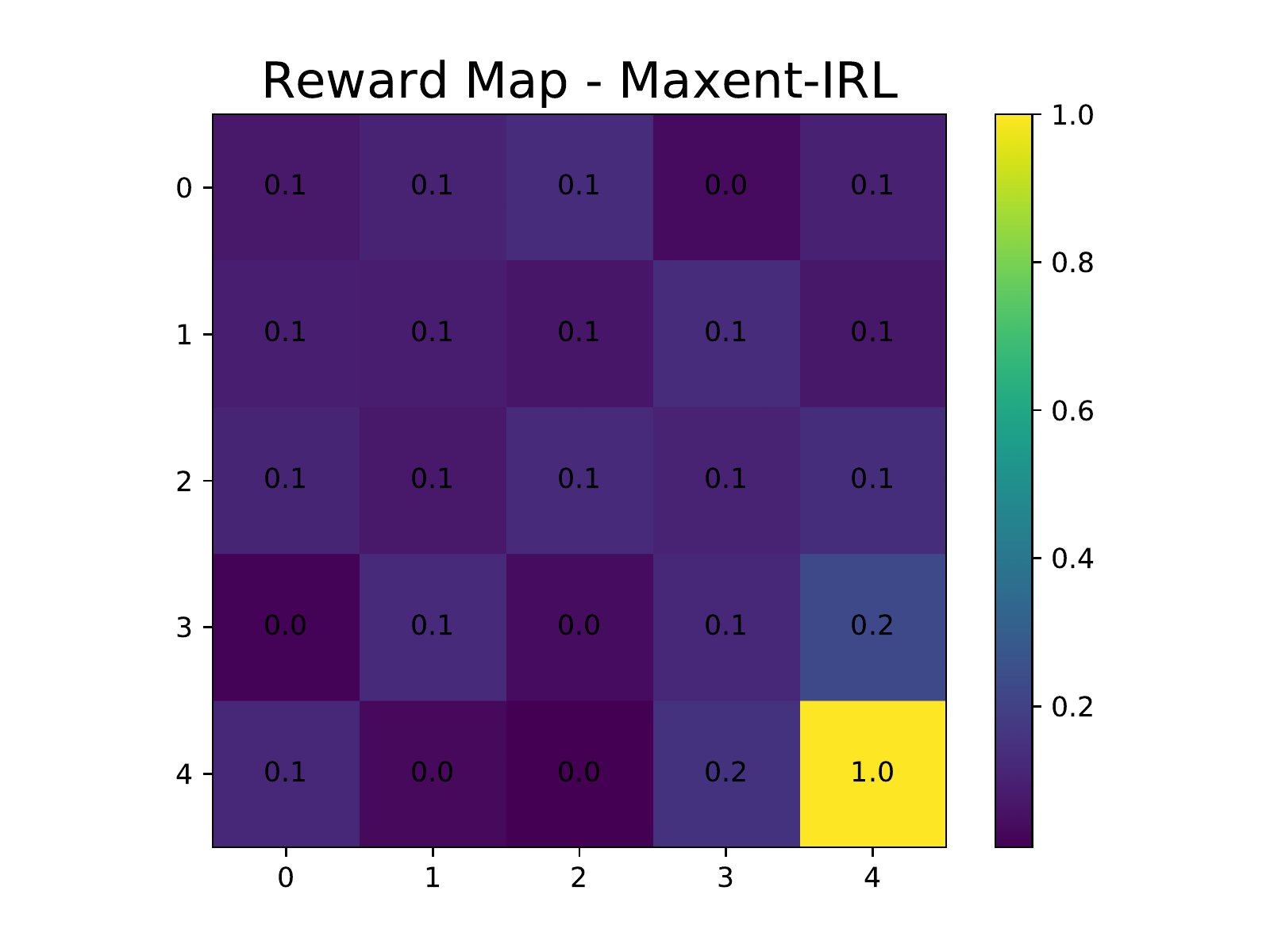}
    \includegraphics[width=0.4\textwidth]{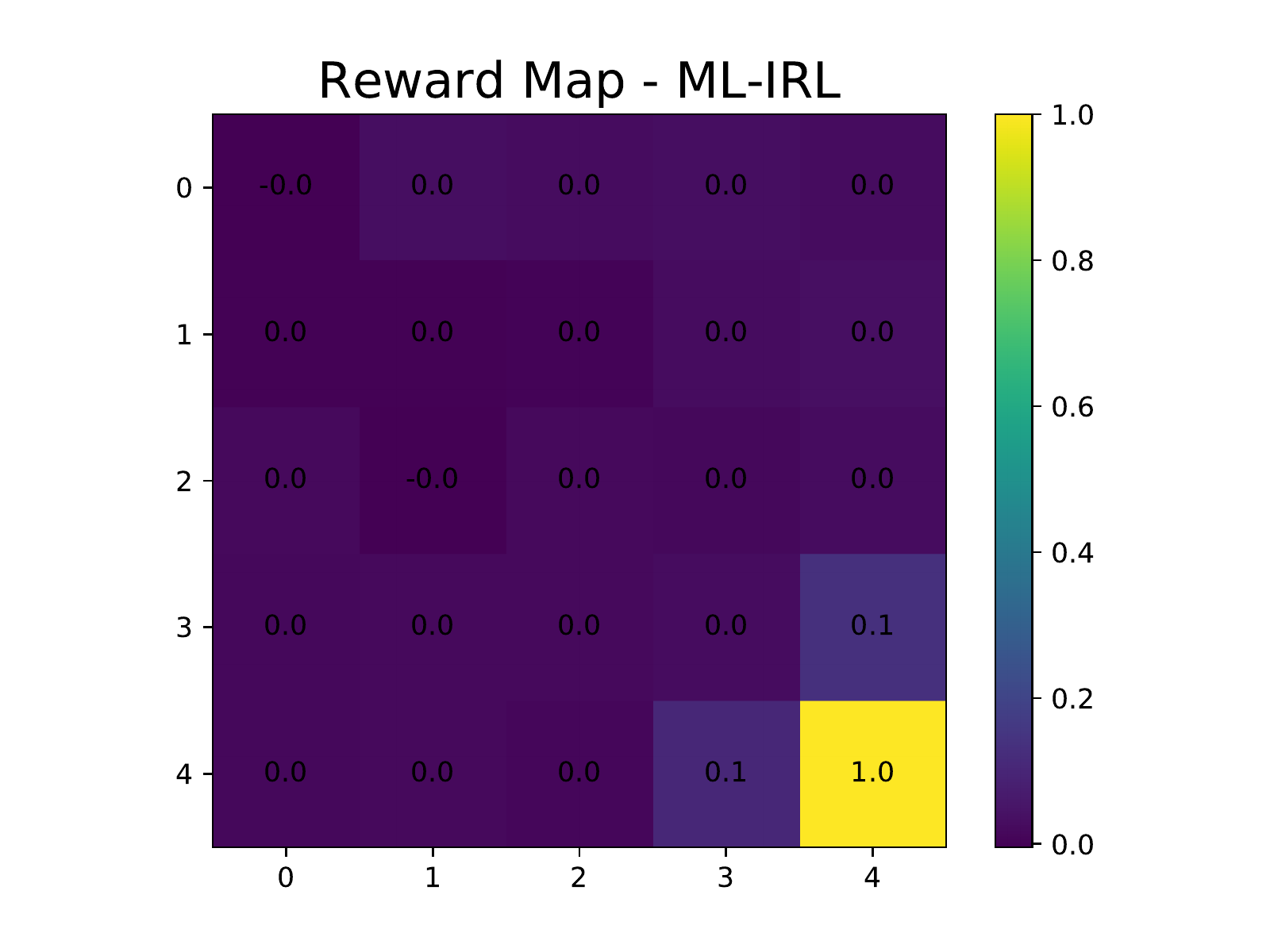}
    \caption{{\bf Tabular
Grid World.} We use a discrete GridWorld environment with 5 possible actions: up, down,
left, right, stay. Agent starts in a random state. (With 30 expert demos).}
    \label{experiment:gridworld}
\end{figure}

{\bf Reward Recovering on a Tabular
Grid World.} In order to validate the proposed algorithm as a method for IRL and show we recover correct rewards, we test our algorithm on a tabular Grid world setting, by using an open-source implementation\footnote{https://github.com/yrlu/irl-imitation}. The classical method \cite{ziebart2008maximum} requires repeated backward and forward passes, to calculate soft-values and action probabilities under a given reward and optimize the rewards respectively. By using a single-loop algorithm structure, our proposed algorithm could avoid the expensive backward pass in optimizing the policy under each given reward without compromising reward estimation accuracy. In Figure \ref{experiment:gridworld}, we visualize our
recovered rewards in the discrete GridWorld environment. According to Fig. 1, we show that ML-IRL converges much faster compared with MaxEnt-IRL while achieving similar accuracy on the recovered reward function.

\begin{figure}[H]
    \centering
    \includegraphics[width=0.45\textwidth]{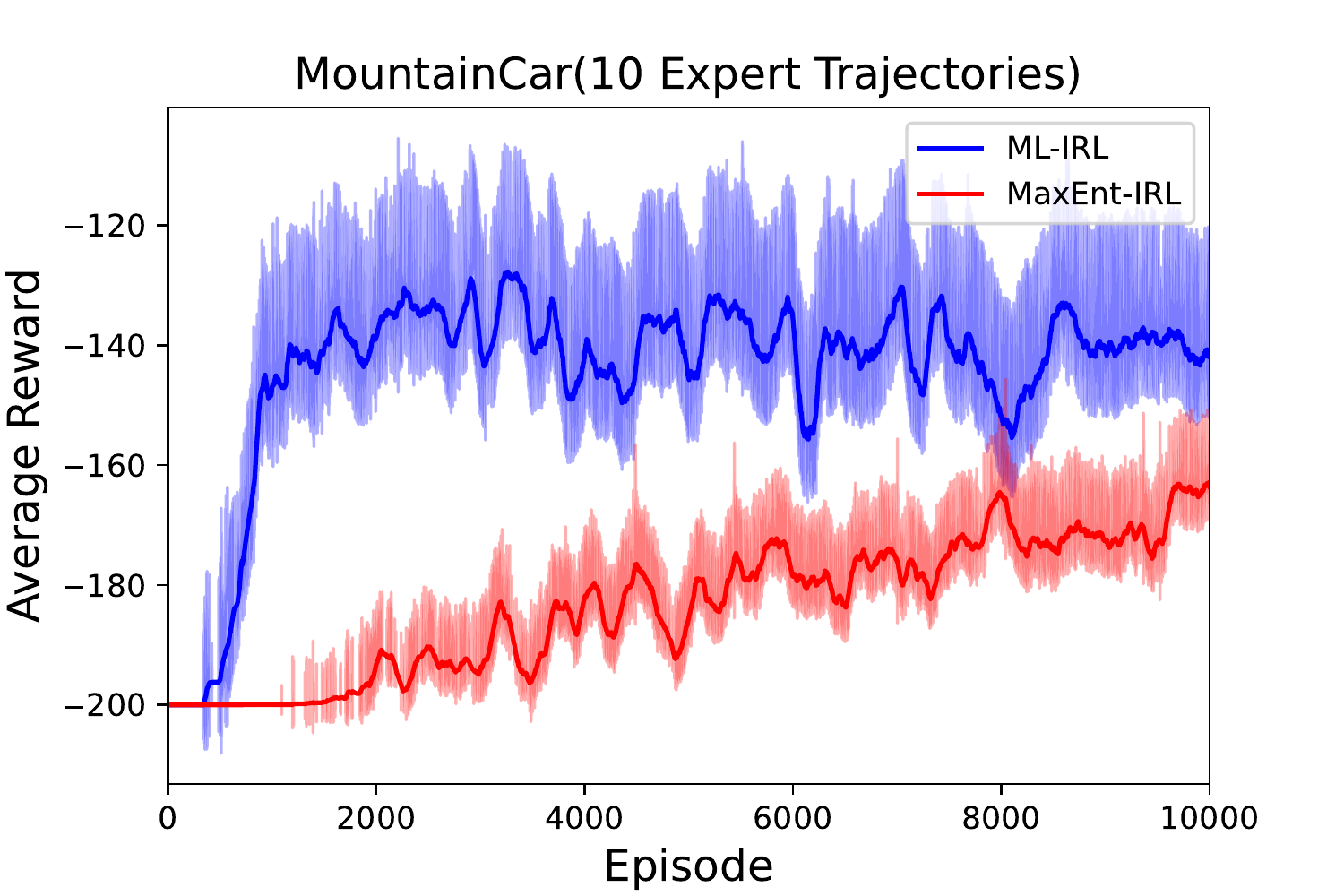}
    \includegraphics[width=0.45\textwidth]{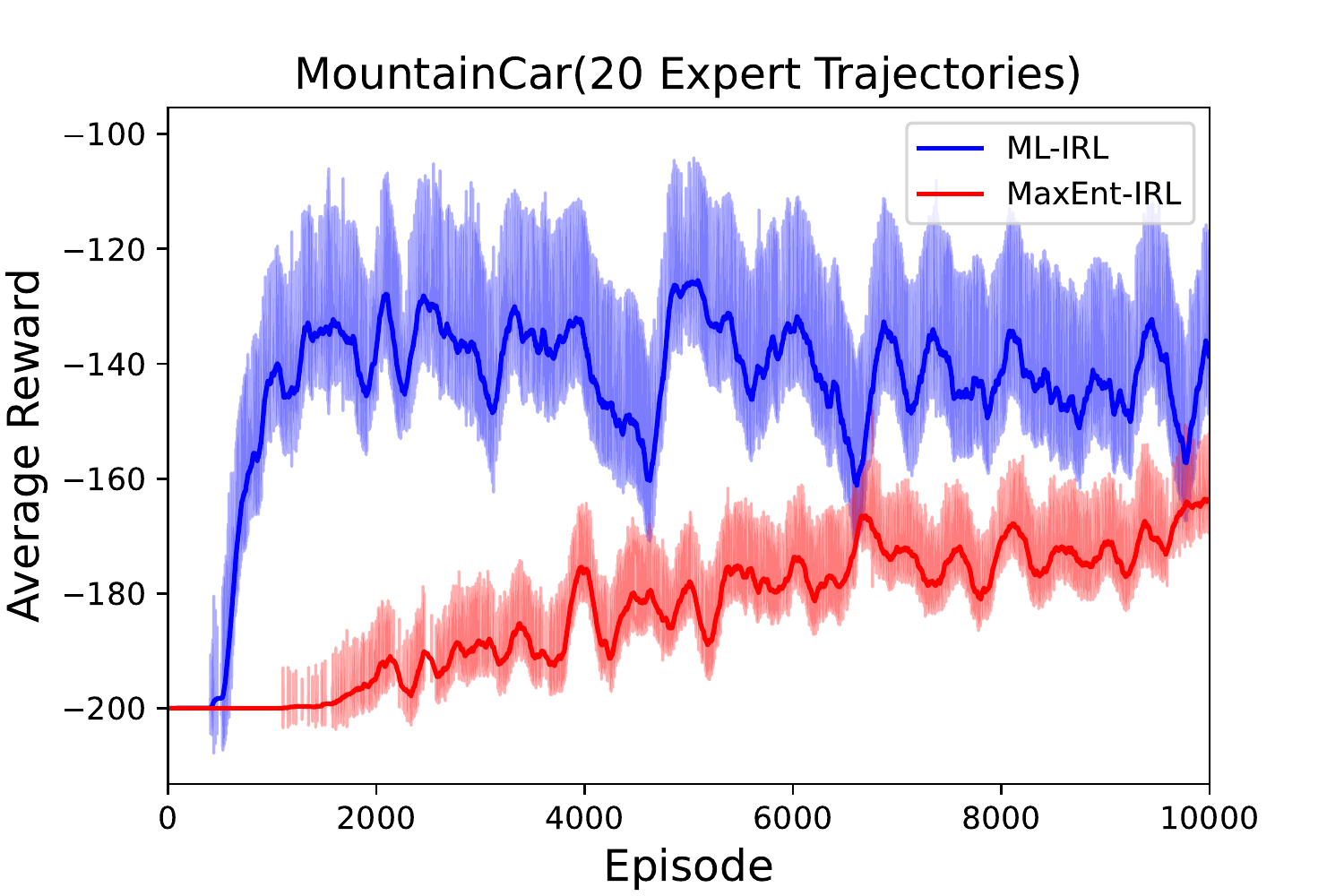}
    \caption{{\bf Mountain Car.} We compare ML-IRL with MaxEnt-IRL under different numbers of expert trajectories.}
    \label{experiment:cartpole}
\end{figure}

{\bf Inverse Reinforcement Learning on Mountain Car.} To further demonstrate the superiority of ML-IRL over MaxEnt-IRL, we test our algorithm on the classic reinforcement learning task -- Mountain Car. According to Fig. \ref{experiment:cartpole}, we show that ML-IRL is able to achieve faster convergence through leveraging the alternating updates between the policy and the reward estimator.

{
\section{Proof of Lemma \ref{lemma:objective_concentration}}
\label{proof:surrogate_objective_approximation}
Recall that in \eqref{rewrite:maximum_likelihood}, the likelihood objective $L(\theta)$ can be expressed as:
\begin{align}
    L(\theta) = \mathbb{E}_{\tau^{\rm E} \sim \pi^{\rm E}} \bigg[ \sum_{t = 0}^{\infty} \gamma^t r(s_t, a_t; \theta) \bigg] - \mathbb{E}_{s_0 \sim \rho} \Big[ V_{\theta}(s_0) \Big].  \nonumber
\end{align}
Moreover, given a dataset of collected expert trajectories, we have defined the estimation problem $\widehat{L}(\theta;\mathcal{D})$ as below:
\begin{align}
    \widehat{L}(\theta;\mathcal{D}) = \mathbb{E}_{\tau^{\rm E} \sim \mathcal{D}} \bigg[ \sum_{t = 0}^{\infty} \gamma^t r(s_t, a_t; \theta) \bigg] - \mathbb{E}_{s_0 \sim \rho} \Big[ V_{\theta}(s_0) \Big].   \nonumber
\end{align}
Then we have the following result:
\begin{align}
    |L(\theta) - \widehat{L}(\theta;\mathcal{D})| = \bigg| \mathbb{E}_{\tau^{\rm E} \sim \pi^{\rm E}} \Big[ \sum_{t = 0}^{\infty} \gamma^t r(s_t, a_t; \theta) \Big] - \mathbb{E}_{\tau^{\rm E} \sim \mathcal{D}} \Big[ \sum_{t = 0}^{\infty} \gamma^t r(s_t, a_t; \theta) \Big] \bigg|. \nonumber
\end{align}
According to Assumption \ref{assumption:bound_reward}, we obtain that the discounted cumulative reward of any trajectory follows $0 \leq \sum_{t = 0}^{\infty} \gamma^t r(s_t, a_t; \theta) \leq \frac{C_r}{1 - \gamma} $. Then by applying Hoeffding’s inequality, for any $\epsilon > 0$, we have the following result:
\begin{align}
    P\bigg(\bigg| \mathbb{E}_{\tau \sim \pi^{\rm E}} \Big[ \sum_{t = 0}^{\infty} \gamma^t r(s_t, a_t; \theta) \Big] - \mathbb{E}_{\tau \sim \mathcal{D}} \Big[ \sum_{t = 0}^{\infty} \gamma^t r(s_t, a_t; \theta) \Big] \bigg| \geq \epsilon \bigg) \leq 2\exp \Big( -\frac{2|\mathcal{D}|\epsilon^2}{(\frac{C_r}{1 - \gamma})^2} \Big). \nonumber
\end{align}
Then by setting $\delta = 2\exp \Big( -\frac{2|\mathcal{D}|\epsilon^2}{(\frac{C_r}{1 - \gamma})^2} \Big) $, with probability greater than $1 - \delta$, we have
    \begin{align}
    &\bigg| \mathbb{E}_{\tau \sim \pi^{\rm E}} \Big[ \sum_{t = 0}^{\infty} \gamma^t r(s_t, a_t; \theta) \Big] - \mathbb{E}_{\tau \sim \mathcal{D}} \Big[ \sum_{t = 0}^{\infty} \gamma^t r(s_t, a_t; \theta) \Big] \bigg| \leq \frac{C_r}{1-\gamma} \sqrt{\frac{\ln(2/\delta)}{2|\mathcal{D}|}},   \label{hoeffding:cumulative_reward}
\end{align}
where $C_r$ is the constant defined in Assumption \ref{assumption:bound_reward}.
According to \eqref{hoeffding:cumulative_reward}, we obtain the concentration bound to quantify the approximation between $L(\theta)$ and $\widehat{L}(\theta; \mathcal{D})$ as below:
\begin{align}
    |L(\theta) - \widehat{L}(\theta;\mathcal{D})| \leq \frac{C_r}{1-\gamma} \sqrt{\frac{\ln(2/\delta)}{2|\mathcal{D}|}}, \quad \text{with probability greater than $1 - \delta$}.  \nonumber
\end{align}
This completes the proof of this lemma.
\hfill $\blacksquare$
}
	
\section{Proof of Lemma \ref{lemma:outer_gradient}}
\label{proof:outer_gradient_derivation}

\noindent \textbf{Proof.} Recall that we have defined $V_{\theta}$, $Q_{\theta}$ as the soft value function, soft Q-function under reward parameter $\theta$ and the optimal policy $\pi_{\theta}$ in an entropy-regularized MDP. Hence, the following expressions hold for any $s \in \mathcal{S}$ and $a \in \mathcal{A}$:
\begin{subequations}
    \begin{align}
    &V_{\theta}(s) := \mathbb{E}_{\tau^{\rm A} \sim \pi_{\theta}} \bigg[ \sum_{t = 0}^{\infty} \gamma^t \bigg( r(s_t, a_t;\theta) + \mathcal{H}(\pi_{\theta}(\cdot | s_t)) \bigg) \bigg | s_0 = s \bigg], \label{def:optimal_soft_V} \\
    &Q_{\theta}(s,a) := r(s,a; \theta) + \gamma \mathbb{E}_{s^\prime \sim P(\cdot| s,a)} \big[ V_{\theta}(s^\prime) \big]. \label{def:optimal_soft_Q}
\end{align}
\end{subequations}
 According to \cite{cen2021fast}, the soft value function $V_{\theta}$ and the policy $\pi_{\theta}$ in the entropy-regularized MDP satisfy the following relations for any $s \in \mathcal{S}$ and $a \in \mathcal{A}$: 
\begin{subequations}
    \begin{align}
    &V_{\theta}(s) =  \log \bigg( \sum_{\tilde{a} \sim \mathcal{A}} \exp Q_{\theta}(s,\tilde{a}) \bigg) , \label{eq:optimal_soft_V_expression} \\
    &\pi_{\theta}(a|s) = \frac{\exp \big( Q_{\theta}(s,a) \big)}{\sum_{\tilde{a} \in \mathcal{A}} \exp \big( Q_{\theta}(s,\tilde{a}) \big)}. \label{eq:policy_closed_form}
\end{align}
\end{subequations}
According to \eqref{rewrite:maximum_likelihood}, we are able to express the objective function $ L(\theta) $ in \eqref{eq:ML} as below:
	\begin{align}
		L(\theta) = \mathbb{E}_{\tau^{\rm E} \sim \pi^{\rm E}} \bigg[  \sum_{t = 0}^{\infty} \gamma^t   r(s_t, a_t; \theta)  \bigg] - \mathbb{E}_{s_0 \sim \rho} \bigg[ V_{\theta}(s_0) \bigg]. \label{eq:obj_expression}
	\end{align}
 
	Based on \eqref{eq:obj_expression}, we calculate the exact gradient of the objective function $ L(\theta) $ as below:
	\begin{align} 
		\nabla_{\theta} L(\theta) &:= \mathbb{E}_{\tau^{\rm E} \sim \pi^{\rm E}} \bigg[  \sum_{t = 0}^{\infty} \gamma^t  \nabla_{\theta} r(s_t, a_t; \theta)  \bigg] - \mathbb{E}_{s_0 \sim \rho} \bigg[ \nabla_{\theta} V_{\theta}(s_0) \bigg] \nonumber \\
		&\overset{(i)}{=} \mathbb{E}_{\tau^{\rm E} \sim \pi^{\rm E}} \bigg[  \sum_{t = 0}^{\infty} \gamma^t  \nabla_{\theta} r(s_t, a_t; \theta)  \bigg] - \mathbb{E}_{s_0 \sim \rho} \bigg[  \nabla_\theta  \log\bigg( \sum_{\tilde{a} \in \mathcal{A}} \exp Q_{\theta}(s_0, \tilde{a}) \bigg) \bigg]  \nonumber \\
		&= \mathbb{E}_{\tau^{\rm E} \sim \pi^{\rm E}} \bigg[  \sum_{t = 0}^{\infty} \gamma^t  \nabla_{\theta} r(s_t, a_t; \theta)  \bigg] - \mathbb{E}_{s_0 \sim \rho} \bigg[  \sum_{a \in \mathcal{A}} \bigg( \frac{ \exp Q_{\theta}(s_0, a) }{\sum_{\tilde{a} \in \mathcal{A}} \exp Q_{\theta}(s_0, \tilde{a}) } \nabla_\theta Q_{\theta}(s_0, a) \bigg) \bigg]  \nonumber \\
		&\overset{(ii)}{=} \mathbb{E}_{\tau^{\rm E} \sim \pi^{\rm E}} \bigg[  \sum_{t = 0}^{\infty} \gamma^t  \nabla_{\theta} r(s_t, a_t; \theta) \bigg] - \mathbb{E}_{s_0 \sim \rho} \bigg[  \sum_{a\in\mathcal{A}}  \pi_{\theta}(a | s_0) \nabla_\theta Q_{\theta}(s_0, a) \bigg].  \label{eq:obj_grad_expression}
	\end{align}
	where (i) follows \eqref{eq:optimal_soft_V_expression} and (ii) is from \eqref{eq:policy_closed_form}.
	Then we calculate $ \nabla_\theta Q_{\theta}(s_0, a_0) $ as below:
	\begin{align}
	    &\nabla_\theta Q_{\theta}(s_0, a_0) \nonumber \\
	    &\overset{(i)}{=} \nabla_{\theta} \bigg( r(s_0,a_0;\theta) + \gamma \ee_{s_1 \sim P(\cdot|s_0, a_0)} \big[ V_{\theta}(s_1) \big]  \bigg) \nonumber \\
	    &\overset{(ii)}{=} \nabla_{\theta} r(s_0,a_0;\theta) + \gamma \ee_{s_1 \sim P(\cdot|s_0, a_0)} \bigg[ \nabla_{\theta} \log\bigg( \sum_{\tilde{a} \in \mathcal{A}} \exp Q_{\theta}(s_0, \tilde{a}) \bigg) \bigg] \nonumber \\
	    &= \nabla_{\theta} r(s_0,a_0;\theta) + \gamma \ee_{s_1 \sim P(\cdot|s_0, a_0)} \bigg[ \sum_{a \in \mathcal{A}} \frac{\exp Q_{\theta}(s_1, a) }{\sum_{\tilde{a} \in \mathcal{A}} \exp Q_{\theta}(s_1, \tilde{a})} \nabla_{\theta} Q_{\theta}(s_1, a) \bigg] \nonumber \\
	    &\overset{(iii)}{=} \nabla_{\theta} r(s_0,a_0;\theta) + \gamma \ee_{s_1 \sim P(\cdot|s_0, a_0)} \bigg[ \sum_{a \in \mathcal{A}} \pi_{\theta}(a|s_1) \nabla_{\theta} Q_{\theta}(s_1, a) \bigg] \nonumber \\
	    &\overset{(iv)}{=} \nabla_{\theta} r(s_0,a_0;\theta) + \gamma \ee_{s_1 \sim P(\cdot|s_0, a_0), a_1 \sim \pi_{\theta}(\cdot|s_1)} \bigg[ \nabla_{\theta} \bigg( r(s_1,a_1;\theta) + \gamma \ee_{s_2 \sim P(\cdot|s_1, a_1)} \big[ V_{\theta}(s_2) \big]  \bigg) \bigg] \nonumber \\
	    &\overset{(v)}{=} \ee_{\tau^{\rm A} \sim \pi_{\theta}} \bigg[ \sum_{t = 0}^{\infty} \gamma^t \nabla_{\theta} r(s_t, a_t;\theta) \mid s_0, a_0 \bigg] \label{eq:optimal_soft_Q_grad}
	\end{align}
	where (i) and (iv) follows the definition of the soft Q-function, see \eqref{def:soft_Q_function}; (ii) follows \eqref{eq:optimal_soft_V_expression}; (iii) is from \eqref{eq:policy_closed_form}; (v) is shown by recursively applying (i) - (iv). 
	
	Finally, plugging equation \eqref{eq:optimal_soft_Q_grad} into \eqref{eq:obj_grad_expression}, the gradient expression of $L(\theta)$ follows:
	\begin{align}
		\nabla_{\theta} L(\theta) &= \mathbb{E}_{\tau^{\rm E} \sim \pi^{\rm E}} \bigg[  \sum_{t = 0}^{\infty} \gamma^t  \nabla_{\theta} r(s_t, a_t; \theta) \bigg] - \mathbb{E}_{s_0 \sim \rho} \bigg[  \sum_{a\in\mathcal{A}}  \pi_{\theta}(a | s_0) \nabla_\theta Q_{\theta}(s_0, a) \bigg] \nonumber \\
		&= \mathbb{E}_{\tau^{\rm E} \sim \pi^{\rm E}} \bigg[  \sum_{t = 0}^{\infty} \gamma^t  \nabla_{\theta} r(s_t, a_t; \theta) \bigg] - \mathbb{E}_{s_0 \sim \rho} \bigg[  \sum_{a\in\mathcal{A}}  \pi_{\theta}(a | s_0) \cdot  \ee_{\tau \sim \pi_{\theta}} \Big[ \sum_{t = 0}^{\infty} \gamma^t \nabla_{\theta} r(s_t, a_t;\theta) \mid s_0, a \Big] \bigg] \nonumber \\
		&=\mathbb{E}_{\tau^{\rm E} \sim \pi^{\rm E}}\bigg[ \sum_{t = 0}^{\infty} \gamma^t  \nabla_{\theta}r(s_t, a_t; \theta) \bigg] - \mathbb{E}_{ \tau^{\rm A} \sim \pi_{\theta}}\bigg[ \sum_{t = 0}^{\infty} \gamma^t \nabla_{\theta} r(s_t, a_t; \theta) \bigg]. \label{eq:obj_grad_exact_form}
	\end{align}
 Following the same proof steps, we can also show the gradient expression of the surrogate objective $\widehat{L}(\theta; \mathcal{D})$ as below:
 \begin{align}
		\nabla_{\theta} \widehat{L}(\theta;\mathcal{D}) =\mathbb{E}_{\tau^{\rm E} \sim \mathcal{D}}\bigg[ \sum_{t = 0}^{\infty} \gamma^t  \nabla_{\theta}r(s_t, a_t; \theta) \bigg] - \mathbb{E}_{ \tau^{\rm A} \sim \pi_{\theta}}\bigg[ \sum_{t = 0}^{\infty} \gamma^t \nabla_{\theta} r(s_t, a_t; \theta) \bigg]. \label{eq:estimation_obj_grad_form}
	\end{align}
	This completes the proof of this lemma.
\hfill $\blacksquare$

\section{Proof of Lemma \ref{lemma:Lipschitz_properties}} 
\label{proof:Lemma_Lipschitz_Properties}

To prove Lemma \ref{lemma:Lipschitz_properties}, we show the inequalities \eqref{ineq:soft_Q_Lipschitz} and  \eqref{ineq:objective_lipschitz_smooth} respectively. The constants $L_q$ and $L_c$ in Lemma \ref{lemma:Lipschitz_properties} has the expression: $$L_q := \frac{L_r}{1 - \gamma}, \quad L_c := \frac{2L_q L_r C_d \sqrt{ | \mathcal{S}| \cdot |\mathcal{A}| } }{1 - \gamma}  + \frac{2 L_g}{1 - \gamma}, $$
where $L_r$, $L_g$ are the constants defined in Assumption \ref{Assumption:reward_grad_bound} and $C_d$ is the constant in Lemma \ref{lemma:Lipschitz_visitation_measure}.

\subsection{Proof of Inequality \eqref{ineq:soft_Q_Lipschitz}}\label{sub:lip}
\noindent \textbf{Proof.} The proof of \eqref{ineq:soft_Q_Lipschitz} consists of two steps: 1) $ Q_{\theta} $ has bounded gradient with respect to any reward parameter $\theta$, 2) the inequality \eqref{ineq:soft_Q_Lipschitz} holds due to the mean value theorem.

Recall that in \eqref{eq:optimal_soft_Q_grad}, we have shown the explicit expression of $ \nabla_\theta Q_{\theta}(s,a) $ for any $s \in \mathcal{S}$ and $a \in \mathcal{A}$. Using this expression, we have the following relations:
	\begin{align}
		 \|  \nabla_\theta Q_{\theta}(s,a) \| & \overset{(i)}{ = }  \bigg \| \ee_{\tau^{\rm A} \sim \pi_{\theta}} \bigg[ \sum_{t = 0}^{\infty} \gamma^t \nabla_{\theta} r(s_t, a_t; \theta) ~ \Big| (s_0, a_0) = (s, a) \bigg]  \bigg \| \nonumber \\
		&\overset{(ii)}{\leq} \ee_{\tau^{\rm A} \sim \pi_{\theta}} \bigg[ \sum_{t = 0}^{\infty} \gamma^t  \big \| \nabla_{\theta} r(s_t, a_t; \theta) \big \| ~ \Big| (s_0, a_0) = (s, a) \bigg] \nonumber \\
		& \overset{(iii)}{\leq } \ee_{\tau^{\rm A} \sim \pi_{\theta}} \bigg[ \sum_{t = 0}^{\infty} \gamma^t  L_r ~ \Big| (s_0, a_0) = (s, a) \bigg] \nonumber \\
		& = \frac{L_r}{1 - \gamma} \label{ineq:soft_Q_grad_bound}
	\end{align}
	where (i) is from the equality \eqref{eq:optimal_soft_Q_grad} in the proof of Lemma \ref{lemma:outer_gradient}, (ii) follows Jensen's inequality and (iii) follows the inequality \eqref{ineq:Lipschitz_smooth_reward} in Assumption \ref{Assumption:reward_grad_bound}. To complete this proof, we use the Mean Value Theorem to show that 
	\begin{align}
	    |  Q_{\theta_1}(s,a)  - Q_{\theta_2}(s,a) | \leq \| \max_{\theta} \nabla_{\theta} Q_{\theta}(s,a) \| \cdot \| \theta_1 - \theta_2 \| \leq L_q \| \theta_{1} - \theta_2 \| \nonumber
	\end{align}
	where the last inequality follows \eqref{ineq:soft_Q_grad_bound} and we denote $L_q := \frac{L_r}{1 - \gamma} $. Therefore, we have proved the Lipschitz continuous inequality in  \eqref{ineq:soft_Q_Lipschitz}.
\hfill $\blacksquare$

\subsection{Proof of Inequality \eqref{ineq:objective_lipschitz_smooth}}
\noindent \textbf{Proof.} In this section, we prove the inequality \eqref{ineq:objective_lipschitz_smooth} in Lemma \ref{lemma:Lipschitz_properties}.

According to Lemma \ref{lemma:outer_gradient}, the gradient $\nabla_{\theta} \widehat{L}(\theta; \mathcal{D})$ has the expression as follows:
\begin{align}
    \nabla_{\theta} \widehat{L}(\theta; \mathcal{D}) = \mathbb{E}_{\tau^{\rm E} \sim \mathcal{D}}\bigg[ \sum_{t = 0}^{\infty} \gamma^t \nabla_{\theta} r(s_t, a_t; \theta) \bigg] - \mathbb{E}_{ \tau^{\rm A} \sim \pi_{\theta}}\bigg[ \sum_{t = 0}^{\infty} \gamma^t \nabla_{\theta} r(s_t, a_t; \theta) \bigg]. \label{restate:gradient_expression}
\end{align}

	Using the above relation, we have  
	\begin{align}
		& \| \nabla_{\theta} \widehat{L}(\theta_1; \mathcal{D}) - \nabla_{\theta} \widehat{L}(\theta_2; \mathcal{D}) \|  \nonumber \\
		&\overset{(i)}{=} \bigg \| \bigg( \mathbb{E}_{\tau^{\rm E} \sim \mathcal{D}}\bigg[ \sum_{t = 0}^{\infty} \gamma^t \nabla_{\theta} r(s_t, a_t; \theta_1) \bigg] - \mathbb{E}_{ \tau^{\rm A} \sim \pi_{\theta_1}}\bigg[ \sum_{t = 0}^{\infty} \gamma^t \nabla_{\theta} r(s_t, a_t; \theta_1) \bigg] \bigg)  \nonumber \\
		& \quad \quad - \bigg( \mathbb{E}_{\tau^{\rm E} \sim \mathcal{D}}\bigg[ \sum_{t = 0}^{\infty} \gamma^t \nabla_{\theta} r(s_t, a_t; \theta_2) \bigg] - \mathbb{E}_{ \tau^{\rm A} \sim \pi_{\theta_2}}\bigg[ \sum_{t = 0}^{\infty} \gamma^t \nabla_{\theta} r(s_t, a_t; \theta_2) \bigg] \bigg)  \bigg  \|  \nonumber \\
		& \leq \underbrace{\bigg \|   \mathbb{E}_{\tau^{\rm E} \sim \mathcal{D}}\bigg[ \sum_{t = 0}^{\infty} \gamma^t \nabla_{\theta} r(s_t, a_t; \theta_1) \bigg] -  \mathbb{E}_{\tau^{\rm E} \sim \mathcal{D}}\bigg[ \sum_{t = 0}^{\infty} \gamma^t \nabla_{\theta} r(s_t, a_t; \theta_2) \bigg] \bigg \|}_{\rm :=term ~ A}  +  \nonumber \\
		& \quad \quad  \underbrace{\bigg \|   \mathbb{E}_{ \tau^{\rm A} \sim \pi_{\theta_1}}\bigg[ \sum_{t = 0}^{\infty} \gamma^t \nabla_{\theta} r(s_t, a_t; \theta_1) \bigg] -  \mathbb{E}_{ \tau^{\rm A} \sim \pi_{\theta_2}}\bigg[ \sum_{t = 0}^{\infty} \gamma^t \nabla_{\theta} r(s_t, a_t; \theta_2) \bigg] \bigg \|}_{\rm :=term ~ B}    \label{ineq:grad_difference_separate}
	\end{align}
	where (i) follows the exact gradient expression in equation \eqref{restate:gradient_expression}. Then we separately analyze term A and term B in \eqref{ineq:grad_difference_separate}. 
	
	For term A, it follows that
	\begin{align}
		&\bigg \|   \mathbb{E}_{\tau^{\rm E} \sim \mathcal{D}}\bigg[ \sum_{t = 0}^{\infty} \gamma^t \nabla_{\theta} r(s_t, a_t; \theta_1) \bigg] -  \mathbb{E}_{\tau^{\rm E} \sim \mathcal{D}}\bigg[ \sum_{t = 0}^{\infty} \gamma^t \nabla_{\theta} r(s_t, a_t; \theta_2) \bigg] \bigg \| \nonumber \\
		& \overset{(i)}{\leq} \ee_{\tau^{\rm E} \sim \mathcal{D}} \bigg[ \sum_{t = 0}^{\infty} \gamma^t \big \|  \nabla_{\theta} r(s_t, a_t; \theta_1)  - \nabla_{\theta} r(s_t, a_t; \theta_2)  \big \| \bigg] \nonumber \\
		& \overset{(ii)}{\leq} \ee_{\tau^{\rm E} \sim \mathcal{D}} \bigg[ \sum_{t = 0}^{\infty} \gamma^t L_g \| \theta_{1}  - \theta_{2} \| \bigg] \nonumber \\
		& = \frac{L_g}{1 - \gamma} \| \theta_{1}  -  \theta_{2} \| \label{ineq:gradient_diff_term_A}
	\end{align}
	where (i) follows Jensen's inequality and (ii) is from \eqref{ineq:Lipschitz_smooth_reward} in Assumption \ref{Assumption:reward_grad_bound}. 
	
	For the term B, it holds that 
	\begin{align}
		& \bigg \|   \mathbb{E}_{ \tau^{\rm A} \sim \pi_{\theta_1}}\bigg[ \sum_{t = 0}^{\infty} \gamma^t \nabla_{\theta} r(s_t, a_t; \theta_1) \bigg] -  \mathbb{E}_{ \tau^{\rm A} \sim \pi_{\theta_2}}\bigg[ \sum_{t = 0}^{\infty} \gamma^t \nabla_{\theta} r(s_t, a_t; \theta_2) \bigg] \bigg \|  \nonumber \\
		&\overset{(i)}{\leq} \bigg \|	\mathbb{E}_{ \tau^{\rm A} \sim \pi_{\theta_1}}\bigg[ \sum_{t = 0}^{\infty} \gamma^t \nabla_{\theta} r(s_t, a_t; \theta_1) \bigg]  -  \mathbb{E}_{ \tau^{\rm A} \sim \pi_{\theta_2}}\bigg[ \sum_{t = 0}^{\infty} \gamma^t \nabla_{\theta} r(s_t, a_t; \theta_1) \bigg]	\bigg \|    \nonumber \\
		&\quad +  \bigg \| \mathbb{E}_{ \tau^{\rm A} \sim \pi_{\theta_2}}\bigg[ \sum_{t = 0}^{\infty} \gamma^t \nabla_{\theta} r(s_t, a_t; \theta_1) \bigg] -  \mathbb{E}_{ \tau^{\rm A} \sim \pi_{\theta_2}} \bigg[ \sum_{t = 0}^{\infty} \gamma^t \nabla_{\theta} r(s_t, a_t; \theta_2) \bigg]  \bigg \| 	\nonumber \\
		&\overset{(ii)}{ \leq }  \frac{1}{1 - \gamma}   \bigg  \|  \mathbb{E}_{(s,a) \sim d(\cdot, \cdot; \pi_{\theta_1})}\big[ \nabla_{\theta} r(s_t, a_t; \theta_1) \big] - \mathbb{E}_{(s,a) \sim d(\cdot, \cdot;\pi_{\theta_2})}\big[ \nabla_{\theta} r(s_t, a_t; \theta_1) \big]   \bigg \|  \nonumber\\
		& \quad + \mathbb{E}_{ \tau^{\rm A} \sim \pi_{\theta_2}} \bigg[ \sum_{t = 0}^{\infty} \gamma^t  \big \|	\nabla_{\theta} r(s_t, a_t; \theta_1)  -  \nabla_{\theta} r(s_t, a_t; \theta_2) \big \|  \bigg]   \nonumber \\
		&\overset{(iii)}{\leq} \frac{1}{1 - \gamma}   \bigg  \| \sum_{s \in \mathcal{S}, a \in \mathcal{A}}  \nabla_{\theta} r(s_t, a_t; \theta_1) \bigg(d(s,a; \pi_{\theta_1}) - d(s,a; \pi_{\theta_2})\bigg) \bigg \|  + \mathbb{E}_{ \tau^{\rm A} \sim \pi_{\theta_2}} \bigg[ \sum_{t = 0}^{\infty} \gamma^t L_g \| \theta_{1}  -  \theta_{2} \|  \bigg] \nonumber \\
		&\overset{(iv)}{\leq} \frac{2L_r}{1 - \gamma} \| d(\cdot, \cdot; \pi_{\theta_1}) - d(\cdot, \cdot; \pi_{\theta_1}) \|_{TV}  + \frac{L_g}{1 - \gamma} \| \theta_{1}  -  \theta_{2} \|  \label{ineq:gradient_diff_term_B}
	\end{align}
	where (i) follows the triangle inequality, (ii) is from Jensen's inequality and the definition of the discounted state-action visitation measure $d(s,a;\pi) := (1-\gamma)\pi(a|s)\sum_{t = 0}^{\infty} \gamma^t P^{\pi}(s_t = s | s_0 \sim \rho)$;  (iii) is from \eqref{ineq:Lipschitz_smooth_reward} in Assumption \ref{Assumption:reward_grad_bound}; (iv) is from \eqref{ineq:Lipschitz_smooth_reward} and the definition of the total variation norm.
	
	Plugging the inequalities \eqref{ineq:gradient_diff_term_A}, \eqref{ineq:gradient_diff_term_B} to \eqref{ineq:grad_difference_separate}, it holds that 
	\begin{align}
		\| \nabla_{\theta} \widehat{L}(\theta_1;\mathcal{D}) - \nabla_{\theta} \widehat{L}(\theta_2;\mathcal{D}) \|   &\leq \frac{2L_r}{1 - \gamma} \| d(\cdot, \cdot;\pi_{\theta_1}) - d(\cdot, \cdot;\pi_{\theta_2}) \|_{TV}  + \frac{2 L_g}{1 - \gamma} \| \theta_{1}  -  \theta_{2} \|   \nonumber \\
		& \overset{(i)}{\leq}  \frac{2L_r C_d}{1 - \gamma} \| Q_{\theta_1} - Q_{\theta_2}  \|  + \frac{2 L_g}{1 - \gamma} \| \theta_{1}  -  \theta_{2} \|   \nonumber \\
		& \overset{(ii)}{\leq}  \frac{2L_r C_d \sqrt{ | \mathcal{S}| \cdot |\mathcal{A}| } }{1 - \gamma} \| Q_{\theta_1} - Q_{\theta_2}   \|_{\infty}  + \frac{2 L_g}{1 - \gamma} \| \theta_{1}  -  \theta_{2} \|  \nonumber \\
		& \overset{(iii)}{\leq}  \bigg( \frac{2L_q L_r C_d \sqrt{ | \mathcal{S}| \cdot |\mathcal{A}| } }{1 - \gamma}  + \frac{2 L_g}{1 - \gamma}  \bigg) \| \theta_{1}  - \theta_{2}  \|.  \label{ineq:Lipschitz_Obj_derivation}
	\end{align}
	Given the fact that $\pi_{\theta}$ is a softmax policy parameterized by $Q_{\theta}$ where $\pi_{\theta}(a|s) \propto \exp(Q_{\theta}(s,a))$, we show the inequality (i) from the inequality \eqref{ineq:lipschitz_measure} in Lemma \ref{lemma:Lipschitz_visitation_measure}. Moreover, the inequality (ii) follows the conversion between the Frobenius norm and the infinty norm, where the inequality $|Q_{\theta_1}(s,a) - Q_{\theta_2}(s,a)| \leq \| Q_{\theta_1} - Q_{\theta_2} \|_{\infty}$ holds for any $s \in \mathcal{S}$ and $a \in \mathcal{A}$ so that $\| Q_{\theta_1} - Q_{\theta_2} \| \leq \sqrt{|\mathcal{S}| \cdot |\mathcal{A}|} ~ \| Q_{\theta_1} - Q_{\theta_2} \|_{\infty}$. Last, (iii) is from the inequality \eqref{ineq:soft_Q_Lipschitz} in Lemma \ref{lemma:Lipschitz_properties}. 
	
	Define the constant $L_c := \frac{2L_q L_r C_d \sqrt{ | \mathcal{S}| \cdot |\mathcal{A}| } }{1 - \gamma}  + \frac{2 L_g}{1 - \gamma}$, we have the following inequality:
	\begin{align}
	    \| \nabla_{\theta} \widehat{L}(\theta_1;\mathcal{D}) - \nabla_{\theta} \widehat{L}(\theta_2;\mathcal{D}) \| \leq L_c \| \theta_1 - \theta_2 \|. \nonumber
	\end{align}
	Therefore, we complete the proof of the inequality \eqref{ineq:objective_lipschitz_smooth} in Lemma \ref{lemma:Lipschitz_properties}. \hfill $\blacksquare$

\section{Proof of Lemma \ref{lemma:formulation_equivalence}}
\label{proof:formulation_equivalence_lemma}

\noindent \textbf{Proof.}
    Suppose the expert trajectories $\tau$ in \eqref{eq:ML} is sampled from an expert policy $\pi^E$. Moreover, we parameterize the state-only reward as $r(s;\theta)$. Then the objective function $L(\theta)$ can be rewritten as follows:
    \begin{align}
        L(\theta)&:= \mathbb{E}_{\tau^{\rm E} \sim \pi^E} \bigg[\sum_{t = 0}^{\infty} \gamma^t \log \pi_{\theta}(a_t|s_t) \bigg] \nonumber \\
		&\overset{(i)}{=} \mathbb{E}_{\tau^{\rm E} \sim \pi^E} \bigg[  \sum_{t = 0}^{\infty} \gamma^t   r(s_t; \theta)  \bigg] - \mathbb{E}_{s_0 \sim \eta(\cdot)} \bigg[ V_{\theta}(s_0)   \bigg] \nonumber \\
		&\overset{(ii)}{=} \mathbb{E}_{s_0 \sim \eta(\cdot)} \big[ V^{\rm E}(s_0)   \big] - \mathbb{E}_{s_0 \sim \eta(\cdot)} \big[ V_{\theta}(s_0) \big] - \mathbb{E}_{\tau^{\rm E} \sim \pi^E} \bigg[  \sum_{t = 0}^{\infty} \gamma^t \mathcal{H}(\cdot|s_t)  \bigg] \label{eq:likelihood_rewrite} 
    \end{align}
    where (i) follows \eqref{eq:obj_expression} and the fact that the reward is a state-only function $r(s;\theta)$; (ii) follows the definitions of the soft value function.
    
    Ignoring the constant term $\mathbb{E}_{\tau^{\rm E} \sim \pi^E} \big[  \sum_{t = 0}^{\infty} \gamma^t \mathcal{H}(\cdot|s_t)  \big]$ in \eqref{eq:likelihood_rewrite}, the maximum likelihood formulation \eqref{eq:ML} is equivalent to the following bi-level problem:
    \begin{align}
        &\min_{\theta} ~~ \ee_{s_0 \sim \eta(\cdot)}\big[ V_{\theta}(s_0) \big] - \ee_{s_0 \sim \eta(\cdot)}\big[ V_{\theta}^{\rm E}(s_0) \big] \nonumber \\
	    & {\rm s.t.} ~~ \pi_{\theta} := \arg \max_{\pi} ~ \mathbb{E}_{\tau^{\rm A} \sim \pi} \bigg[ \sum_{t = 0}^{\infty} \gamma^t \bigg( r(s_t; \theta) + \mathcal{H}(\pi(\cdot | s_t)) \bigg) \bigg]. \nonumber
    \end{align}
    Therefore, we complete the proof of Lemma \ref{lemma:formulation_equivalence}. As an alternative interpretation to \eqref{eq:ML:problem}, the formulation above aims to minimize the gap between the soft value function of $\pi_{\theta}$ and $\pi^E$ under the state-only IRL setting.
$\hfill \blacksquare$

\section{Proof of Lemma \ref{lemma:Lipschitz_Q_different_r}}
\noindent \textbf{Proof.}
	Based on the definition of soft Q-functions $Q_{k+
	\frac{1}{2}}$ and $Q_{k+1}$, it follows {\small
	\begin{align}
	    &Q_{k+\frac{1}{2}}(s,a) := r(s,a;\theta_k) +  \mathbb{E}_{\tau^{\rm A} \sim \pi_{k+1}} \bigg[ \sum_{t = 1}^{\infty} \gamma^t \bigg( r(s_t, a_t;\theta_k) + \mathcal{H}(\pi_{k+1}(\cdot | s_t)) \bigg) \bigg | (s_0, a_0) = (s,a) \bigg], \label{def:soft_Q_semi} \\
	    &Q_{k+1}(s,a) := r(s,a;\theta_{k+1}) +  \mathbb{E}_{\tau^{\rm A} \sim \pi_{k+1}} \bigg[ \sum_{t = 1}^{\infty} \gamma^t \bigg( r(s_t, a_t;\theta_{k+1}) + \mathcal{H}(\pi_{k+1}(\cdot | s_t)) \bigg) \bigg | (s_0, a_0) = (s,a) \bigg]. \label{def:soft_Q_updated}
	\end{align}}
	Then it holds that
	\begin{align}
		| Q_{k+\frac{1}{2}}(s,a) -  Q_{k+1}(s,a) | 
		&= \bigg | \mathbb{E}_{\tau^{\rm A} \sim \pi_{k+1}} \bigg[ \sum_{t = 0}^{\infty} \gamma^t \Big( r(s_t, a_t; \theta_k) -  r(s_t, a_t; \theta_{k+1})  \Big) \Big | (s_0, a_0) = (s,a) \bigg] \bigg|  \nonumber  \\
		&\overset{(i)}{\leq}  \mathbb{E}_{\tau^{\rm A} \sim \pi_{k+1}} \bigg[ \sum_{t = 0}^{\infty} \gamma^t \big| r(s_t, a_t; \theta_k) -  r(s_t, a_t; \theta_{k+1})  \big| \Big | (s_0, a_0) = (s,a) \bigg]  \nonumber  \\
		&\overset{(ii)}{=} \mathbb{E}_{\tau^{\rm A} \sim \pi_{k+1}} \bigg[ \sum_{t = 0}^{\infty} \gamma^t \big | \big \langle \nabla_{\theta} r(s_t, a_t; \theta^c),  \theta_k - \theta_{k+1} \big \rangle \big | \Big | (s_0, a_0) = (s,a) \bigg]  \nonumber  \\
		&\overset{(iii)}{=} \mathbb{E}_{\tau^{\rm A} \sim \pi_{k+1}} \bigg[ \sum_{t = 0}^{\infty} \gamma^t \big \| \nabla_{\theta} r(s_t, a_t; \theta^c) \big \| \cdot \big \|  \theta_k - \theta_{k+1} \big \| \Big | (s_0, a_0) = (s,a)  \bigg]  \nonumber  \\
		&\leq  \mathbb{E}_{\tau^{\rm A} \sim \pi_{k+1}} \bigg[ \sum_{t = 0}^{\infty} \gamma^t \big \|\max_{\theta} \nabla_{\theta} r(s_t, a_t; \theta)
		 \big \| \cdot \big \| \theta_k - \theta_{k+1} \big \| \Big | (s_0, a_0) = (s,a) \bigg]  \nonumber  \\
		&\overset{(iv)}{\leq} \mathbb{E}_{\tau^{\rm A} \sim \pi_{k+1}} \bigg[ \sum_{t = 0}^{\infty} \gamma^t L_r \| \theta_k - \theta_{k+1} \| \bigg]  \nonumber \\
		&= \frac{ L_r }{1 - \gamma}  \| \theta_k - \theta_{k+1} \| \label{difference:Q_different_r}
	\end{align}
	where (i) follows Jensen's inequality; (ii) follows the mean value theorem where $\theta^c$ is a convex combination of $\theta_k$ and $\theta_{k+1}$; (iii) follows the Cauchy–Schwarz inequality; (iv) follows inequality \eqref{ineq:Lipschitz_smooth_reward} in Assumption \ref{Assumption:reward_grad_bound}. 
	\hfill $\blacksquare$

\section{Proof of Lemma \ref{lemma:approx_policy_improvement}}
\label{proof:approx_policy_improvement}
In this section, we prove the inequalities \eqref{ineq:approx_soft_policy_improvement} and \eqref{ineq:soft_Q_contraction} respectively.

\subsection{Proof of Inequality \eqref{ineq:approx_soft_policy_improvement}}
\noindent \textbf{Proof.} Recall the definitions of the soft value function and the soft Q-function in \eqref{def:soft_Value_function} - \eqref{def:soft_Q_function}. Moreover, we also defined the soft Q-function $Q_{k+\frac{1}{2}}$ in \eqref{def:soft_Q_semi}. Similarly, let us define the corresponding soft value function $V_{k+\frac{1}{2}}$ (under reward parameter $\theta_k$ and policy $\pi_{k+1}$) as: 
\begin{align}
    V_{k+\frac{1}{2}}(s) := \mathbb{E}_{\tau^{\rm A} \sim \pi_{k+1}} \bigg[ \sum_{t = 0}^{\infty} \gamma^t \bigg( r(s_t, a_t;\theta_k) + \mathcal{H}(\pi_{k+1}(\cdot | s_t)) \bigg) \bigg | s_0 = s \bigg], \quad \forall s \in \mathcal{S}. \label{def:soft_value_function_semi}
\end{align}

According to the definitions of $V_k$ in \eqref{def:soft_Value_function} and $V_{k + \frac{1}{2}}$ in \eqref{def:soft_value_function_semi}, we could rewrite $Q_{k}$ and $Q_{k+\frac{1}{2}}$ as:
\begin{align}
	    &Q_{k}(s,a) := r(s,a;\theta_k) + \gamma \mathbb{E}_{s^\prime \sim P(\cdot|s,a)} \big[ V_k(s^\prime) \big], \nonumber \\
	    &Q_{k+\frac{1}{2}}(s,a) := r(s,a;\theta_k) + \gamma \mathbb{E}_{s^\prime \sim P(\cdot|s,a)} \big[ V_{k+\frac{1}{2}}(s^\prime) \big]. \nonumber
	\end{align}
According to the expressions above, the following relation holds for any $s\in \mathcal{S}$ and $a \in \mathcal{A}$:
\begin{align}
    Q_{k}(s,a) - Q_{k+\frac{1}{2}}(s,a) = \gamma \mathbb{E}_{s^\prime \sim P(\cdot|s,a)}\big[ V_{k}(s^\prime) - V_{k+\frac{1}{2}}(s^\prime) \big]. \label{ineq:approx_soft_q_gap}
\end{align}
In order to measure the difference between $Q_{k+\frac{1}{2}}$ and $Q_{k}$, we need to bound the gap between $V_{k+\frac{1}{2}}$ and $V_{k}$. Here, we could define an auxiliary sequence $\{ \tilde{\pi}_k \}_{k\geq0}$ generated as below:
\begin{align}
    \tilde{\pi}_{k+1}(\cdot|s) \propto \exp \big( Q_k(s,\cdot) \big) \quad \forall s \in \mathcal{S}. \label{def:auxiliary_policy}
\end{align}
As a comparison, let us recall the approximated soft policy iteration \eqref{def:approximated_SPI}:
\begin{align}
    \pi_{k+1}(a|s) \propto \exp\big( \widehat{Q}_{k}(s,a) \big), \text{ where } \| \widehat{Q}_{k} - Q_k \|_{\infty} \leq \epsilon_{\rm app}. \label{update:approx_policy_updated}
\end{align}
Then for any $s \in \mathcal{S}$, we have the following series of relations:
\begin{align}
    V_k(s) &\overset{(i)}{=} \mathbb{E}_{a \sim \pi_k(\cdot|s)}\big[ -\log \pi_k(a|s) + Q_k(s,a) \big] \nonumber \\
    &= \mathbb{E}_{a \sim \pi_k(\cdot|s)}\big[ -\log \tilde{\pi}_{k+1}(a|s) + Q_k(s,a) \big] + \mathbb{E}_{a \sim \pi_k(\cdot|s)}\big[ \log \tilde{\pi}_{k+1}(a|s) - \log \pi_k(a|s) \big] \nonumber \\
    &\overset{(ii)}{=} \mathbb{E}_{a \sim \pi_k(\cdot|s)}\bigg[ \log \bigg( \sum_{\tilde{a} \in \mathcal{A}} \exp \big( Q_k(s, \tilde{a}) \big) \bigg) \bigg] - D_{KL}\bigg(\pi_k(\cdot|s) ~ || ~ \tilde{\pi}_{k+1}(\cdot|s)\bigg) \nonumber \\
    &\overset{(iii)}{\leq} \mathbb{E}_{a \sim \pi_{k+1}(\cdot|s)}\bigg[ \log \bigg( \sum_{\tilde{a} \in \mathcal{A}} \exp \big( Q_k(s, \tilde{a}) \big) \bigg) \bigg] \nonumber \\
    &\overset{(iv)}{=} \mathbb{E}_{a \sim \pi_{k+1}(\cdot|s)}\bigg[ -\log \tilde{\pi}_{k+1}(a|s) + Q_k(s,a) \bigg] \nonumber \\
    &= \mathbb{E}_{a \sim \pi_{k+1}(\cdot|s)}\bigg[ -\log \pi_{k+1}(a|s) + Q_k(s,a) \bigg] + \mathbb{E}_{a \sim \pi_{k+1}(\cdot|s)}\bigg[ \log \pi_{k+1}(a|s) - \log \tilde{\pi}_{k+1}(a|s) \bigg] \nonumber \\
    &\overset{(iv)}{\leq} \mathbb{E}_{a \sim \pi_{k+1}(\cdot|s)}\bigg[ -\log \pi_{k+1}(a|s) + Q_k(s,a) \bigg] + \| \log \pi_{k+1} - \log \tilde{\pi}_{k+1}  \|_{\infty} \label{bound:soft_value_function} 
\end{align}
where (i) follows the definition of the soft value function $V_k$; (ii) follows the fact in \eqref{def:auxiliary_policy} that $ \tilde{\pi}_{k+1}(a|s) := \frac{\exp Q_k(s,a)}{\sum_{\tilde{a} \in \mathcal{A}} \exp Q_k(s,\tilde{a})} $ and the definition of the KL divergence; (iii) follows the non-negativity of the KL divergence and the fact that $ \log \bigg( \sum_{\tilde{a} \in \mathcal{A}} \exp \big(Q_k(s, \tilde{a}) \big) \bigg) $ is independent of any action $a \in \mathcal{A}$; (iv) follows the fact that $-\log \tilde{\pi}_{k+1}(a|s) = \log\big( \sum_{\tilde{a} \in \mathcal{A}} \exp Q_k(s,\tilde{a}) \big) - Q_k(s,a)$, which is derived from the definition of $\tilde{\pi}_{k+1}$ in \eqref{def:auxiliary_policy};  (iv) follows the definition of the infinity norm $\| \cdot \|_{\infty}$ such that $\| \log \pi_{k+1} - \log \tilde{\pi}_{k+1}  \|_{\infty} = \max_{s\in \mathcal{S}} \max_{a \in \mathcal{A}} \big(\log \pi_{k+1}(a|s) - \log \tilde{\pi}_{k+1}(a|s)\big)$. 

We further analyze the approximation error $ \| \log \pi_{k+1} - \log \tilde{\pi}_{k+1}  \|_{\infty} $ in \eqref{bound:soft_value_function}. We first show that for any $s \in \mathcal{S}$ and $a \in \mathcal{A}$, the following relations hold:
	\begin{align}
	    &\big| \log \pi_{k+1}(a|s)  - \log \tilde{\pi}_{k+1}(a|s) \big| \nonumber \\
	    &\overset{(i)}{=} \bigg| \log \bigg(\frac{\exp \widehat{Q}_{k}(s,a) }{\sum_{\tilde{a} \in \mathcal{A}} \exp \widehat{Q}_{k}(s,\tilde{a})} \bigg) - \log \bigg(\frac{\exp Q_{k}(s,a)}{\sum_{\tilde{a} \in \mathcal{A}} \exp Q_{k}(s,\tilde{a}) } \bigg) \bigg|  \nonumber \\
	    &\overset{(ii)}{\leq} \bigg| \widehat{Q}_{k}(s,a) - Q_{k}(s,a) \bigg| + \bigg| \log\bigg( \sum_{\tilde{a} \in \mathcal{A}} \exp \widehat{Q}_{k}(s,\tilde{a}) \bigg) - \log \bigg( \sum_{\tilde{a} \in \mathcal{A}} \exp Q_{k}(s,\tilde{a}) \bigg) \bigg| \label{ineq:log_policy_gap}
	\end{align}
	where (i) follows \eqref{def:auxiliary_policy} and \eqref{update:approx_policy_updated}; (ii) follows the triangle inequality. We further analyze the second term in \eqref{ineq:log_policy_gap}. 
	
	We first define a short-handed notation $\log(\| \exp(v) \|_1) := \log(\| \sum_{\tilde{a} \in \mathcal{A}} \exp(v_{\tilde{a}}) \|_1)$, where the vector $v \in \mathbb{R}^{|\mathcal{A}|}$ and $v = [v_1, v_2, \cdots, v_{|\mathcal{A}|}]$. Then for any $v^\prime, v^{\prime\prime} \in \mathbb{R}^{|\mathcal{A}|}$, we have the following relation: 
	\begin{align}
	    \big| \log\big( \| \exp(v^\prime) \|_1 \big) - \log\big( \| \exp(v^{\prime\prime}) \|_1 \big) \big| &\overset{(i)}{=} \big | \big \langle v^\prime - v^{\prime\prime}, \nabla_{v} \log\big( \| \exp(v) \|_1 \big) |_{v = v^c}  \big \rangle \big | \nonumber \\
	    &\leq \|  v^\prime - v^{\prime\prime}  \|_{\infty} \cdot \|  \nabla_{v} \log\big( \| \exp(v) \|_1 \big) |_{v = v^c}  \|_1 \nonumber \\
	    &\overset{(ii)}{=} \|  v^\prime - v^{\prime\prime}  \|_{\infty} \label{ineq:Lipscitz_log_exp_operator}
	\end{align}
	where (i) follows the mean value theorem and $v_c$ is a solution lies between $v^\prime$ and $v^{\prime\prime}$; (ii) follows the following relations:
	\begin{align}
	    [\nabla_{v} \log\big( \| \exp(v) \|_1 \big)]_i = \frac{\exp(v_i)}{\sum_{1 \leq a \leq |\mathcal{A}|} \exp(v_a)}, \quad  \|\nabla_{v} \log\big( \| \exp(v) \|_1 \big)\|_1 = 1, \quad \forall ~ v \in \mathbb{R}^{|\mathcal{A}|}. \nonumber
	\end{align}
	Through plugging \eqref{ineq:Lipscitz_log_exp_operator} into \eqref{ineq:log_policy_gap}, it holds that 
	\begin{align}
	    &\big| \log \pi_{k+1}(a|s) - \log \tilde{\pi}_{k+1}(a|s) \big| \leq \big| \widehat{Q}_{k}(s,a) - Q_{k}(s,a) \big| + \max_{\tilde{a} \in \mathcal{A}} \big| \widehat{Q}_{k}(s,\tilde{a}) - Q_{k}(s,\tilde{a}) \big|. \label{ineq:log_pi_soft_Q}
	\end{align}
	Taking the infinity norm over $\mathbb{R}^{|\mathcal{S}| \times |\mathcal{A}|}$, the following result holds:
	\begin{align}
	   \| \log \pi_{k+1} - \log \tilde{\pi}_{k+1} \|_{\infty} \leq 2 \| \widehat{Q}_{k} - Q_{k} \|_{\infty}.
	   \label{ineq:policy_lipschitzness}
	\end{align}
Through plugging \eqref{ineq:policy_lipschitzness} into \eqref{bound:soft_value_function}, it follows that 
\begin{align}
    V_k(s) &\leq \mathbb{E}_{a \sim \pi_{k+1}(\cdot|s)}\bigg[ -\log \pi_{k+1}(a|s) + Q_k(s,a) \bigg] + \| \log \pi_{k+1} - \log \tilde{\pi}_{k+1}  \|_{\infty} \nonumber \\
    &\overset{(i)}{\leq} \mathbb{E}_{a \sim \pi_{k+1}(\cdot|s)}\bigg[ -\log \pi_{k+1}(a|s) + Q_k(s,a) \bigg] + 2\| \widehat{Q}_k - Q_k \|_{\infty} \nonumber \\
    &\overset{(ii)}{=} \mathbb{E}_{a \sim \pi_{k+1}(\cdot|s), s^\prime \sim P(\cdot|s,a)}\bigg[ -\log \pi_{k+1}(a|s) + r(s,a; \theta_k) + \gamma V_k(s^\prime) \bigg] + 2\epsilon_{\rm app} \label{bound:soft_value_function_iterate}
\end{align}
where (i) is from \eqref{ineq:policy_lipschitzness}; (vi) follows the definition of the soft Q-function and the fact that the approximation error $ \| \widehat{Q}_k - Q_k \|_{\infty} $ is bounded by $\epsilon_{\rm app}$. By recursively using the inequality \eqref{bound:soft_value_function_iterate}, the following result hold for any $s \in \mathcal{S}$: 
\begin{align}
    V_k(s) &\leq \mathbb{E}_{\tau^{\rm A} \sim \pi_{k+1}} \bigg[ \sum_{t = 0}^{\infty} \bigg( -\log \pi_{k+1}(a_t|s_t) + r(s_t, a_t; \theta_k) \bigg) \bigg{|} s_0 = s\bigg] + \frac{2\epsilon_{\rm app}}{1 - \gamma} \overset{(i)}{=} V_{k+\frac{1}{2}}(s) + \frac{2\epsilon_{\rm app}}{1 - \gamma} \label{gap:approx_soft_V}
\end{align}
where (i) follows the definition of $V_{k+\frac{1}{2}}$ in \eqref{def:soft_value_function_semi}. By plugging \eqref{gap:approx_soft_V} into \eqref{ineq:approx_soft_q_gap}, we finish the proof.
\hfill $\blacksquare$

\subsection{Proof of Inequality \eqref{ineq:soft_Q_contraction}}
\noindent \textbf{Proof. } 
For any $s \in \mathcal{S}$ and $a \in \mathcal{A}$, the following results hold{\small
\begin{align}
    & \quad Q_{\theta_k}(s,a) - Q_{k+\frac{1}{2}}(s,a) \nonumber \\
    &\overset{(i)}{=} \bigg( r(s,a;\theta_k) + \gamma \mathbb{E}_{s^\prime \sim P(\cdot| s,a)} \bigg[ \log \big( \sum_{\tilde{a} \in \mathcal{A}} \exp Q_{\theta_k}(s^\prime, \tilde{a}) \big) \bigg] \bigg) \nonumber \\
    &\quad - \bigg( r(s,a;\theta_k) + \gamma \mathbb{E}_{s^\prime \sim P(\cdot| s,a), a^\prime \sim \pi_{k+1}(\cdot|s^\prime)} \bigg[ -\log \pi_{k+1}(a^\prime | s^\prime) + Q_{k+\frac{1}{2}}(s^\prime, a^\prime) \bigg] \bigg) \nonumber \\
    &\overset{(ii)}{=}  \gamma \mathbb{E}_{s^\prime \sim P(\cdot| s,a)} \bigg[ \log \big( \sum_{\tilde{a} \in \mathcal{A}} \exp Q_{\theta_k}(s^\prime, \tilde{a}) \big) \bigg] - \gamma \mathbb{E}_{s^\prime \sim P(\cdot| s,a), a^\prime \sim \pi_{k+1}(\cdot|s^\prime)} \bigg[ -\log \bigg( \frac{\exp \widehat{Q}_k(s^\prime, a^\prime)}{\sum_{\tilde{a} \in \mathcal{A}}\exp \widehat{Q}_k(s^\prime, \tilde{a})} \bigg) + Q_{k+\frac{1}{2}}(s^\prime, a^\prime) \bigg] \nonumber \\
    &= \gamma \mathbb{E}_{s^\prime \sim P(\cdot| s,a)} \bigg[ \log \big( \sum_{\tilde{a} \in \mathcal{A}} \exp Q_{\theta_k}(s^\prime, \tilde{a}) \big) -  \log \big( \sum_{\tilde{a} \in \mathcal{A}} \exp \widehat{Q}_{k}(s^\prime, \tilde{a} ) \big) \bigg] \nonumber \\
    &\quad - \gamma \mathbb{E}_{s^\prime \sim P(\cdot| s,a), a^\prime \sim \pi_{k+1}(\cdot|s^\prime)} \bigg[ Q_{k+\frac{1}{2}}(s^\prime, a^\prime) - \widehat{Q}_{k}(s^\prime, a^\prime) \bigg]  \nonumber \\
    &\overset{(iii)}{\leq} \gamma \mathbb{E}_{s^\prime \sim P(\cdot| s,a)} \bigg[ \max_{\tilde{a} \in \mathcal{A}}\big| Q_{\theta_k}(s^\prime, \tilde{a}) - \widehat{Q}_k(s^\prime, \tilde{a}) \big| \bigg] - \gamma \min_{s \in \mathcal{S}, a \in \mathcal{A}} \bigg( Q_{k+\frac{1}{2}}(s,a) - \widehat{Q}_{k}(s,a) \bigg) \nonumber \\
    &\overset{(iv)}{\leq} \gamma \big \|  Q_{\theta_k} - \widehat{Q}_k \big \|_{\infty} - \gamma \min_{s \in \mathcal{S}, a \in \mathcal{A}} \bigg( Q_{k+\frac{1}{2}}(s,a) - \widehat{Q}_{k}(s,a) \bigg) \nonumber \\
    &\leq \gamma \| \widehat{Q}_k - Q_k \|_{\infty} + \gamma \| Q_{\theta_k} - Q_k \|_{\infty} - \gamma \min_{s \in \mathcal{S}, a \in \mathcal{A}} \bigg( Q_{k+\frac{1}{2}}(s,a) - \widehat{Q}_{k}(s,a) \bigg) \nonumber \\
    &\leq \gamma \epsilon_{\rm app} + \gamma \| Q_{\theta_k} - Q_k \|_{\infty} - \gamma \min_{s \in \mathcal{S}, a \in \mathcal{A}} \bigg( Q_{k+\frac{1}{2}}(s,a) - \widehat{Q}_{k}(s,a) \bigg) \label{bound:soft_Q_contraction}
\end{align}}
where (i) follows \eqref{def:optimal_soft_Q}, \eqref{eq:optimal_soft_V_expression} and the definition of the soft Q-function $Q_{k+\frac{1}{2}}$ in \eqref{def:soft_Q_semi}; (ii) is from the definition of $\pi_{k+1}$ in \eqref{def:approximated_SPI}; (iii) follows \eqref{ineq:Lipscitz_log_exp_operator}; (iv) follows the definition of the infinity norm such that $ \big \|  Q_{\theta_k} - \widehat{Q}_k \big \|_{\infty} = \max_{s \in \mathcal{S}} \max_{a \in \mathcal{A}} | Q_{\theta_k}(s,a) - \widehat{Q}_k(s,a) | $.

Given the fact that $Q_{\theta_k}$ is the soft Q-function under the reward function $r(\cdot, \cdot; \theta_k)$ and the optimal policy $\pi_{\theta_k}$, we have the relation that 
\begin{align}
    Q_{\theta_k}(s,a) - Q_{k+\frac{1}{2}}(s,a) \geq 0, \quad \forall s \in \mathcal{S}, a \in \mathcal{A}. \label{gap:soft_Q}
\end{align}
This implies that
\begin{align}
    \max_{s\in \mathcal{S},a\in\mathcal{A}}\{Q_{\theta_k}(s,a) - Q_{k+\frac{1}{2}}(s,a)\} = \| Q_{\theta_k} - Q_{k+\frac{1}{2}} \|_{\infty}.
\end{align}
Combining \eqref{gap:soft_Q} and \eqref{bound:soft_Q_contraction}, we bound the absolute difference between $Q_{\theta_k}$ and $Q_{k+\frac{1}{2}}$ as below:
\begin{align}
    \| Q_{\theta_k} - Q_{k+\frac{1}{2}} \|_{\infty} \leq \gamma \epsilon_{\rm app} + \gamma \| Q_{\theta_k} - Q_k \|_{\infty} - \gamma \min_{s \in \mathcal{S}, a \in \mathcal{A}} \bigg( Q_{k+\frac{1}{2}}(s,a) - \widehat{Q}_{k}(s,a) \bigg). \label{ineq:soft_Q_optimality_gap}
\end{align}
Then we further bound the last term above. For any $s \in \mathcal{S}$ and $a \in \mathcal{A}$, note that:
\begin{align}
    &Q_{k+\frac{1}{2}}(s,a) - \widehat{Q}_{k}(s,a) \nonumber \\
    &= \big( Q_{k+\frac{1}{2}}(s,a) - Q_{k}(s,a) \big) + \big( Q_{k}(s,a) - \widehat{Q}_{k}(s,a) \big) \nonumber \\
    &\geq - \frac{2\gamma \epsilon_{\rm app}}{1 - \gamma} - \epsilon_{\rm app} \nonumber
\end{align}
where the last inequality follows \eqref{ineq:approx_soft_policy_improvement} and the definition of the approximation error $\epsilon_{\rm app}$.

Through plugging the inequality above into \eqref{ineq:soft_Q_optimality_gap}, it holds that:
\begin{align}
    \| Q_{\theta_k} - Q_{k+\frac{1}{2}} \|_{\infty} \leq \gamma \| Q_{\theta_k} - Q_k \|_{\infty} + \gamma \epsilon_{\rm app} + \gamma \big(\epsilon_{\rm app} + \frac{2\gamma\epsilon_{\rm app}}{1-\gamma} \big) = \gamma \| Q_{\theta_k} - Q_k \|_{\infty} + \frac{2 \gamma \epsilon_{\rm app}}{ 1 - \gamma}. \label{absolute_bound:soft_Q_optimality_gap}
\end{align}
This completes the proof of the lemma. \hfill $\blacksquare$

\end{document}